\documentclass{article}

    \PassOptionsToPackage{numbers, compress}{natbib}

\usepackage[final]{neurips_2024}

\usepackage[utf8]{inputenc} 
\usepackage[T1]{fontenc}    
\usepackage{hyperref}       
\usepackage{url}            
\usepackage{booktabs}       
\usepackage{amsfonts}       
\usepackage{nicefrac}       
\usepackage{microtype}      
\usepackage{xcolor}         

\usepackage{amsmath}
\usepackage{amssymb}
\usepackage{amsthm}
\usepackage{abbreviations}
\usepackage{algorithm}
\usepackage{algorithmic}
\usepackage{cleveref}
\usepackage{subcaption} 
\usepackage{graphicx} 
\usepackage{pifont}
\usepackage{optidef} 
\usepackage{tikz} 
\usetikzlibrary{shapes,arrows,positioning}
\usepackage{bigints} 

\addtocontents{toc}{\protect\setcounter{tocdepth}{0}}

\title{Randomized Exploration for Reinforcement Learning with Multinomial Logistic Function Approximation}

\author{%
    Wooseong Cho\thanks{Equal contribution}
    \\
    Seoul National University
    \\
    Seoul, South Korea\\
    \texttt{wooseong\_cho@snu.ac.kr}
    \And
    Taehyun Hwang\footnotemark[1]
    \\
    Seoul National University
    \\
    Seoul, South Korea\\
    \texttt{th.hwang@snu.ac.kr}
    \AND
    Joongkyu Lee
    \\
    Seoul National University
    \\
    Seoul, South Korea\\
    \texttt{jklee0717@snu.ac.kr}
    \And
    Min-hwan Oh\thanks{Corresponding author}
    \\
    Seoul National University
    \\
    Seoul, South Korea\\
    \texttt{minoh@snu.ac.kr}
}

\begin{document}

\maketitle

\begin{abstract}
We study reinforcement learning with \textit{multinomial logistic} (MNL) function approximation where the underlying transition probability kernel of the \textit{Markov decision processes} (MDPs) is parametrized by an unknown transition core with features of state and action.
For the finite horizon episodic setting with inhomogeneous state transitions, we propose provably efficient algorithms with randomized exploration having frequentist regret guarantees.
For our first algorithm, $\texttt{RRL-MNL}$, we adapt optimistic sampling to ensure the optimism of the estimated value function with sufficient frequency.
We establish that $\texttt{RRL-MNL}$ achieves a $\tilde{\mathcal{O}}(\kappa^{-1} d^{\frac{3}{2}} H^{\frac{3}{2}} \sqrt{T})$ frequentist regret bound with constant-time computational cost per episode.
Here, $d$ is the dimension of the transition core, $H$ is the horizon length, $T$ is the total number of steps, and $\kappa$ is a problem-dependent constant.
Despite the simplicity and practicality of $\texttt{RRL-MNL}$, its regret bound scales with $\kappa^{-1}$, which is potentially large in the worst case.
To improve the dependence on $\kappa^{-1}$, we propose $\texttt{ORRL-MNL}$, which estimates the value function using the local gradient information of the MNL transition model. We show that its frequentist regret bound is $\tilde{\mathcal{O}}(d^{\frac{3}{2}} H^{\frac{3}{2}} \sqrt{T} + \kappa^{-1} d^2 H^2)$.
To the best of our knowledge, these are the first randomized RL algorithms for the MNL transition model that achieve statistical guarantees with constant-time computational cost per episode.
Numerical experiments demonstrate the superior performance of the proposed algorithms.
\end{abstract}

\section{Introduction} \label{sec:introduction}

\textit{Reinforcement learning} (RL) is a sequential decision-making problem in which an agent tries to maximize its expected cumulative reward by interacting with an unknown environment over time.
Despite significant empirical progress in RL algorithms for various applications~\citep{kober2013reinforcement, mnih2015human, silver2017mastering, silver2018general, fawzi2022discovering}, the theoretical understanding of RL algorithms had long been limited to tabular methods~\citep{jaksch2010near, osband2014modelbased, azar2017minimax, zhang2020almostoptimal, zhang2021model}, which explicitly enumerate the entire state and action spaces and learn the value (or the policy) for each state and action.
Recently, there has been an increasing body of research in RL with function approximation to extend beyond the tabular problem setting.
In particular, \textit{linear function approximation} has served as a foundational model~\citep{jin2020provably, zanette2020frequentist, du2020is, ayoub2020model, ishfaq2021randomized}.
On the other hand, the linear transition model assumption poses significant constraints: 1) the output of the function must be within $[0,1]$, and 2) the sum of the probabilities for all possible next states must be exactly 1.
These constraints make it challenging to apply RL with linear function approximation to real-world applications~\cite{hwang2023model}.
To overcome such challenges, there has been literature on RL with general function approximation
~\cite{du2021bilinear, foster2021statistical, ishfaq2021randomized, jin2021bellman, agarwal2022model, chen2023a}.
Despite the guarantee of sample efficiency achieved by their algorithms, this accomplishment might be impeded by computational intractability or the necessity to rely on stronger assumptions.
As a result, the resulting methods may not be as general or practical.

On the other hand,~\citet{hwang2023model} introduce specific non-linear parametric MDPs called MNL-MDPs (Assumption~\ref{assm:mnl-mdp}) where the transition probability of MDPs is given by an MNL model.
They consider an \textit{upper confidence bound} (UCB) approach to balance exploration and exploitation. 
Since it is costly or even intractable to compute UCB explicitly, randomized exploration methods such as \textit{Thompson Sampling} (TS) are widely studied in RL with linear function approximation as well as tabular MDPs.
This is because, in various decision-making problems ranging from multi-armed bandits to RL, randomized exploration algorithms have been shown to perform better than UCB methods in empirical evaluations~\citep{chapelle2011empirical, osband2017posterior, russo2018tutorial, kveton2020perturbed}. 
Furthermore, randomized exploration can be easily integrated with linear function approximation.
This is because the value function in linear MDPs can be linearly parameterized, allowing perturbations of the estimator to directly control the perturbations of the value function.
However, although there has been some literature aiming to propose randomized algorithms for general function classes~\cite{ishfaq2021randomized, agarwal2022model, agarwal2022non, zhang2022feel}, these methods do not discuss how to define the posterior distribution supported by the given function class and how to draw the optimistic sample from the posterior~\cite{agarwal2022model, agarwal2022non, zhang2022feel}, or they require stronger assumptions on stochastic optimism~\cite{ishfaq2021randomized}, which is one of the most challenging elements in frequentist regret analysis.
Thus, the design of a tractable randomized exploration RL algorithm and the feasibility of frequentist regret analysis for randomized exploration remain open challenges.
Hence, the following question arises:

\textit{Can we design a provably efficient and tractable randomized algorithm for RL with MNL function approximation?}

We answer the above question by proposing the first randomized algorithm, $\texttt{RRL-MNL}$, achieving $\tilde{\Ocal}(\kappa^{-1} d^{\frac{3}{2}} H^{\frac{3}{2}} \sqrt{T})$ frequentist regret with constant-time computational cost per episode. 
$\texttt{RRL-MNL}$ is not only the first algorithm with randomized exploration for MNL-MDPs, but also, to the best of our knowledge, it provides the first frequentist regret analysis for a \textit{non-linear model-based} algorithm with randomized exploration without assuming stochastic optimism~\cite{ishfaq2021randomized}.

While $\texttt{RRL-MNL}$ is \textit{statistically} efficient, the current method used to analyze the regret of MNL function approximation introduces a problem-dependent constant $\kappa$ (Assumption~\ref{assm:positive kappa}), which reflects the level of non-linearity of the MNL transition model.
This constant $\kappa$ originates from the use of generalized linear models (GLMs) for contextual bandit settings~\citep{filippi2010parametric, li2017provably, jun2017scalable} and MNL bandit settings~\citep{oh2019thompson, chen2020dynamic, oh2021multinomial}. 
The magnitude of the constant $\kappa$ can be exponentially small with respect to the size of the decision set, hence the regret bound scaling with $\kappa^{-1}$ could be prohibitively large in the worst case~\citep{faury2020improved}.
Worse yet, the situation is even more challenging in RL, as in the worst case, $\kappa^{-1}$ can be much larger than in the case of bandits.
To overcome the prohibitive dependence on $\kappa$, algorithms based on new Bernstein-like inequalities and the self-concordant-like property of the log-loss have been proposed for logistic bandits~\citep{faury2020improved, abeille2021instance, faury2022jointly} and for MNL bandits~\citep{perivier2022dynamic, agrawal2023tractable, lee2024nearly}. 
As an extension of these works, the following fundamental question remains open:

\textit{Is it possible for RL algorithms with MNL function approximation to have a sharper dependence on the problem-dependent constant $\kappa$?}

For the above question, we propose the second randomized algorithm referred to as $\texttt{ORRL-MNL}$, which establishes a regret bound of $\tilde{\Ocal}(d^{\frac{3}{2}} H^{\frac{3}{2}} \sqrt{T} + \kappa^{-1}d^2 H^2)$ with constant-time computational cost per episode. 
We summarize our main contributions as follows:

\begin{itemize}
    \item 
    We propose computationally tractable randomized algorithms for RL with MNL function approximation: $\texttt{RRL-MNL}$ and $\texttt{ORRL-MNL}$.
    To the best of our knowledge, these are the first randomized model-based RL algorithms with MNL function approximation that achieve the frequentist regret bounds with constant-time computational cost per episode.

    \item 
    We establish that $\texttt{RRL-MNL}$ enjoys $\tilde{\Ocal}(\kappa^{-1} d^{\frac{3}{2}} H^{\frac{3}{2}} \sqrt{T})$ frequentist regret bound with constant-time computational cost per episode, where $d$ is the dimension of the transition core, $H$ is horizon length, $T$ is the total number of rounds, and $\kappa$ is a problem-dependent constant.
    We derive the stochastic optimism of $\texttt{RRL-MNL}$, and to our knowledge, this is the first frequentist regret analysis for a non-linear model-based algorithm with randomized exploration without assuming stochastic optimism.

    \item 
    To achieve a regret bound with improved dependence on $\kappa$, we introduce $\texttt{ORRL-MNL}$, which constructs the optimistic randomized value functions
    by taking into account the effects of the local gradient information for the MNL transition model at each reachable state.
    We prove that $\texttt{ORRL-MNL}$ enjoys an $\tilde{\Ocal}(d^{\frac{3}{2}} H^{\frac{3}{2}} \sqrt{T} + \kappa^{-1}d^2 H^2)$ regret with constant-time computational cost per episode, significantly improving the regret of $\texttt{RRL-MNL}$ without requiring prior knowledge of $\kappa$.

    \item
    We evaluate our algorithms on tabular MDPs and demonstrate the superior performance of our proposed algorithms compared to the existing state-of-the-art MNL-MDP algorithm~\cite{hwang2023model}. 
    The experiments provide evidence that our proposed algorithms are both computationally and statistically efficient.
\end{itemize}
Related works on RL with function approximation and MNL contextual bandits are provided in Appendix~\ref{appx:related work}.

\section{Problem Setting}
We consider the episodic \textit{Markov decision processes} (MDPs) denoted by $\Mcal (\Scal, \Acal, H, \{P\}_{h=1}^H, r) $, where $\Scal$ is the state space, $\Acal$ is the action space, $H$ is the horizon length of each episode, $\{P\}_{h=1}^H$ is the collection of probability distributions, and $r$ is the reward function. Every episodes start from the initial state $s_1$ and for every step $h \in [H]:= \{1,..., H\}$ in an episode, the learning agent interacts with the environment represented as $\Mcal$. The agent observes the state $s_h \in \Scal$, chooses an action $a_h \in \Acal$, receives a reward $r (s_h, a_h) \in [0, 1]$ and the next state $s_{h+1}$ is given by the transition probability distribution $P_h(\cdot | s_h, a_h)$. Then this process is repeated throughout the episode. A policy $\pi : \Scal \times [H] \rightarrow \Acal$ is a function that determines the action of the agent at state $s_h$, i.e., $a_h = \pi(s_h, h):= \pi_h(s_h)$.

We define the value function of the policy $\pi$, denoted by $V_h^\pi(s)$, as the expected sum of rewards under the policy $\pi$ until the end of the episode starting from $s_h = s$, i.e., $V_h^\pi (s) = \mathbb{E}_\pi \left[ \displaystyle \sum_{h' = h}^H r(s_{h'}, \pi_{h'} (s_{h'})) \mid s_h = s \right]$.
Similarly, we define the action-value function $Q_h^\pi (s, a) = r(s, a) + \mathbb{E}_{s' \sim P_h(\cdot \mid s,a)} \left[ V^\pi_{h+1}(s') \right]$.
We define an optimal policy $\pi^*$ to be a policy that achieves the highest possible value at every $(s, h) \in \Scal \times [H]$. We denote the optimal value function by $V_h^*(s) = V_h^{{\pi}^*}(s)$ and the optimal action-value function by $Q_h^*(s, a) = Q_h^{\pi^*} (s, a)$.
To simplify, we introduce the notation $P_h V_{h+1} (s, a) = \EE_{s' \sim P_h (\cdot | s, a)} [ V_{h+1} (s') ]$. Recall that the Bellman equations are,
\begin{equation*}        
    Q_h^\pi(s,a) = r(s,a) + P_h V_{h+1}^\pi (s,a) \, ,
    \quad 
    Q_h^*(s,a) = r(s,a) + P_h V_{h+1}^* (s,a) \, ,
\end{equation*}
where $V_{H+1}^\pi(s) = V_{H+1}^*(s) = 0$ and $V_h^*(s) = \max_{a \in \Acal} Q_h^* (s,a)$ for all $s \in \Scal $.

The goal of the agent is to maximize the sum of rewards for K episodes. In other words, the goal is to minimize the cumulative regret of the policy $\pi$ over K episodes where $\pi = \{ \pi^k \}_{k=1}^{K}$ is a collection of policies $\pi^k$ at k-th episode. The regret is defined as 
\begin{equation*}
    \Regret_\pi (K) := \sum_{k=1}^K (V_1^* - V_1^{\pi^k})(s^k_1)
\end{equation*}
where $s^k_1$ is the initial state at the $k$-th episode.

\subsection{Multinomial Logistic Markov Decision Processes (MNL-MDPs)}
Even though a lot of provable RL algorithms for linear MDPs are proposed, there is a simple but fundamental problem with the linear transition model assumption on the linear MDPs.
In other words, the output of a linear function approximating the transition model must be in $[0,1]$ and the probability of all possible following states must sum to $1$ exactly.
Such restrictive assumption can affect the regret performances of algorithm suggested under the linearity assumption.
To resolve these challenges, \citet{hwang2023model} propose a setting of a \textit{multinomial logistic Markov decision processes} (MNL-MDPs), where the state transition model is given by a multinomial logistic model.
We introduce the formal definition for MNL-MDP as follows:
\begin{assumption}[MNL-MDPs~\cite{hwang2023model}] \label{assm:mnl-mdp}
    An MDP $\Mcal(\Scal, \Acal, H, \{P_h\}_{h=1}^H, r)$ is an \textnormal{MNL-MDP} with a feature map $\varphib: \Scal \times \Acal \times \Scal \rightarrow \RR^d$, if for each $h \in [H]$, there exists $\thetab^*_h \in \RR^{d}$, such that for any $(s,a) \in \Scal \times \Acal$ and $s' \in \Scal_{s,a}:= \{ s' \in \Scal : \PP(s' \mid s, a) \ne 0 \}$, the state transition kernel of $s'$ when an action $a$ is taken at a state $s$ is given by,
    \begin{equation}
        P_h (s' \mid s,a)
        = \frac{\exp(\varphib(s,a,s')^\top \thetab^*_h)}{\sum_{\tilde{s} \in \Scal_{s,a}} \exp(\varphib(s,a,\tilde{s})^\top \thetab^*_h)} \, .
        \label{eq:MNL_original}
    \end{equation}    
    We call each unknown vector $\thetab^*_h$ transition core.
    Furthermore, we denote the maximum cardinality of the set of reachable states as $\Ucal$, i.e., $\Ucal := \max_{s,a} |\Scal_{s,a} |$.
\end{assumption}

\begin{remark}
    While \citet{hwang2023model} assume a homogeneous transition kernel, we assume an inhomogeneous transition kernel, in which the probability varies depending on the current time step $h$ even for the same state transition, which is a more general setting.
    Also, for notational simplicity, we denote the true transition kernel $P_h$ as $P_{\thetab^*_h}$, and the estimated transition kernel by $\thetab$ as $P_{\thetab}$.
\end{remark}

\subsection{Assumptions} \label{sec:assumptions}
We introduce some standard regularity assumptions. 
\begin{assumption}[Boundedness] \label{assm:bdd feature & param}
    We assume $ \| \varphib(s,a,s') \|_2 \le \Fnorm$ for all $(s,a, s') \in \Scal \times \Acal \times \Scal_{s,a}$, and $\| \thetab^*_h \|_2 \le \Pnorm$ for all $h \in [H]$.
\end{assumption}

\begin{assumption}[Known reward] \label{assm:known reward}
    We assume that the reward function $r$ is known to the agent.    
\end{assumption}

\begin{assumption}[Problem-dependent constant]\label{assm:positive kappa}
    Let $\Bcal_d(\Pnorm):= \{ \thetab \in \RR^d : \| \thetab \|_2 \le \Pnorm \}$. 
    There exists $\kappa > 0$ such that for any $(s,a) \in \Scal \times \Acal$ and $s', \tilde{s} \in \Scal_{s,a}$ with $s' \ne \tilde{s}$,
    \begin{equation*}
        \inf_{\thetab \in \Bcal_d(\Pnorm)} P_{\thetab} (s' \mid s, a) P_{\thetab} ( \tilde{s} \mid s, a) \ge \kappa \, .    
    \end{equation*}
\end{assumption}

\paragraph{Discussion of assumptions}
Assumption~\ref{assm:bdd feature & param} is common in the literature on RL with function approximation~\citep{jin2020provably, yang2020reinforcement, zanette2020frequentist, ishfaq2021randomized, hwang2023model} to make the regret bounds scale-free.
Assumption~\ref{assm:known reward} is used to focus on the main challenge of model-based RL that learning about $P$ of the environment is more difficult than learning $r$. 
In the model-based RL literature~\citep{yang2019sample, ayoub2020model, yang2020reinforcement, zhou2021nearly, hwang2023model}, the known reward $r$ assumption is widely used.
Assumption~\ref{assm:positive kappa} is typical in generalized linear contextual bandit~\cite{filippi2010parametric, li2017provably, faury2020improved, abeille2021instance, faury2022jointly} and MNL contextual bandit literature~\citep{oh2019thompson, amani2021ucb, oh2021multinomial, perivier2022dynamic, agrawal2023tractable, zhang2023online, lee2024nearly} to guarantee non-singular Fisher information matrix.

\section{Randomized Algorithm for MNL-MDPs having constant-time computational cost} \label{sec:rrl-mnl}

Previous work for MNL-MDPs~\cite{hwang2023model} proposed a UCB-based exploration algorithm. 
Constructing a UCB-based optimistic value function is not only computationally intractable but also tends to overly optimistically estimate the true optimal value function.
Additionally, their algorithm incurs increasing computation costs as episodes progress, as it requires all samples from the previous episode to estimate the transition core.
In this section, we present a novel model-based RL algorithm that incorporates \textit{randomized exploration} and \textit{online parameter estimation} for MNL-MDPs.

\subsection{Algorithm: \texorpdfstring{$\texttt{RRL-MNL}$}{RRL-MNL}}

\begin{algorithm}[t!] 
    \caption{$\texttt{RRL-MNL}$ (Randomized RL for MNL-MDPs)} \label{alg:Algorithm 1}
    \begin{algorithmic}[1]
        \STATE \textbf{Inputs:} Episodic MDP $\Mcal$, Feature map $\varphib:\Scal \times \Acal \times \Scal \rightarrow \RR^{d}$, Number of episodes~$K$, Regularization parameter $\lambda$, Exploration variance $\{ \sigma_k \}_{k=1}^K$, Sample size $M$, Problem-dependent constant $\kappa$
        \STATE \textbf{Initialize:} $\thetab^1_h = \zero_d$, $\Ab_{1,h} = \lambda \Ib_d$ for $h \in [H]$
        \FOR{episode $k=1,2, \cdots, K$}
            \STATE Observe $s^k_1$ and sample \textit{i.i.d.} noise vector $\xib^{(m)}_{k,h} \sim \Ncal(\zero_d, \sigma_k^2 \Ab_{k,h}^{-1})$ for $m \in [M]$ and $h \in [H]$
            \STATE Set $\left\{ Q^k_h (\cdot, \cdot) \right\}_{h \in [H]}$ as described in~\eqref{eq:stochastically q f.t for alg 1}
            \FOR{horizon $h=1, 2, \cdots, H$} 
                \STATE Select $a^k_h = \argmax_{a \in \Acal} Q^k_h(s^k_h, a)$ and observe $s^k_{h+1}$
                \STATE Update $\Ab_{k+1,h} = \Ab_{k,h} + \frac{\kappa}{2} \sum_{s' \in \Scal_{k,h}} \!\! \varphib(s^k_h, a^k_h, s') \varphib(s^k_h, a^k_h, s')^\top$ and $\thetab^{k+1}_h$ as in~\eqref{eq:online transition core estimation}
            \ENDFOR
        \ENDFOR
    \end{algorithmic}
\end{algorithm}

\paragraph{Online transition core estimation}
While \citet{hwang2023model} estimate the transition core using maximum likelihood estimation over all samples from previous episodes, we employ an efficient online parameter estimation method by exploiting the particular structure of the MNL transition model. 
The key insight is that the negative log-likelihood function for the MNL model in each episode is strongly convex over a bounded domain. 
This property allows us to utilize a variation of the online Newton step~\cite{hazan2007logarithmic, hazan2014logistic}, which inspired online algorithms for logistic bandits~\cite{zhang2016online} and MNL contextual bandits~\cite{oh2021multinomial}.
Specifically, for $(k,h) \in [K] \times [H]$, we define the response variable $y^k_h = \left[ y^k_h (s') \right]_{s' \in \Scal_{k,h}}$ such that $y^k_h (s') = \ind (s^k_{h+1} = s')$ for $s' \in \Scal_{k,h} := \Scal_{s^k_h, a^k_h}$. 
Then, $y^k_h$ is sampled from the following multinomial distribution:
$\displaystyle y^k_h \sim \mathrm{multinomial}( 1, \left[ P_{\thetab^*_h}(s' \mid s^k_h, a^k_h) \right]_{s' \in \Scal_{k,h}})$,
where $1$ represents that $y^k_h$ is a single-trial sample.
We define the per-episode loss $\ell_{k,h}(\thetab)$ as follows:
\begin{equation*}
    \ell_{k,h} (\thetab) := - \sum_{s' \in \Scal_{k,h}} y^k_h(s') \log P_{\thetab}(s' \mid s^k_h, a^k_h) \, .
\end{equation*}
Then, the estimated transition core for $\thetab^*_h$ is given by
\begin{equation} \label{eq:online transition core estimation}
    \thetab^k_h = \argmin_{\thetab \in \Bcal_d (\Pnorm)} \frac{1}{2} \| \thetab - \thetab^{k-1}_{h} \|_{\Ab_{k,h}}^2
        + (\thetab - \thetab^{k-1}_h)^\top \nabla \ell_{k-1,h}(\thetab^{k-1}_h) \, ,    
\end{equation}
where $\thetab^1_h$ can be initialized as any point in $\Bcal_d(\Pnorm)$ and $\Ab_{k,h}$ is the Gram matrix defined by
\begin{equation} \label{eq:global gram matrix}
    \Ab_{k,h} := \lambda \Ib_d + \frac{\kappa}{2} \sum_{i=1}^{k-1} \sum_{s' \in \Scal_{i, h}} \varphib(s^{i}_h, a^{i}_h, s') \varphib(s^{i}_h, a^{i}_h, s')^\top \, .
\end{equation}

\paragraph{Stochastically optimistic value function}
First of all, we introduce the key challenges of regret analysis for randomized algorithms, explain how previous works have overcome these challenges, and then describe why the techniques from previous works cannot be applied to MNL-MDPs.
Ensuring that the estimated value function is optimistic with sufficient frequency is a crucial challenge in analyzing the frequentist regret of randomized algorithms.
A common way to promote sufficient exploration in randomized algorithms is by perturbing the estimated value function or by performing posterior sampling in the transition model class.
Frequentist regret analysis of randomized exploration in an RL setting has been conducted for tabular~\cite{osband2016generalization, agrawal2017posterior, russo2019worst, pacchiano2021towards, tiapkin2022optimistic}, linear MDPs~\cite{zanette2020frequentist,ishfaq2021randomized}, and general function classes~\cite{ishfaq2021randomized, agarwal2022model, agarwal2022non, zhang2022feel}.
In the case of linear MDPs~\cite{zanette2020frequentist,ishfaq2021randomized}, since the property that the action-value function is linear in the feature map allows perturbing the estimated parameter directly to control the perturbation of the estimated value function.
Also, even though~\citet{ishfaq2021randomized} presented a randomized algorithm for the general function class using eluder dimension, they assume stochastic optimism (anti-concentration), which is in fact one of the most challenging aspects of frequentist analysis.
Other posterior sampling algorithms in RL for the general function class such as~\cite{agarwal2022model, agarwal2022non, zhang2022feel}, except for very limited examples, do not discuss how to define the posterior distribution supported by the given function class and how to draw the optimistic sample from the posterior.
That is why even after there exists a so-called \textit{general function class}-based result, it is often the case that results in specific parametric models are still needed.

Note that in episodic RL, the perturbed estimated value functions are propagated back through horizontal steps, requiring careful adjustment of the perturbation scheme to maintain a sufficient probability of optimism without decaying too quickly with the horizon.
For example, if the probability of the estimated value function being optimistic at horizon $h$ is denoted as $p$, this would result in the probability that the estimated value function in the initial state is optimistic being on the order of $p^H$, implying that the regret can increase exponentially with the length of the horizon $H$.
Additionally, the non-linearity and substitution effect of the next state transition in the MNL-MDPs make applying the existing TS techniques infeasible to guarantee optimism in MNL-MDPs with sufficient frequency.
Instead, we design the \textit{stochastically optimistic value function} by exploiting the structure of the MNL transition model. 
In other words, the prediction error of MNL transition model (Definition~\ref{def:prediction error & bellman error}) can be bounded by the weighted norm of the dominant feature $\DomFeat$ (Lemma~\ref{lemma:bound of prediction error}).
Based on such dominant feature, we perturb the estimated value function by injecting Gaussian noise whose variance is proportional to the inverse of the Gram matrix to encourage the perturbation with higher variance in less explored directions.
To guarantee the optimism with fixed probability, we adapt optimistic sampling technique~\citep{agrawal2017posterior, oh2019thompson, ishfaq2021randomized, hwang2023combinatorial}.
For each $m \in [M]$, sample \textit{i.i.d.} Gaussian noise vector $\xib^{(m)}_{k,h} \sim \Ncal(\zero_d, \sigma_k^2 \Ab_{k,h}^{-1})$ where $\sigma_k$ is an exploration parameter, and add the most optimistic inner product value $\max_{m \in [M]} \DomFeat_{k,h}(s,a)^\top \xib^{(m)}_{k,h}$ to the estimated value function.
To summarize for any $(s,a) \in \Scal \times \Acal$, set $Q^k_{H+1}(s,a) = 0$ and for $h \in [H]$, 
\begin{align} \label{eq:stochastically q f.t for alg 1}
    Q^k_h (s,a) =  \min \bigg\{ r(s,a) + \sum_{s' \in \Scal_{s,a}} P_{\thetab^k_h} (s' \mid s,a) V^k_{h+1}(s')
    + \max_{m \in [M]} \hat{\varphib}_{k,h}(s,a)^\top \xib_{k,h}^{(m)}, H \bigg\} \, ,       
\end{align}
where $V^k_h(s) = \max_{a'} Q^k_h(s,a')$ and 
$\DomFeat_{k,h}(s,a):= \varphib(s,a,\hat{s})$ for $\hat{s} = \argmax_{s' \in \Scal_{s,a}} \| \varphib(s,a,s') \|_{\Ab_{k,h}^{-1}}$.
Based on these stochastically optimistic value function, the agent plays a greedy action $a^k_h=\argmax_{a'} Q^k_h (s^k_h, a')$. 
We layout the procedure in Algorithm~\ref{alg:Algorithm 1}.

\begin{remark}
    Note that $\texttt{RRL-MNL}$ only requires constant-time computational cost and storage cost per episode, as it does not require storing all samples from previous episodes, and the Gram matrix $\Ab_{k,h}$ can be updated incrementally.
\end{remark}

\subsection{Regret bound of \texorpdfstring{$\texttt{RRL-MNL}$}{RRL-MNL}}
We present the regret upper bound of $\texttt{RRL-MNL}$. The complete proof is deferred to~\Cref{appx:regret bound of alg1}.

\begin{theorem}[Regret Bound of $\texttt{RRL-MNL}$] \label{thm:alg 1}
    Suppose that Assumption~\ref{assm:mnl-mdp}-~\ref{assm:positive kappa} hold.
    For any $0 < \delta < \frac{\Phi(-1)}{2}$, if we set the input parameters in Algorithm~\ref{alg:Algorithm 1} as $\lambda = \Fnorm^2, \sigma_k = \tilde{\Ocal}(H \sqrt{d})$ and $M = \lceil 1 - \frac{\log H}{\log \Phi(1)} \rceil$ where $\Phi$ is the normal CDF, then
    with probability at least $1 - \delta$, the cumulative regret of the $\textup{\texttt{RRL-MNL}}$ policy $\pi$ is upper-bounded as follows:
    \begin{equation*}
        \Regret_\pi (K)
        = \tilde{\Ocal} \left( \kappa^{-1} d^{\frac{3}{2}} H^{\frac{3}{2}} \sqrt{T} \right),
    \end{equation*}
    where $T=KH$ is the total number of steps.
\end{theorem}

\paragraph{Discussion of Theorem~\ref{thm:alg 1}}
To our best knowledge, this is the first result to provide a frequentist regret bound for the MNL-MDPs.
Among the previous RL algorithms using function approximation, the most comparable techniques to our method are \textit{model-free} algorithms with randomized exploration~\citep{zanette2020frequentist, ishfaq2021randomized}.
To guarantee stochastic optimism,~\citet{zanette2020frequentist} established a lower bound on the difference between the estimated value and the optimal value by the summation of linear terms with respect to the average feature (Lemma F.1 in~\cite{zanette2020frequentist}).
This property is achievable due to the linear expression of the value function in linear MDPs.
Instead, we established a lower bound on the difference between value functions by the summation of the Bellman errors (Definition~\ref{def:prediction error & bellman error}) along the sample path obtained through the optimal policy (Lemma~\ref{supporting lemma:pessimism decomposition}). 
Hence, our analysis significantly differs from that of~\citet{zanette2020frequentist} since the value function in MNL-MDPs is no longer linearly parametrized, and there is no closed-form expression for it.

Compared to~\cite{ishfaq2021randomized}, they also used an optimistic sampling technique; however, our theoretical sampling size $M = \Ocal(\log H)$ is much tighter than that of~\cite{ishfaq2021randomized}, i.e., $\Ocal(d)$ for the linear function class, $\Ocal(\log (T |\Scal| |\Acal|))$ for the general function class.
While~\citet{ishfaq2021randomized} extend the results of the linear function class to general function class under the assumption of stochastic optimism (Assumption C in~\cite{ishfaq2021randomized}), 
we provide the frequentist regret analysis for a \textit{non-linear model-based} algorithm with randomized exploration \textit{without assuming stochastic optimism}.

Compared to the optimistic exploration algorithm for MNL-MDPs~\cite{hwang2023model}, our randomized exploration requires a more involved proof technique to ensure that the perturbation of the estimated value function has enough variance to maintain optimism with sufficient frequency (Lemma~\ref{lemma:stochastic optimism}).
As a result, the established regret of $\texttt{RRL-MNL}$ differs by a factor of $\sqrt{d}$, which aligns with the difference in the existing bounds of linear bandits between a TS-based algorithm~\citep{abeille2017Linear} and a UCB-based algorithm~\citep{abbasi2011improved}.
Additionally, we achieve statistical efficiency for the \textit{inhomogeneous transition model}, which is a more general setting than that of~\citet{hwang2023model}.
Our computation cost per episode is $\Ocal(1)$ while the computation cost per episode of~\citet{hwang2023model} is $\Ocal(K)$. 

\paragraph{Proof Sketch of Theorem~\ref{thm:alg 1}}
We provide the proof sketch of Theorem~\ref{thm:alg 1}. 
By decomposing the regret into the estimation part and the pessimism part, we have
\begin{equation*}
    \sum_{k=1}^K (V^*_1 - V^{\pi_k}_1)(s^k_1)
    = \sum_{k=1}^K \Big( \underbrace{V^*_1 - V^k_1}_\text{Pessimism} + \underbrace{V^k_1 - V^{\pi_k}_1}_\text{Estimation} \Big)  (s^k_1) \, .
\end{equation*}
We bound these two parts separately.
For the estimation part, for each $k \in [K], h \in [H]$, we first show that the online estimated transition core $\thetab^k_h$~\eqref{eq:online transition core estimation} concentrates around the unknown transition core parameter $\thetab^*_h$ with high probability (Lemma~\ref{lemma:concentration of online estimated transition core}).
Then, we show that the prediction error induced by the estimated transition core can be bounded by the weighted norm of the dominant feature $\hat{\varphib}$, multiplied by the confidence radius of the estimated transition core (Lemma~\ref{lemma:bound of prediction error}).
The bounded prediction error, together with the concentration of Gaussian noise, implies the desired bound on the estimation part (Lemma~\ref{lemma:bound of estimation part}).
For the pessimism part, we first show that the stochastically optimistic value function $V^k_1$ is optimistic than the true optimal value function $V^*_1$ with sufficient frequency (Lemma~\ref{lemma:stochastic optimism}).
In the next step, we show that the pessimism part is upper bounded by a bound of the estimation part times the inverse probability of being optimistic (Lemma~\ref{lemma:bound of pessimism part}).
Combining all the results, we can conclude the proof.
Refer to Appendix~\ref{appx:regret bound of alg1} for detailed proofs.

\section{Statistically Improved Algorithm for MNL-MDPs}
Although $\texttt{RRL-MNL}$ is provably efficient and achieves constant-time computational cost per episode, the current analysis makes its regret bound scale with $\kappa^{-1}$.
Recall that the problem-dependent constant $\kappa$ introduced in Assumption~\ref{assm:positive kappa} indicates the curvature of the MNL function, i.e., how difficult it is to learn the true transition core parameter.
It is required to ensure the non-singular Fisher information matrix, hence is typically used in GLM or MNL bandit algorithms that use the maximum likelihood estimator.
As introduced in~\citet{faury2020improved}, $\kappa^{-1}$ can be exponentially large in the worst case.
The appearance of $\kappa$ in existing bounds originates in the connection between the difference of estimators and the difference of gradients of negative log-likelihood, usually denoted as $\Gb$ in~\citet{filippi2010parametric}. 
Without considering local information at all, using a loose lower bound for $\Gb$ incurs $\kappa^{-1}$ in regret bound (see  Section 4.1 in~\citet{agrawal2023tractable}).    
Recently, improved dependence on $\kappa$ has been achieved in bandit literature~\citep{faury2020improved, abeille2021instance, perivier2022dynamic, agrawal2023tractable, zhang2023online, lee2024nearly} through the use of generalization of the Bernstein-like tail inequality~\citep{faury2020improved} and the self-concordant-like property of the log loss~\citep{bach2010self}.
However, a direct adaptation of the MNL bandit technique would result in sub-optimal dependence on the assortment size in MNL bandit, which corresponds to the size of the set of reachable states, such as $\Ucal$.
In this section, we introduce a new randomized algorithm for MNL-MDPs, equipped with a tight online parameter estimation and feature centralization technique that achieves a regret bound with improved dependence on $\kappa$ and $\Ucal$.

\subsection{Algorithms: \texorpdfstring{$\texttt{ORRL-MNL}$}{ORRL-MNL}}
\begin{algorithm}[t!] 
    \caption{$\texttt{ORRL-MNL}$ (Optimistic Randomized RL for MNL-MDPs)} \label{alg:Algorithm 2}
    \begin{algorithmic}[1]
        \STATE \textbf{Inputs:} Episodic MDP $\Mcal$, Feature map $\varphib:\Scal \times \Acal \times \Scal \rightarrow \RR^{d}$, Number of episodes~$K$, Regularization parameter $\lambda$, Exploration variance $\{ \sigma_k \}_{k=1}^K$, Confidence radius $\{ \beta_k \}_{k=1}^K$, Sample size $M$, Step size $\eta$
        \STATE \textbf{Initialize: } $\omdtheta{1}{h} = \zero_d $, $\Bb_{1,h} = \lambda \Ib_d$ for all $h \in [H]$
        \FOR{episode $k=1,2, \cdots, K$}
            \STATE Observe $s^k_1$ and sample \textit{i.i.d.} noise vector $\xib^{(m)}_{k,h} \sim \Ncal(\zero_d, \sigma_k^2 \Bb_{k,h}^{-1})$ for $m \in [M]$ and $h \in [H]$
            \STATE Set $\left\{ \tilde{Q}^k_h (\cdot, \cdot) \right\}_{h \in [H]}$ as described in~\eqref{eq:q-function for alg 2}
            \FOR{horizon $h=1, 2, \cdots, H$} 
                \STATE Select $a^k_h = \argmax_{a \in \Acal} \tilde{Q}^k_h(s^k_h, a)$ and observe $s^k_{h+1}$
                \STATE Update $\tilde{\Bb}_{k,h} = \Bb_{k,h} + \eta \nabla^2 \ell_{k,h} (\omdtheta{k}{h})$ and $\omdtheta{k+1}{h}$ as in~\eqref{eq:online mirror descent theta}
                \STATE Update $\Bb_{k+1,h} = \Bb_{k,h} + \nabla^2 \ell_{k,h} (\omdtheta{k+1}{h})$
            \ENDFOR
        \ENDFOR
    \end{algorithmic}
\end{algorithm}

\paragraph{Tight online transition core estimation}
\citet{zhang2023online} presented a jointly efficient UCB-based MNL contextual bandit algorithm using online mirror descent algorithm.
Adapting the update rule from~\cite{zhang2023online}, the estimated transition core run by the online mirror descent is given by
\begin{equation} \label{eq:online mirror descent theta}
    \omdtheta{k+1}{h} = \argmin_{\thetab \in \Bcal_d(\Pnorm)} \frac{1}{2 \eta} \| \thetab - \omdtheta{k}{h} \|_{\tilde{\Bb}_{k,h}}^2 + \thetab^\top \nabla \ell_{k,h}(\omdtheta{k}{h}) \, ,
\end{equation}
where $\omdtheta{1}{h}$ can be initialized as any point in $\Bcal_d(\Pnorm)$, $\eta$ is a step size, and $\tilde{\Bb}_{k,h}$ is defined as
\begin{equation}
    \tilde{\Bb}_{k,h} := \Bb_{k,h} + \eta \nabla^2 \ell_{k,h} (\omdtheta{k}{h}) \, , \quad 
    \Bb_{k,h} := \lambda \Ib_d + \sum_{i=1}^{k-1} \nabla^2 \ell_{i,h} (\omdtheta{i+1}{h}) \, .
\end{equation}
Note that the MNL model in~\citet{zhang2023online} operates in a \textit{multiple-parameter} setting, where there are multiple unknown choice parameters and one given context feature.
In contrast, our MNL model operates in a \textit{single-parameter} setting, where there is one unknown transition core and features for up to $\Ucal$ reachable states.
This difference results in variations in applying the self-concordant-like property of the log-loss for the MNL model. For instance,~\citet{zhang2023online} utilized the fact that the log-loss for the multiple parameter MNL model is $\sqrt{6}$-self-concordant-like (Lemma 2 in \citet{zhang2023online}).
On the other hand,~\citet{lee2024nearly} revisit the self-concordant-like property and demonstrate that the log-loss of the single-parameter MNL model is $3 \sqrt{2}$-self-concordant-like (Proposition B.1 in \citet{lee2024nearly}).
This results in a concentration bound that is independent of $\kappa$ and $\Ucal$, introduced in Lemma~\ref{lemma:tight concentration}.
\begin{remark}
    Note that the online estimated parameters $\thetab^k_h$~\eqref{eq:online transition core estimation} and $\omdtheta{k}{h}$~\eqref{eq:online mirror descent theta} do not aim to minimize the sum of negative log-likelihoods, $\sum_{k'=1}^k \ell_{k',h}(\thetab)$. Instead, we show that the online estimated parameter concentrates around the unknown transition core $\thetab^*_h$ with high probability (Lemma~\ref{lemma:concentration of online estimated transition core} \&~\ref{lemma:tight concentration}).
    This online update approach allows us to estimate the transition core with constant-time computational cost per episode, as the agent does not need to store all samples from previous episodes.
\end{remark}

\paragraph{Optimistic randomized value function}
To achieve improved dependence on $\kappa$, a crucial point is to utilize the local gradient information of MNL transition probabilities for each reachable state when constructing the Gram matrix.
In MNL bandit problems~\cite{perivier2022dynamic, zhang2023online}, this can be accomplished by substituting the Hessian of the negative log-likelihood with the Gram matrix using global gradient information $\kappa$.
However, there are fundamental differences between the settings in~\citet{perivier2022dynamic, zhang2023online} and ours. 
~\citet{perivier2022dynamic} address the case where the reward for each product is \textit{uniform} (i.e., all products have a reward of 1), and the reward for not selecting a product from the given assortment (also known as the outside option) is 0.
On the other hand,~\citet{zhang2023online} deal with \textit{non-uniform} rewards where the reward for each product may vary; however, the rewards for individual products are known a priori to the agent.
In contrast, in MNL-MDPs, the value for each reachable state may vary (non-uniform) and is \textit{not known} beforehand. 
Due to these differences, the analysis techniques in MNL bandits~\cite{perivier2022dynamic, zhang2023online} cannot be directly applied to our setting.
Instead, we adapt the feature centralization technique~\cite{lee2024nearly}.
Then, the Hessian of the per-round loss $\ell_{k,h}(\thetab)$ is expressed in terms of the centralized feature as follows:
\begin{equation*}
    \nabla^2 \ell_{k,h} (\thetab) = \sum_{s' \in \Scal_{k,h}} P_{\thetab}(s' \mid s^k_h,a^k_h) \bar{\varphib}(s^k_h,a^k_h,s'; \thetab) \bar{\varphib}(s^k_h,a^k_h,s';\thetab)^\top \, .
\end{equation*}
where $\bar{\varphib}(s,a,s'; \thetab) := \varphib(s,a,s') - \EE_{\tilde{s} \sim P_{\thetab}(\cdot \mid s,a)}[\varphib(s, a, \tilde{s})]$ is the centralized feature by $\thetab$.
For more details, please refer to Appendix~\ref{appx:prediction error of alg 2}.

Now we introduce the \textit{optimistic randomized value function} $\tilde{Q}^k_h(\cdot, \cdot)$ for $\texttt{ORRL-MNL}$. 
The key point is that when perturbing the estimated value function, we use the centralized feature by the estimated transition parameter $\omdtheta{k}{h}$.
For any $(s,a) \in \Scal \times \Acal$, set $\tilde{Q}^k_{H+1}(s,a) = 0$ and for each $h \in [H]$, 
\begin{equation} \label{eq:q-function for alg 2}
    \tilde{Q}^k_h(s,a) := \min \bigg\{ r(s,a) + \sum_{s' \in \Scal_{s,a}} P_{\omdtheta{k}{h}}(s' \mid s,a) \tilde{V}^k_{h+1}(s') + \rbonus_{k,h}(s,a)\, , H \bigg\} \, ,
\end{equation}
where $\tilde{V}^k_{h}(s) := \max_{a \in \Acal} \tilde{Q}^k_h(s,a)$ and $\rbonus_{k,h}(s,a)$ is the \textit{randomized bonus term} defined by
\begin{equation*}
    \rbonus_{k,h}(s,a) := \sum_{s' \in \Scal_{s,a}} P_{\omdtheta{k}{h}} (s' \mid s, a) \bar{\varphib}(s, a, s'; \omdtheta{k}{h})^\top \xib^{s'}_{k,h}
    + 3 H \beta_k^2 \max_{s' \in \Scal_{s,a}} \| \varphib (s,a,s') \|^2_{\Bb^{-1}_{k,h}} \, .
\end{equation*}
Here we sample \textit{i.i.d.} Gaussian noise $\xib^{(m)}_{k,h} \sim \Ncal(\zero_d, \sigma_k^2 \Bb_{k,h}^{-1})$ for each $m \in [M]$ and set $\xib^{s'}_{k,h} := \xib^{m(s')}_{k,h}$ where $m(s') := \argmax_{m \in [M]} \bar{\varphib}(s, a, s'; \omdtheta{k}{h})^\top \xib^{m}_{k,h}$ is the most optimistic sampling index for a reachable state $s'$.
Based on these optimistic randomized value function, at each episode the agent plays a greedy action with respect to $\tilde{Q}^k_h$ as summarized in Algorithm~\ref{alg:Algorithm 2}.

\begin{remark}
    Note that the second term in the randomized bonus always has a positive value, but it rapidly decreases as episode proceeds.
    While due to the randomness of $\xib$, the randomized bonus $\rbonus_{k,h}$ itself cannot be guaranteed to always have a positive value. 
    Consequently, the constructed value function $\tilde{Q}^k_h(\cdot, \cdot)$ can be optimistic or pessimistic. 
    However, as shown in Lemma~\ref{lemma:stochastic optimism of alg2}, optimistic sampling technique ensures that the optimistic randomized value function $\tilde{Q}^k_h$ has at least a constant probability of being optimistic than the true optimal value function.
\end{remark}

\begin{remark}
    As with $\texttt{RRL-MNL}$, since the transition core is estimated in an online manner and the Gram matrices with local gradient information $\Bb_{k,h}$ and $\tilde{\Bb}_{k,h}$ are updated incrementally, $\texttt{ORRL-MNL}$ also requires constant-time computational cost and storage cost per-episode.
    Although $\texttt{ORRL-MNL}$ requires an additional $\Ocal(\Ucal)$ computation cost for feature centralization, the computation complexity order is the same as that of $\texttt{RRL-MNL}$ because it also needs to go over reachable states to calculate the dominant feature $\DomFeat$, which also incurs a $\Ocal(\Ucal)$ computation cost.
    On the other hand, $\texttt{ORRL-MNL}$ does not require prior knowledge of $\kappa$ and achieves a regret with a better dependence on $\kappa$.
\end{remark}

\subsection{Regret Bound of \texorpdfstring{$\texttt{ORRL-MNL}$}{ORRL-MNL}}
We present the regret upper bound of $\texttt{ORRL-MNL}$. The complete proof is deferred to Appendix~\ref{appx:alg2}.
\begin{theorem}[Regret Bound of $\texttt{ORRL-MNL}$] \label{thm:alg 2}
    Suppose that Assumption~\ref{assm:mnl-mdp}-~\ref{assm:positive kappa} hold.
    For any $0 < \delta < \frac{\Phi(-1)}{2}$, if we set the input parameters in Algorithm~\ref{alg:Algorithm 2} as $\lambda = \Ocal(\Fnorm^2 d \log \Ucal), \beta_k = \Ocal(\sqrt{d} \log \Ucal \log(kH)), \sigma_k = H \beta_k,$ $M=\lceil 1 - \frac{\log (H \Ucal)}{\log \Phi(1)} \rceil$, and $\eta = \Ocal(\log \Ucal)$, then with probability at least $1 - \delta$, the cumulative regret of the $\texttt{ORRL-MNL}$ policy $\pi$ is upper-bounded as follows: 
    \begin{align*}
        \Regret_\pi (K)
        = \BigOTilde \left( d^{3/2} H^{3/2} \sqrt{T} + \kappa^{-1} d^2 H^2 \right) \, ,
    \end{align*}    
    where $T=KH$ is the total number of time steps.
\end{theorem}

\paragraph{Dicussion of Theorem~\ref{thm:alg 2}}
Theorem~\ref{thm:alg 2} establishes that the leading term in the regret bound does not suffer from the problem-dependent constant $\kappa^{-1}$ and the second term of the regret bound is independent of the size of set of reachable states. 
To the extent of our knowledge, this is the first algorithm that provides a frequentist regret guarantee with improved dependence on $\kappa^{-1}$ in MNL-MDPs.
Compared to $\texttt{RRL-MNL}$, the technical challenge lies in ensuring the stochastic optimism of the estimated value for $\texttt{ORRL-MNL}$. 
Note that the prediction error (Definition~\ref{def:prediction error & bellman error}) for $\texttt{ORRL-MNL}$ is characterized by two components: one related to the gradient information of the MNL transition model at each reachable state, and the other related to the dominant feature with respect to the Gram matrix \(\Bb_{k,h}\) (Lemma~\ref{lemma:prediction error bound (alg2)}). 
Hence, the probability of the Bellman error at each horizon, when following the optimal policy, being negative can depend on the size of the reachable states. 
This implies that the probability of stochastic optimism can be exponentially small, not only in the horizon $H$ but also in the size of the reachable states $\Ucal$.
However, as shown in Lemma~\ref{lemma:stochastic optimism of alg2}, this challenge has been overcome by using a sample size $M$ that \textit{logarithmically} increases with $\Ucal$, effectively addressing the issue.

\paragraph{Proof Sketch of Theorem~\ref{thm:alg 2}}
The overall proof pipeline for Theorem~\ref{thm:alg 2} is similar to that of Theorem~\ref{thm:alg 1}.
The main differences lie in the concentration of the estimated transition core (Lemma~\ref{appx:prediction error of alg 2}), the bound on the prediction error (Lemma~\ref{appx:prediction error of alg 2}), and the stochastic optimism (Lemma~\ref{lemma:stochastic optimism of alg2}).
Please refer to Appendix~\ref{appx:alg2} for detailed proofs.

\paragraph{Optimistic exploration extension}
In general, since TS-based randomized exploration requires a more rigorous proof technique than UCB-based algorithms, our technical ingredients enable the use of optimistic exploration in a straightforward manner.
We introduce $\AlgUCB$ (Algorithm~\ref{alg:ucrl-mnl plus}) in the Appendix~\ref{appx:ucrl-mnl plus}, an optimism-based algorithm for MNL-MDPs. 
It is both \textit{computationally} and \textit{statistically} efficient compared to $\texttt{UCRL-MNL}$~\cite{hwang2023model}, achieving \textit{the tightest regret bound} for MNL-MDPs.

\begin{corollary} \label{coro:ucrl-mnl plus}
    $\AlgUCB$ (Algorithm~\ref{alg:ucrl-mnl plus}) has $\tilde{\Ocal}(d H^{3/2} \sqrt{T} + \kappa^{-1} d^2 H^2)$ regret with high probability.
\end{corollary}

\section{Numerical Experiments} \label{sec:numerical experiments}
\begin{figure}[t!]
    \centering
    \begin{subfigure}[b]{0.32\textwidth} 
        \includegraphics[width=\textwidth, trim=0mm 0mm 0mm 0mm, clip]{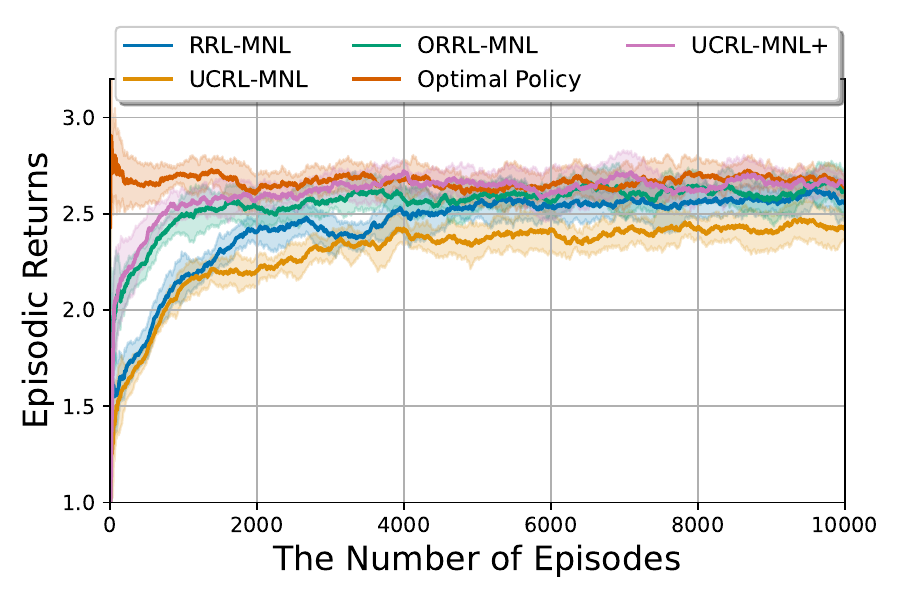}
        \vspace{-5mm}
        \caption{$S=4, H=12$}
        \label{fig:exp_instance_1}
    \end{subfigure}
    \begin{subfigure}[b]{0.32\textwidth}
        \includegraphics[width=\textwidth, trim=0mm 0mm 0mm 0mm, clip]{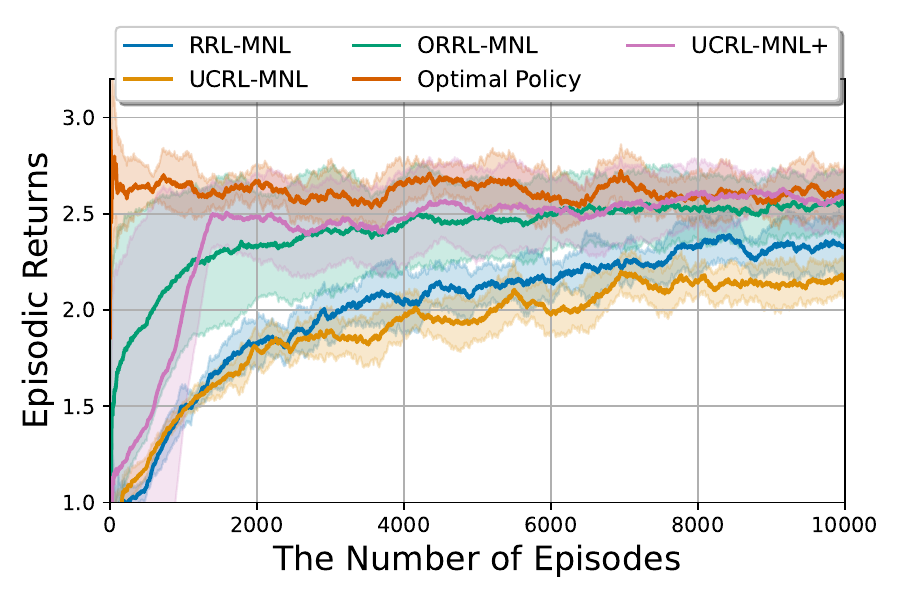}
        \vspace{-5mm}
        \caption{$S=8, H=24$}
        \label{fig:exp_instance_2}
    \end{subfigure}
    \begin{subfigure}[b]{0.305\textwidth}
        \includegraphics[width=\textwidth, trim=0mm -4mm 0mm 0mm, clip]{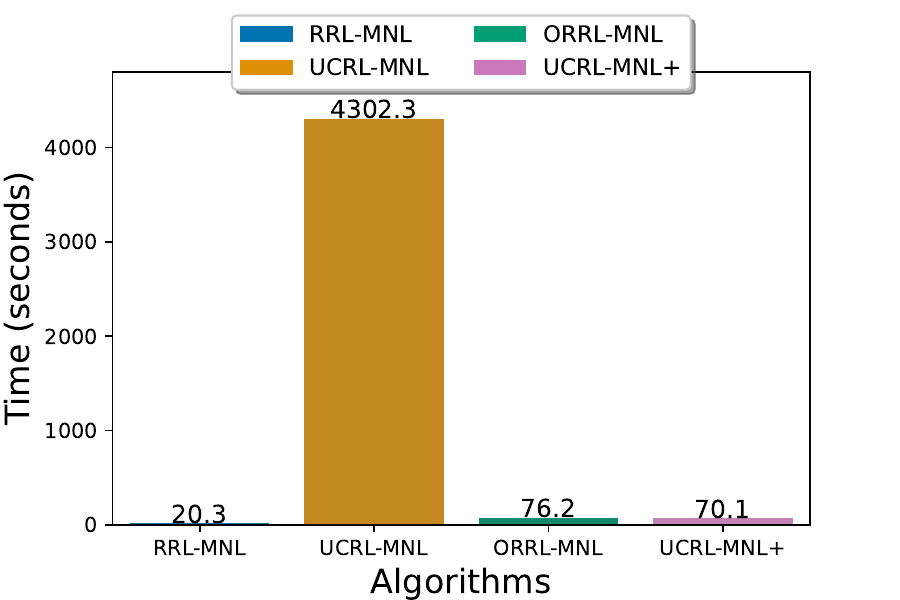}
        \vspace{-5mm}
        \caption{Runtime for 1,000 episodes}
        \label{fig:exp_barchart}
    \end{subfigure}
    \caption{Riverswim experiment results}
    \label{fig:riverswim}
    \vspace{-2.5mm}
\end{figure}

We perform a numerical evaluation on a variant of RiverSwim~\citep{osband2013more} to demonstrate practicality of our proposed algorithms.
We compare our algorithms ($\texttt{RRL-MNL}, \texttt{ORRL-MNL}, \AlgUCB$) with the state-of-the-art $\texttt{UCRL-MNL}$~\cite{hwang2023model} for MNL-MDPs.
For each configuration, we report the averaged results over 10 independent runs.
Figure~\ref{fig:exp_instance_1} and~\ref{fig:exp_instance_2} show the episodic return of each algorithm, which is the sum of all the rewards obtained in one episode. 
First, our proposed algorithms ($\texttt{RRL-MNL}$, $\texttt{ORRL-MNL}$, $\AlgUCB$) outperform $\texttt{UCRL-MNL}$~\citep{hwang2023model} for both cases of $|\Scal|=4, 8$.
Second, $\texttt{ORRL-MNL}$ and $\AlgUCB$ reach the optimal values quickly compared to the other algorithms, demonstrating improved statistical efficiency. 
Figure~\ref{fig:exp_barchart} illustrates the comparison in running time of the algorithms for the first 1,000 episodes.
Our proposed algorithms are at least 50 times faster than $\texttt{UCRL-MNL}$.
These differences become more pronounced as the episodes progress because our algorithms have a constant computation cost, whereas the computation cost of $\texttt{UCRL-MNL}$ increases over time.

\section{Conclusions}
We propose randomized algorithms with provable efficiency and constant-time computational cost for MNL-MDPs.
For the first algorithm, $\texttt{RRL-MNL}$, we use an optimistic sampling technique to ensure the stochastic optimism of the estimated value functions and provide the frequentist regret analysis.
This is the first frequentist regret analysis for a non-linear model-based algorithm with randomized exploration without assuming stochastic optimism.
To achieve a statistically improved regret bound, we propose $\texttt{ORRL-MNL}$ by constructing the optimistic randomized value function using the effects of the local gradient of the MNL transition model equipped with the centralized feature.
As a result, we achieve a frequentist regret guarantee with improved dependence on $\kappa$ in RL with the MNL transition model, which is a significant contribution.
The effectiveness and practicality of our methods are supported by numerical experiments.

\clearpage

\section*{Acknowledgements}
We sincerely thank the anonymous reviewers for their constructive feedback.
This work was supported by the National Research Foundation of Korea(NRF) grant funded by the Korea government(MSIT) (No. 2022R1C1C1006859, 2022R1A4A1030579, and RS-2023-00222663) and by AI-Bio Research Grant through Seoul National University.

\bibliography{references.bib}
\bibliographystyle{plainnat}

\clearpage
\appendix

\renewcommand{\contentsname}{Contents of Appendix}
\addtocontents{toc}{\protect\setcounter{tocdepth}{2}}
{
  \hypersetup{hidelinks}
  \tableofcontents
}

\section{Related Work} \label{appx:related work}

\begin{table*}[ht]
    \centering
    \caption{This table compares the problem settings, online update, performance of the this paper with those of other methods in provable RL with function approximation. For computation cost, we only keep the dependence on the number of episode $K$.}
    \label{tab:my-table}
    \resizebox{\textwidth}{!}{%
    \begin{tabular}{cccccc}
        \hline
        Algorithm                               & Model-based                & Transition model & Reward & Computation cost              & Regret                      
        \\
        \hline
        \texttt{LSVI-UCB}~\cite{jin2020provably} & \ding{55} & Linear           & Linear & $\Ocal(K)$ & $\tilde{\Ocal}(d^{\frac{3}{2}} H^{\frac{3}{2}} \sqrt{T})$
        \\
        \texttt{OPT-RLSVI}~\cite{zanette2020frequentist} & \ding{55} & Linear           & Linear & $\Ocal(K)$ & $\tilde{\Ocal}(d^2 H^2 \sqrt{T})$
        \\
        \texttt{LSVI-PHE}~\cite{ishfaq2021randomized} & \ding{55} & Linear           & Linear & $\Ocal(K)$ & $\tilde{\Ocal}(d^{\frac{3}{2}} H^{\frac{3}{2}} \sqrt{T})$
        \\
        \texttt{UC-MatrixRL}~\cite{yang2020reinforcement} & \ding{51} & Linear           & Known  & $\Ocal(K)$ & $\tilde{\Ocal}(d^{\frac{3}{2}} H^2 \sqrt{T})$
        \\
        \texttt{UCRL-VTR}~\cite{ayoub2020model} & \ding{51} & Linear mixture   & Known  & $\Ocal(K)$ & $\tilde{\Ocal}(d H^{\frac{3}{2}}\sqrt{T})$
        \\
        \texttt{UCRL-MNL}~\cite{hwang2023model} & \ding{51} & MNL              & Known  & $\Ocal(K)$ & $\tilde{\Ocal}(\kappa^{-1} d H^{\frac{3}{2}} \sqrt{T})$
        \\
        $\texttt{RRL-MNL}$ (\textbf{this work}) & \ding{51} & MNL              & Known  & $\Ocal(1)$ & $\tilde{\Ocal}(\kappa^{-1} d^{\frac{3}{2}} H^{\frac{3}{2}} \sqrt{T})$
        \\
        $\texttt{ORRL-MNL}$ (\textbf{this work}) & \ding{51} & MNL              & Known  & $\Ocal(1)$ & $\BigOTilde \left( d^{\frac{3}{2}} H^{\frac{3}{2}} \sqrt{T} + \kappa^{-1} d^2 H^2 \right)$ 
        \\
        $\AlgUCB$ (\textbf{this work}) & \ding{51} & MNL              & Known  & $\Ocal(1)$ & $\BigOTilde \left( d H^{\frac{3}{2}} \sqrt{T} + \kappa^{-1} d^2 H^2 \right)$ 
        \\        
        \hline
    \end{tabular}%
    }
\end{table*}

\paragraph{RL with linear function approximation}
There has been a growing interest in studies that extend beyond tabular MDPs and focus on function approximation methods with provable guarantees~\citep{jiang2017contextual, yang2019sample, jin2020provably, zanette2020frequentist, modi2020sample, du2020is, cai2020provably, ayoub2020model, wang2020reinforcement_eluder, weisz2021exponential, he2021logarithmic, zhou2021nearly, zhou2021provably, ishfaq2021randomized, hwang2023model, ishfaq2024provable}.
In particular, for minimizing regret in linear MDPs,~\citet{jin2020provably} propose an optimistic variant of the Least-Squares Value Iteration (LSVI) algorithm~\citep{bradtke1996linear, osband2016generalization} under the assumption that the transition model and reward function of the MDPs are linear function of a $d$-dimensional feature mapping and they guarantee $\tilde{\Ocal}(d^{\frac{3}{2}}H^{\frac{3}{2}}\sqrt{T})$ regret. 
~\citet{zanette2020frequentist} propose a randomized LSVI algorithm that incorporates exploration by perturbing the least-square approximation of the action-value function, and this algorithm guarantees $\tilde{\Ocal}(d^2 H^2 \sqrt{T})$ regret.
~\citet{ishfaq2021randomized} propose a variant of the randomized LSVI algorithm that combines optimism and TS by perturbing the training data with \textit{i.i.d.} scalar noise, achieving a regret bound of \(\tilde{\mathcal{O}}(d^{\frac{3}{2}} H^{\frac{3}{2}} \sqrt{T})\).  
Similarly, ~\citet{ishfaq2024provable} introduce a randomized RL algorithm that employs Langevin Monte Carlo (LMC) to approximate the posterior distribution of the action-value function, also ensuring a regret bound of \(\tilde{\mathcal{O}}(d^{\frac{3}{2}} H^{\frac{3}{2}} \sqrt{T})\).  
Also, there have been studies on model-based methods with function approximation in linear MDPs, such as~\citet{yang2020reinforcement}, which assume that the transition probability kernel is a bilinear model parametrized by a matrix and propose a UCB-based algorithm with an upper bound of $\tilde{\Ocal}(d^{\frac{3}{2}} H^2 \sqrt{T})$ for regret. 
~\citet{he2023nearly} propose an algorithm achieving nearly minimax optimal regret $\tilde{\Ocal}(d H \sqrt{T})$.
~\citet{jia2020model} consider a specific type of MDPs called linear mixture MDPs in which the transition probability kernel is a linear combination of different basis kernels.
This model encompasses various types of MDPs studied previously in~\citet{modi2020sample, yang2020reinforcement}.
For this model,~\citet{jia2020model} propose a UCB-based RL algorithm with value-targeted model parameter estimation that guarantees an upper bound of $\tilde{\Ocal}(d H^{\frac{3}{2}} \sqrt{T})$ for regret.
The same linear mixture MDPs have been used in other studies such as~\citet{ayoub2020model, zhou2021nearly, zhou2021provably}. 
Specifically, in~\citet{zhou2021nearly}, a variant of the method proposed by~\citet{jia2020model} is suggested and proved that the algorithm guarantees an upper bound of $\tilde{\Ocal}(d H \sqrt{T})$ regret with a matching lower bound of $\Omega(d H \sqrt{T})$ for linear mixture MDPs.
More recently, there are also works achieving horizon-free regret bounds for linear mixture MDPs~\cite{zhang2021improved, kim2022improved, zhou2022computationally}.

\paragraph{RL with non-linear function approximation}
Studies have been conducted on extending function approximation beyond linear models.
\citet{ayoub2020model, wang2020reinforcement_eluder, ishfaq2021randomized} provide upper bound for regret based on eluder dimension~\citep{russo2013eluder}.
Also, there has been an effort to develop sample-efficient methods with more ``general'' function approximation~\cite{krishnamurthy2016pac, jiang2017contextual, dann2018oracle, du2019provably, du2021bilinear, foster2021statistical, ishfaq2021randomized, jin2021bellman, agarwal2022model, agarwal2022non, zhang2022feel, chen2023a, ishfaq2024more}
However, these attempts may have been hindered by the difficulty of solving computationally intractable problems~\citep{krishnamurthy2016pac, jiang2017contextual, dann2018oracle, du2021bilinear, foster2021statistical, jin2021bellman, chen2023a}, the necessity of relying on stronger assumptions~\citep{du2019provably, ishfaq2021randomized}, or the lack of discussion on how to define the posterior distribution supported by a given function class and how to draw the optimistic sample from the posterior~\cite{agarwal2022model, agarwal2022non, zhang2022feel}.
That is why even after there exists a so-called ``general function class''-based result, it is often the case that the results in specific parametric models are still needed.
Despite the large number of studies on RL with linear function approximation, there is limited research on extending beyond linear models to other parametric models. 
\citet{wang2021glmRL} use generalized linear function approximation, where the Bellman backup of any value function is assumed to be a generalized linear function of feature mapping.
\citet{hwang2023model} discuss the limitations of linear function approximation and propose a UCB-based algorithm for MNL transition model in feature space achieving $\tilde{\Ocal}(d H^{\frac{3}{2}} \sqrt{T})$.
~\citet{ishfaq2024more} present TS-based RL algorithms that utilize approximate samplers, such as LMC or Underdamped LMC, to enhance the implementation and computational tractability of TS for RL with general function classes.

\paragraph{Contextual bandits}
~\citet{faury2020improved} first provide a UCB-based algorithm with $\kappa$-independent regret for binary logistic bandit and~\citet{abeille2021instance} present UCB \& TS based algorithms achieving nearly minimax optimal regret for the same setting.
~\citet{faury2022jointly} propose a jointly efficient UCB-based algorithm that achieve $\kappa$-independent regret bound with $\Ocal(\log t)$ computation cost.
In the context of MNL model,~\citet{oh2019thompson} employ TS approach, while~\citet{oh2021multinomial} incorporate a combination of UCB exploration and online parameter updates for MNL bandits. 
Both of the methods have $\Ocal(\kappa^{-1} \sqrt{T})$ regret.
~\citet{amani2021ucb} propose an optimistic algorithm with better dependence on $\kappa$.
~\citet{agrawal2023tractable} design a UCB-based algorithm with $\Ocal(\sqrt{T})$ regret bound without $\kappa$ in its leading term, and~\citet{perivier2022dynamic} establish $\Ocal(\sqrt{T/\kappa_*})$ regret for the uniform reward setting.
~\citet{zhang2023online} develop jointly efficient UCB-based algorithm for non-uniform MNL bandit problem.
~\citet{lee2024nearly} propose nearly minimax optimal MNL bandit algorithm for both uniform and non-uniform reward structures.

\section{Notations \& Definitions}
In this section, we formally summarize some definitions and notations used to analyze the proposed algorithm.

\subsection*{Inhomogeneous MNL transition model}
For $h \in [H]$, the probability of state transition to $s' \in \Scal_{s,a}$ when an action $a$ is taken at a state $s$ is given by
\begin{equation*}
    P_h(s' \mid s,a) := P_{\thetab^*_h}(s' \mid s,a) = \frac{\exp(\varphib(s,a,s')^\top \thetab^*_h)}{\sum_{\tilde{s} \in \Scal_{s,a}} \exp(\varphib(s,a,\tilde{s})^\top \thetab^*_h)} \, .
\end{equation*}
The estimated transition probability parameterized by $\thetab$ is denoted as
\begin{equation*}
    P_{\thetab}(s' \mid s,a) := \frac{\exp(\varphib(s,a,s')^\top \thetab)}{\sum_{\tilde{s} \in \Scal_{s,a}} \exp(\varphib(s,a,\tilde{s})^\top \thetab)}  \, .
\end{equation*}

\subsection*{Feature vector}
We abbreviate the feature vector as follows:
\begin{align*}
    & \varphib_{s,a,s'} := \varphib(s,a,s') \; \text{ for } (s,a,s') \in \Scal \times \Acal \times \Scal_{s,a} \, ,
    \\
    & \varphib_{k,h,s'} := \varphib(s^k_h, a^k_h, s') \; \text{ for } (k,h) \in [K] \times [H] \text{ and } s' \in \Scal_{k,h}:= \Scal_{s^k_h, a^k_h} \, ,
    \\
    & \DomFeat_{k,h}(s,a) := \varphib(s,a,\hat{s}) \; \text{ for } \hat{s} := \argmax_{s' \in \Scal_{s,a}} \| \varphib(s,a,s') \|_{\Ab_{k,h}^{-1}} \, ,
    \\
    & \bar{\varphib}_{s,a,s'}(\thetab) := \bar{\varphib}(s,a,s'; \thetab) = \varphib(s,a,s') - \EE_{\tilde{s} \sim P_{\thetab}(\cdot \mid s,a)} [\varphib(s,a, \tilde{s})] \, ,
    \\
    & \bar{\varphib}_{k,h,s'}(\thetab) := \bar{\varphib}(s^k_h,a^k_h,s'; \thetab) \, .
\end{align*}

\subsection*{Response variable \& per-episode loss}
The response variable $y^k_h$ is given by
\begin{equation*}
    y^k_h := [y^k_h(s')]_{s' \in \Scal_{k,h}} \; \text{ where } y^k_h(s') := \ind(s^k_{h+1} = s') \; \text{ for } s' \in \Scal_{k,h} \, .
\end{equation*}
The per-episode loss $\ell_{k,h}(\thetab)$ is given by
\begin{align*}
    & \ell_{k,h} (\thetab) := - \sum_{s' \in \Scal_{k,h}} y^k_h(s') \log P_{\thetab}(s' \mid s^k_h, a^k_h) \, , 
    \\
    & \Gb_{k,h} (\thetab) := \nabla \ell_{k,h}(\thetab) = \sum_{s' \in \Scal_{k,h}} (P_{\thetab}(s' \mid s^k_h, a^k_h) - y^k_h(s')) \varphib_{k,h,s'} \, ,
    \\
    & \Hb_{k,h} (\thetab) := \nabla^2 \ell_{k,h}(\thetab)
        \\
        & \phantom{{}}
        = \sum_{s' \in \Scal_{k,h}} P_{\thetab}(s' \mid s^k_h, a^k_h) \varphib_{k, h,s'} \varphib_{k, h,s'}^\top
                - \sum_{s' \in \Scal_{k,h}} \sum_{\tilde{s} \in \Scal_{k,h}} P_{\thetab} (s' \mid s^k_h, a^k_h) P_{\thetab}(\tilde{s} \mid s^k_h, a^k_h) \varphib_{k, h,s'} \varphib_{k, h,\tilde{s}}^\top \, .
\end{align*}

\subsection*{Regularity constants}
\begin{align*}
    & H : \text{Horizon length}
    \\
    & K : \text{Episode number}
    \\
    & T = KH : \text{Total number of interactions}
    \\
    & \Fnorm : \ell_2\text{-norm upper bound of } \varphib(s,a,s), \; \text{i.e., } \| \varphib(s,a,s') \|_2 \le \Fnorm \, ,
    \\
    & \Pnorm : \ell_2\text{-norm upper bound of } \thetab^*_h, \; \text{i.e., } \| \thetab^*_h \|_2 \le \Pnorm \, ,
    \\
    & \kappa : \text{Problem-dependent constant such that } \inf_{\thetab \in \Bcal_d(\Pnorm)} P_{\thetab} (s' \mid s, a) P_{\thetab} ( \tilde{s} \mid s, a) \ge \kappa \, ,
    \\
    & \Ucal : \text{Maximum cardinality of the set of reachable states, i.e., } \Ucal:= \max_{s,a} |\Scal_{s,a}| \, .
\end{align*}

\subsection*{Estimated transition core}
The estimated transition core for $\texttt{RRL-MNL}$ is given by
\begin{equation*}
    \thetab^k_h = \argmin_{\thetab \in \Bcal_d (\Pnorm)} \frac{1}{2} \| \thetab - \thetab^{k-1}_{h} \|_{\Ab_{k,h}}^2
        + (\thetab - \thetab^{k-1}_h)^\top \nabla \ell_{k-1,h}(\thetab^{k-1}_h) \, ,
\end{equation*}
and the estimated transition core for $\texttt{ORRL-MNL}$ is given by
\begin{equation*}
    \omdtheta{k+1}{h} = \argmin_{\thetab \in \Bcal_d(\Pnorm)} \frac{1}{2 \eta} \left\| \thetab - \omdtheta{k}{h} \right\|_{\tilde{\Bb}_{k,h}}^2 + \thetab^\top \nabla \ell_{k,h}(\omdtheta{k}{h}) \, .
\end{equation*}

\subsection*{Gram matrices}
The Gram matrix with global gradient information $\kappa$ is given by
\begin{equation*}
    \Ab_{k,h} := \lambda \Ib_d + \frac{\kappa}{2} \sum_{i=1}^{k-1} \sum_{s' \in \Scal_{i, h}} \varphib(s^{i}_h, a^{i}_h, s') \varphib(s^{i}_h, a^{i}_h, s')^\top \, .
\end{equation*}
The Gram matrices with local gradient information are given by
\begin{equation*}
    \tilde{\Bb}_{k,h} := \Bb_{k,h} + \eta \nabla^2 \ell_{k,h} (\omdtheta{k}{h}) \; \text{ and } \;
    \Bb_{k,h} := \lambda \Ib_d + \sum_{i=1}^{k-1} \nabla^2 \ell_{i,h} (\omdtheta{i+1}{h}) \, .
\end{equation*}

\subsection*{Confidence radius}
For some absolute constants $C_\beta, C_{\xib} > 0$,
\begin{align*}
    &\alpha_k := \alpha_{k} (\delta) 
    \\ &=
    \sqrt{\frac{8 d}{\kappa} \log \left( 1 + \frac{k \Ucal \Fnorm^2}{d \lambda} \right)
    + \left( \frac{32 \Fnorm \Pnorm}{3} + \frac{16}{\kappa} \right) \log \frac{ \left( 1 + \lceil 2 \log_2 k \Ucal \Fnorm \Pnorm \rceil \right) k^2}{\delta}
    + 2 \sqrt{2} + 2 \lambda \Pnorm^2} 
    \\
    &= \tilde{\Ocal}(\kappa^{-1/2} d^{1/2})  \, ,
    \\
    & \beta_k := \beta_k (\delta)
        = C_\beta \sqrt{ \log \Ucal \left(  \lambda \log (\Ucal k)
        + \log (\Ucal k)  \log \left( \frac{H\sqrt{1 + 2k}}{\delta} \right) 
        + d \log \left( 1 + \frac{k}{d \lambda}\right) \right)
        + \lambda L_{\thetab}^2}
        \\
        & \phantom{{}={}}
        = \Ocal(\sqrt{d} \log \Ucal \log (kH)) \, ,    
    \\
    & \gamma_k := \gamma_k (\delta) = C_{\xib} \sigma_k \sqrt{d \log (M d/ \delta)} \, .
\end{align*}

\subsection*{Filtration}
For an arbitrary set $X$, we denote the $\Sigma$-algebra generated by $X$ as $\Sigma(X)$.
Then we define the following filtrations
\begin{align*}
    & \Fcal_k := \Sigma \left( \left\{ s^i_j, a^i_j, r(s^i_j, a^i_j) \mid i < k, j \le H \right\} \cup \left\{ \xib^{(m)}_{i,j} \mid i < k, j \le H, 1 \le m \le M \right\} \right) \, ,
    \\
    & \Fcal_{k,h} := \Sigma \left( \Fcal_k \cup \left\{ s^k_j, a^k_j, r(s^k_j, a^k_j) \mid j \le h \right\} 
    \cup \left\{ \xib_{k, j}^{(m)} \mid j \ge h, 1 \le m \le M \right\} \right) \, .
\end{align*}

\subsection*{Pseudo-noise}
For $\texttt{RRL-MNL}$, the pseudo-noise is sampled as
\begin{equation*}
    \xib_{k,h}^{(m)} \sim \Ncal( \zero_d, \sigma_k^2 \Ab_{k,h}^{-1}) \, ,
\end{equation*}
and for $\texttt{ORRL-MNL}$, the pseudo-noise is sampled as
\begin{equation*}
    \xib_{k,h}^{(m)} \sim \Ncal( \zero_d, \sigma_k^2 \Bb_{k,h}^{-1}) \, ,
\end{equation*}
for $M$ times independently.

\subsection*{Estimated value functions}
The stochastically optimistic value function for $\texttt{RRL-MNL}$ is defined as follows:
\begin{align*}
    & Q^k_{H+1} (s,a) = 0 \, ,
    \\
    & Q^k_h (s,a) =  \min \bigg\{ r(s,a) + \sum_{s' \in \Scal_{s,a}} P_{\thetab^k_h} (s' \mid s,a) V^k_{h+1}(s')
    + \max_{m \in [M]} \hat{\varphib}_{k,h}(s,a)^\top \xib_{k,h}^{(m)}, H \bigg\} \; \text{for } h \in [H] \, .
\end{align*}
The optimistic randomized value function for $\texttt{ORRL-MNL}$ is defined as follows:
\begin{align*}
    & \tilde{Q}^k_{H+1}(s,a) = 0 \, ,
    \\
    & \tilde{Q}^k_h(s,a) := \min \bigg\{ r(s,a) + \sum_{s' \in \Scal_{s,a}} P_{\omdtheta{k}{h}}(s' \mid s,a) \tilde{V}^k_{h+1}(s') + \rbonus_{k,h}(s,a)\, , H \bigg\} \; \text{for } h \in [H] \, , 
\end{align*}
where 
\begin{align*}
    & \rbonus_{k,h}(s,a) := \sum_{s' \in \Scal_{s,a}} P_{\omdtheta{k}{h}} (s' \mid s, a) \bar{\varphib}(s, a, s'; \omdtheta{k}{h})^\top \xib^{s'}_{k,h}
    + 3 H \beta_k^2 \max_{s' \in \Scal_{s,a}} \| \varphib (s,a,s') \|^2_{\Bb^{-1}_{k,h}} \, ,
    \\
    & \xib^{s'}_{k,h} := \xib^{m(s')}_{k,h} \; \text{for } m(s') := \argmax_{m \in [M]} \bar{\varphib}(s, a, s'; \omdtheta{k}{h})^\top \xib^{m}_{k,h} \, .
\end{align*}

\subsection*{Prediction error \& Bellman error}
\begin{definition}[Prediction error \& Bellman error] \label{def:prediction error & bellman error}
    For any $(s, a) \in \Scal \times \Acal$ and $(k,h) \in [K] \times [H]$, we define the prediction error about $\thetab^k_h$ as
    \begin{equation*}
        \PE^k_h (s,a)
        := \sum_{s' \in \Scal_{s,a}} \left( P_{\thetab^k_h} (s' \mid s, a ) - P_{\thetab^*_h}(s' | s, a) \right) V^k_{h+1}(s') \, .
    \end{equation*}
    Also we define the Bellman error as follows:
    \begin{equation*}
        \BE^k_h (s,a)
        := r(s,a) + P_h V^k_{h+1}(s,a) - Q^k_h(s,a) \, .
    \end{equation*}        
\end{definition}

\subsection*{Good events}
For any $\delta \in (0,1)$, we define the following good events:

For $\texttt{RRL-MNL}$,
\begin{align*}
        & \Gcal_{k,h}^\Delta (\delta) := \left\{ | \Delta^k_h(s,a) | \le H \alpha_k(\delta) \lVert \hat{\varphib}_{k,h}(s,a) \rVert_{\Ab_{k,h}^{-1}} \right\} \, ,
        \\
        & \Gcal_{k,h}^{\xib} (\delta) := \left\{ \max_{m \in [M]} \lVert \xib_{k,h}^{(m)} \rVert_{\Ab_{k,h}} \le \gamma_k (\delta) \right\} \, ,
        \\
        & \Gcal_{k,h} (\delta) := \left\{ \Gcal_{k,h}^\Delta (\delta) \cap \Gcal_{k,h}^{\xib} (\delta) \right\} \, ,
        \\
        & \Gcal_k (\delta) := \bigcap_{h \in [H]} \Gcal_{k,h}(\delta) \, ,
        \\
        & \Gcal(K, \delta) := \bigcap_{k \le K} \Gcal_{k} (\delta) \, .    
\end{align*}

For $\texttt{ORRL-MNL}$, 
\begin{align*}
    & \Gfrak_{k,h}^\Delta (\delta) 
    := \bigg\{
        | \Delta^k_h(s,a) | \le H \beta_k (\delta) \!\!\!\! \sum_{s' \in \Scal_{s,a}} \!\!\!\! P_{\omdtheta{k}{h}} (s' \mid s, a) \left\| \bar{\varphib}_{s,a,s'} (\omdtheta{k}{h}) \right\|_{\Bb^{-1}_{k,h}}
        \\
        &\phantom{{}={}={}={}={}} + 3H \beta_k (\delta)^2 \max_{s' \in \Scal_{s,a}} \| \varphib_{s,a,s'} \|^2_{\Bb^{-1}_{k,h}}
        \bigg\} \, ,
    \\
    & \Gfrak_{k,h}^{\xib} (\delta) := \left\{ \max_{m \in [M]} \lVert \xib_{k,h}^{(m)} \rVert_{\Bb_{k,h}} \le \gamma_k (\delta) \right\} \, ,
    \\
    & \Gfrak_{k,h} (\delta) := \left\{ \Gfrak_{k,h}^\Delta (\delta) \cap \Gfrak_{k,h}^{\xib} (\delta) \right\} \, ,
    \\
    & \Gfrak_k (\delta) := \bigcap_{h \in [H]} \Gfrak_{k,h}(\delta) \, ,
    \\
    & \Gfrak(K, \delta) := \bigcap_{k \le K} \Gfrak_{k} (\delta) \, .    
\end{align*}

\subsection*{Derivative of MNL transition model}
\begin{proposition}[Derivative of MNL transition model] \label{prop:derivative of transition prob}
    The gradient and Hessian of $P_{\thetab} (\cdot \mid \cdot, \cdot)$ can be calculated as follows:
    \begin{equation}
        \begin{split}
            \nabla P_{\thetab} (s' \mid s, a) 
            & = P_{\thetab} (s' \mid s, a) \left( {\varphib}_{s, a, s'} - \sum_{s'' \in \Scal_{s,a}} P_{\thetab} (s'' \mid s, a) {\varphib}_{s, a, s''} \right)
            \\
            & = P_{\thetab} (s' \mid s, a) \bar{\varphib}_{s,a,s'}(\thetab) \, ,
        \end{split}
    \end{equation}
    and 
    \begin{equation}
        \begin{split}
            &\nabla^2 P_{\thetab} (s' \mid s, a) \\
            &\,= P_{\thetab} (s' \mid s, a) \varphib_{s,a,s'} \varphib_{s,a,s'}^\top \\
            &\phantom{{}={}} - P_{\thetab} (s' \mid s, a) \sum_{s'' \in \Scal_{s,a}} P_{\thetab} (s'' \mid s, a) \left( \varphib_{s,a,s'} \varphib_{s,a,s''}^\top + \varphib_{s,a,s''} \varphib_{s,a,s'}^\top + \varphib_{s,a,s''} \varphib_{s,a,s''}^\top \right) \\
            &\phantom{{}={}} + 2P_{\thetab} (s' \mid s, a) \left( \sum_{s'' \in \Scal_{s,a}}  P_{\thetab}(s'' \mid s, a) \varphib_{s,a,s''}  \right) \left(\sum_{s'' \in \Scal_{s,a}} P_{\thetab}(s'' \mid s, a) \varphib_{s,a,s''} \right)^\top \, .
        \end{split}
    \end{equation}
\end{proposition}

\begin{proof}[Proof of Proposition~\ref{prop:derivative of transition prob}]
    Let $\thetab = (\theta_1, \ldots, \theta_d)$ and $[\varphib_{s,a,s'}]_i$ be the $i$-th component of $\varphib_{s,a,s'}$.
    Then, we have
    \begin{align*}
        &\frac{\partial}{\partial \theta_j} P_{\thetab} (s' \mid s, a) \\
        & = \frac{\exp \left( \varphib_{s,a,s'}^\top \thetab \right) [\varphib_{s,a,s'}]_j }{  \sum_{s'' \in \Scal_{s,a}} \exp \left( \varphib_{s,a,s''}^\top \thetab \right) }
            - \frac{\exp \left( \varphib_{s,a,s'}^\top \thetab \right) \sum_{s'' \in \Scal_{s,a}} \exp \left( \varphib_{s,a,s''}^\top \thetab \right) [\varphib_{s,a,s''}]_j }{ \left( \sum_{s'' \in \Scal_{s,a}} \exp \left( \varphib_{s,a,s''}^\top \thetab \right) \right)^2 }
        \\
        & = P_{\thetab} (s' \mid s, a)  \left( [\varphib_{s,a,s'}]_j - \sum_{s'' \in \Scal_{s,a}} P_{\thetab} (s'' \mid s,a) [\varphib_{s,a,s''}]_j \right) \, .
    \end{align*}
    Then, the gradient of $P_{\thetab}(s' \mid s,a)$ is given by 
    \begin{align*}
        \nabla P_{\thetab} (s' \mid s, a)
        & = P_{\thetab} (s' \mid s,a) \varphib_{s,a,s'}
            - P_{\thetab} (s' \mid s, a) \sum_{s'' \in \Scal_{s,a}} P_{\thetab} ( s'' \mid s,a ) \varphib_{s,a,s''}
        \\
        & = P_{\thetab} (s' \mid s,a) \left( \varphib_{s,a,s'} - \sum_{s'' \in \Scal_{s,a} } P_{\thetab} ( s'' \mid s,a ) \varphib_{s,a,s''} \right)
        \\
        & = P_{\thetab} (s' \mid s,a) \bar{\varphib}_{s,a,s'} (\thetab) \, .
    \end{align*}
    
    On the other hand, the second derivative $\frac{\partial}{\partial \theta_i \partial \theta_j} P_{\thetab} (s' \mid s, a)$ can be obtained as follows:
    \begin{align*}
        & \frac{\partial}{\partial \theta_i \partial \theta_j} P_{\thetab} (s' \mid s, a)
        \\
        & = P_{\thetab} (s' \mid s, a) 
            \left( [\varphib_{s,a,s'}]_i - \sum_{s'' \in \Scal_{s,a}} P_{\thetab} (s'' \mid s,a) [\varphib_{s,a,s''}]_i \right)
            \\
            &\phantom{{}={}} \cdot \left( [\varphib_{s,a,s'}]_j - \sum_{s'' \in \Scal_{s,a}} P_{\thetab} (s'' \mid s,a) [\varphib_{s,a,s''}]_j \right)
            \\
            & \,\,
            + P_{\thetab} (s' \mid s, a) \left(  - \sum_{s'' \in \Scal_{s,a}} P_{\thetab} (s'' \mid s,a)  \left( [\varphib_{s,a,s''}]_i - \sum_{\tilde{s} \in \Scal_{s,a}} P_{\thetab} (\tilde{s} \mid s,a) [\varphib_{s, a, \tilde{s}}]_i \right) [\varphib_{s,a,s''}]_j \right)
            \\
        & = P_{\thetab} (s' \mid s, a)
            \Bigg\{ 
            [\varphib_{s,a,s'}]_i [\varphib_{s,a,s'}]_j
            \\
            & \phantom{{}={}} - \sum_{s'' \in \Scal_{s,a}} P_{\thetab}(s'' \mid s, a) 
            \left( [\varphib_{s,a,s''}]_i [\varphib_{s,a,s'}]_j + [\varphib_{s,a,s'}]_i [\varphib_{s,a,s''}]_j \right)
            \\
            & \phantom{{}={}}
            + \left( \sum_{s'' \in \Scal_{s,a}} P_{\thetab}(s'' \mid s, a) [\varphib_{s,a,s''}]_i \right)
            \left( \sum_{s'' \in \Scal_{s,a}} P_{\thetab}(s'' \mid s, a) [\varphib_{s,a,s''}]_j \right)      
            \\     
            & \phantom{{}={}}
            - \sum_{s'' \in \Scal_{s,a}} P_{\thetab}(s'' \mid s, a) [\varphib_{s,a,s''}]_i [\varphib_{s,a,s''}]_j \\
            & \phantom{{}={}}
            \left. + \left( \sum_{s'' \in \Scal_{s,a}} P_{\thetab}(s'' \mid s, a) [\varphib_{s,a,s''}]_j \right) \left( \sum_{\tilde{s} \in \Scal_{s,a}} P_{\thetab}(\tilde{s} \mid s, a) [\varphib_{s,a,\tilde{s}}]_i \right) \right\}
            \\            
        & = P_{\thetab} (s' \mid s, a)
            \left\{ [\varphib_{s,a,s'}]_i [\varphib_{s,a,s'}]_j \right.
            \\ 
            &- \sum_{s'' \in \Scal_{s,a}} P_{\thetab}(s'' \mid s, a) \left( [\varphib_{s,a,s''}]_i [\varphib_{s,a,s'}]_j + [\varphib_{s,a,s'}]_i [\varphib_{s,a,s''}]_j \right)
            \\            
            & \phantom{{}={}}
            - \sum_{s'' \in \Scal_{s,a}} P_{\thetab}(s'' \mid s, a) [\varphib_{s,a,s''}]_i [\varphib_{s,a,s''}]_j \\
            & \phantom{{}={}} 
            \left. +2\left( \sum_{s'' \in \Scal_{s,a}} P_{\thetab}(s'' \mid s, a) [\varphib_{s,a,s''}]_i \right) \left( \sum_{s'' \in \Scal_{s,a}} P_{\thetab}(s'' \mid s, a) [\varphib_{s,a,s''}]_j \right)
            \right\} \, .
    \end{align*}
    Thus, we get the desired result as follows:
    \begin{align*}
        &\nabla^2 P_{\thetab} (s' \mid s, a) \\
        &\, = P_{\thetab} (s' \mid s, a) \varphib_{s,a,s'} \varphib_{s,a,s'}^\top \\
        &\phantom{{}={}} - P_{\thetab} (s' \mid s, a) \sum_{s'' \in \Scal_{s,a}} P_{\thetab} (s'' \mid s, a) \left( \varphib_{s,a,s'} \varphib_{s,a,s''}^\top + \varphib_{s,a,s''} \varphib_{s,a,s'}^\top + \varphib_{s,a,s''} \varphib_{s,a,s''}^\top \right) \\
        &\phantom{{}={}} + 2P_{\thetab} (s' \mid s, a) \left( \sum_{s'' \in \Scal_{s,a}}  P_{\thetab}(s'' \mid s, a) \varphib_{s,a,s''}  \right) \left(\sum_{s'' \in \Scal_{s,a}} P_{\thetab}(s'' \mid s, a)  \varphib_{s,a,s''} \right)^\top 
        \, .
    \end{align*}
\end{proof}

\section{Detailed Regret Analysis for \texorpdfstring{$\texttt{RRL-MNL}$}{RRL-MNL} (Theorem~\ref{thm:alg 1})} \label{appx:regret bound of alg1}
In this section, we provide the complete proof of Theorem~\ref{thm:alg 1}. First, we introduce all the technical lemmas needed to prove Theorem~\ref{thm:alg 1} along with their proofs. At the end of this section, we present the proof of Theorem~\ref{thm:alg 1}.

\subsection{Concentration of Estimated Transition Core \texorpdfstring{$\thetab^k_h$}{}}
In this section, we provide the concentration inequality for the estimated transition core run by the approximate online Newton step. The proof is similar to that given by \citet{oh2021multinomial}. For completeness, we provide the detailed proof.
\begin{lemma}[Concentration of online estimated transition core] \label{lemma:concentration of online estimated transition core}
    For each $h \in [H]$, if $\lambda \ge \Fnorm^2$, then we have
    \begin{equation*}
        \PP \left(\forall k \ge 1 , \| \thetab^k_h - \thetab_h^* \|_{\Ab_{k,h}} \le \alpha_{k}(\delta) \right) \ge 1 - \delta \, .
    \end{equation*}
    where $\alpha_{k}(\delta)$ is given by
    \begin{align*} 
        &\alpha_{k} (\delta) 
        \\
        &:= 
            \sqrt{\frac{8d}{\kappa} \log \left( 1 + \frac{k \Ucal \Fnorm^2}{d \lambda} \right)
            + \left( \frac{32 \Fnorm \Pnorm}{3} + \frac{16}{\kappa} \right) \log \frac{ \left( 1 + \lceil 2 \log_2 k \Ucal \Fnorm \Pnorm \rceil \right) k^2}{\delta}
            + 2 \sqrt{2} + 2 \lambda \Pnorm^2} \, .
    \end{align*}
\end{lemma}

\begin{proof}[Proof of~\cref{lemma:concentration of online estimated transition core}]
    Recall that the per-round loss $\ell_{k,h}(\thetab)$ and its gradient $\Gb_{k,h}(\thetab)$ is defined as follows:
    \begin{equation*}
        \ell_{k,h} (\thetab) := - \sum_{s' \in \Scal_{k,h}} y^k_h(s') \log P_{\thetab}(s' \mid s^k_h, a^k_h) \, ,
        \quad
        \Gb_{k,h}(\thetab) := \nabla_{\thetab} \ell_{k,h}(\thetab) \, .
    \end{equation*}
    For the analysis, we define the conditional expectations of $\ell_{k,h}(\thetab)$ \& $\Gb_{k,h}(\thetab)$ as follows:
    \begin{equation*}
        \bar{\ell}_{k,h}(\thetab):= \EE_{y^k_h} \left[ \ell_{k,h}(\thetab) \mid \Fcal_{k,h} \right] \, , \quad \bar{\Gb}_{k,h}(\thetab) := \EE_{y^k_h} [\Gb_{k,h}(\thetab) \mid \Fcal_{k,h}] \, .
    \end{equation*}

    By Taylor expansion with $\bar{\thetab}= \nu \thetab^k_h + (1- \nu) \thetab_h^*$ for some $\nu \in (0,1)$, we have
    \begin{equation} \label{eq:online concentration talyor}
        \ell_{k,h}(\thetab_h^*) = \ell_{k,h}(\thetab^k_h) + \Gb_{k,h}(\thetab^k_h)^\top (\thetab_h^* - \thetab^k_h)
        + \frac{1}{2} (\thetab_h^* - \thetab^k_h)^\top \Hb_{k,h}(\bar{\thetab}) (\thetab_h^* - \thetab^k_h) \, ,
    \end{equation}
    where $\Hb_{k,h}(\thetab)$ is the Hessian of the per-round loss evaluated at $\thetab$, i.e., 
    \begin{align*}
        \Hb_{k,h}(\thetab)
            & := \nabla^2 \ell_{k,h}(\thetab)
            \numberthis 
            \label{eq:hessian of per-round loss}   
            \\
            & = \sum_{s' \in \Scal_{k,h}} P_{\thetab}(s' \mid s^k_h, a^k_h) \varphib_{k, h,s'} \varphib_{k, h,s'}^\top    
            \\ 
            & \phantom{{}={}} - \sum_{s', \tilde{s} \in \Scal_{k,h}} P_{\thetab} (s' \mid s^k_h, a^k_h) P_{\thetab}(\tilde{s} \mid s^k_h, a^k_h) \varphib_{k, h,s'} \varphib_{k, h,\tilde{s}}^\top \, .     
            \nonumber
    \end{align*}

    Note that for $\bar{\thetab}= \nu \thetab^k_h + (1- \nu) \thetab_h^*$ with $\nu \in (0,1)$, we have
    \begin{align*}
        \Hb_{k,h} (\bar{\thetab}) 
        & = \sum_{s' \in \Scal_{k,h}} P_{\bar{\thetab}}(s' \mid s^k_h, a^k_h) \varphib_{k, h,s'} \varphib_{k, h,s'}^\top
            \\
            & \quad 
            - \sum_{s' \in \Scal_{k,h}} \sum_{\tilde{s} \in \Scal_{k,h}}  P_{\bar{\thetab}} (s' \mid s^k_h, a^k_h) P_{\bar{\thetab}}(\tilde{s} \mid s^k_h, a^k_h) \varphib_{k, h,s'} \varphib_{k, h,\tilde{s}}^\top
        \\
        & = \sum_{s' \in \Scal_{k,h} } P_{\bar{\thetab}}(s' \mid s^k_h, a^k_h) \varphib_{k, h, s'} \varphib_{k, h, s'}^\top \nonumber
            \\
            & \phantom{{}={}} - \frac{1}{2} \sum_{s' \in \Scal_{k,h} } \sum_{\tilde{s} \in \Scal_{k,h} } P_{\bar{\thetab}}(s' \mid s^k_h, a^k_h) P_{\bar{\thetab}}(\tilde{s} \mid s^k_h, a^k_h)  (\varphib_{k, h, s'} \varphib_{k, h, \tilde{s}}^\top + \varphib_{k, h, \tilde{s}} \varphib_{i, h, s'}^\top) \nonumber   
        \\
        & \succeq \sum_{s' \in \Scal_{k,h} } P_{\bar{\thetab}}(s' \mid s^k_h, a^k_h)  \varphib_{k, h, s'} \varphib_{k, h, s'}^\top \nonumber
            \\
            & \phantom{{}={}}- \frac{1}{2} \sum_{s' \in \Scal_{k,h} } \sum_{\tilde{s} \in \Scal_{k,h} } P_{\bar{\thetab}}(s' \mid s^k_h, a^k_h) P_{\bar{\thetab}}( \tilde{s} \mid s^k_h, a^k_h)  (\varphib_{k, h, s'} \varphib_{k, h, s'}^\top + \varphib_{k, h, \tilde{s}} \varphib_{k, h, \tilde{s}}^\top) \nonumber
        \\
        & = \sum_{s' \in \Scal_{k,h} } \!\!\! P_{\bar{\thetab}}(s' \mid s^k_h, a^k_h)  \varphib_{k, h, s'} \varphib_{k, h, s'}^\top
            \\
            & \quad 
            - \sum_{s' \in \Scal_{k,h} } \sum_{\tilde{s} \in \Scal_{k,h} } \!\!\! P_{\bar{\thetab}}(s' \mid s^k_h, a^k_h) P_{\bar{\thetab}}(\tilde{s} \mid s^k_h, a^k_h)  \varphib_{k, h, s'} \varphib_{k, h, s'}^\top \, , \nonumber
    \end{align*}
    where the inequality utilizes the fact that $\xb \xb^\top + \yb \yb^\top \succeq \xb \yb^\top + \yb \xb^\top $ for any $\xb, \yb \in \RR^d$.    
    Therefore, we have
    \begin{align*}
        \Hb_{k,h} (\bar{\thetab}) 
        & \succeq \sum_{s' \in \Scal_{k,h} } P_{\bar{\thetab}}(s' \mid s^k_h, a^k_h)  \varphib_{k, h, s'} \varphib_{k, h, s'}^\top
            \\
            & \quad 
            - \sum_{s' \in \Scal_{k,h} } \sum_{\tilde{s} \in \Scal_{k,h} } P_{\bar{\thetab}}(s' \mid s^k_h, a^k_h) P_{\bar{\thetab}}(\tilde{s} \mid s^k_h, a^k_h)  \varphib_{k, h, s'} \varphib_{k, h, s'}^\top \nonumber
        \\
        & = \sum_{s' \ne \dot{s}_{k,h}} P_{\bar{\thetab}}(s' \mid s^k_h, a^k_h)  \varphib_{k, h, s'} \varphib_{k, h, s'}^\top
            \\
            & \quad
            - \sum_{s' \ne \dot{s}_{k,h} } \sum_{\tilde{s} \ne \dot{s}_{k,h}} P_{\bar{\thetab}}(s' \mid s^k_h, a^k_h) P_{\bar{\thetab}}(\tilde{s} \mid s^k_h, a^k_h)  \varphib_{k, h, s'} \varphib_{k, h, s'}^\top \nonumber
        \\
        & = \sum_{s' \ne \dot{s}_{k,h}} P_{\bar{\thetab}}(s' \mid s^k_h, a^k_h) \left( 1 - \sum_{\tilde{s} \ne \dot{s}_{k,h}} P_{\bar{\thetab}}(\tilde{s} \mid s^k_h, a^k_h) \right) \varphib_{k,h, s'} \varphib_{k, h, s'}^\top 
        \\
        & = \sum_{s' \ne \dot{s}_{k,h}} P_{\bar{\thetab}}(s' \mid s^k_h, a^k_h) P_{\bar{\thetab}}(\dot{s}_{k,h} \mid s^k_h, a^k_h)  \varphib_{k,h, s'} \varphib_{k, h, s'}^\top 
        \\
        & \succeq \sum_{s' \ne \dot{s}_{k,h} } \kappa \varphib_{k,h, s'} \varphib_{k, h, s'}^\top 
        \\
        & = \sum_{s' \in \Scal_{k,h} } \kappa \varphib_{k,h, s'} \varphib_{k, h, s'}^\top  \, ,
    \end{align*}
    where $\dot{s}_{k,h}$ is the state satisfying $\varphib(s^k_h, a^k_h, \dot{s}_{k,h})=\zero_d$ and the last inequality comes from the Assumption~\ref{assm:positive kappa}.    

   Using the lower bound of the Hessian of the per-round loss evaluated at $\bar{\thetab}$, from~\eqref{eq:online concentration talyor} we have
    \begin{equation*}
        \ell_{k,h}(\thetab_h^*) \ge \ell_{k,h}(\thetab^k_h) + \Gb_{k,h}(\thetab^k_h)^\top (\thetab_h^* - \thetab^k_h)
        + \frac{\kappa}{2} (\thetab_h^* - \thetab^k_h)^\top \left( \sum_{s' \in \Scal_{k,h}} \varphib_{k, h,s'} \varphib_{k, h,s'}^\top  \right) (\thetab_h^* - \thetab^k_h) \, .
    \end{equation*}
    By rearranging, we have
    \begin{equation*}
        \ell_{k,h}(\thetab^k_h)
        \le 
        \ell_{k,h}(\thetab_h^*) 
        + \Gb_{k,h}(\thetab^k_h)^\top (\thetab^k_h - \thetab_h^*)
        - \frac{\kappa}{2} (\thetab_h^* - \thetab^k_h)^\top \Wb_{k,h}  (\thetab_h^* - \thetab^k_h) \, ,
    \end{equation*}
    where we denote $\Wb_{k,h}:= \sum_{s' \in \Scal_{k,h}} \varphib_{k, h,s'} \varphib_{k, h,s'}^\top$.
    By taking expectation over $y^k_h$, we have
    \begin{equation} \label{eq:online concentration eq 1}
        \bar{\ell}_{k,h}(\thetab^k_h)
        \le 
        \bar{\ell}_{k,h}(\thetab_h^*) 
        + \bar{\Gb}_{k,h}(\thetab^k_h)^\top (\thetab^k_h - \thetab_h^*)
        - \frac{\kappa}{2} (\thetab_h^* - \thetab^k_h)^\top \Wb_{k,h}  (\thetab_h^* - \thetab^k_h) \, .      
    \end{equation}

    On the other hand, for any $\thetab \in \RR^d$, since we have
    \begin{align*}
        & \bar{\ell}_{k,h}(\thetab) - \bar{\ell}_{k,h}(\thetab_h^*)
        \\
        & = - \sum_{s' \in \Scal_{k,h}} P_{\thetab^*_h}(s' \mid s^k_h, a^k_h) \log P_{\thetab}(s' \mid s^k_h, a^k_h) 
            + \sum_{s' \in \Scal_{k,h}} P_{\thetab_h^*}(s' \mid s^k_h, a^k_h) \log P_{\thetab_h^*}(s' \mid s^k_h, a^k_h)
        \\
        & = \sum_{s' \in \Scal_{k,h}} P_{\thetab^*_h}(s' \mid s^k_h, a^k_h) \left( \log P_{\thetab^*_h}(s' \mid s^k_h, a^k_h) - \log P_{\thetab}(s' \mid s^k_h, a^k_h) \right) 
        \\
        & = \sum_{s' \in \Scal_{k,h}} P_{\thetab^*_h}(s' \mid s^k_h, a^k_h) \log \frac{P_{\thetab^*_h}(s' \mid s^k_h, a^k_h)} {P_{\thetab}(s' \mid s^k_h, a^k_h)} 
        \\
        & = D_{\text{KL}}(P_{\thetab^*_h} \parallel P_{\thetab})
        \\
        & \ge 0 \, , 
    \end{align*}
    where $D_{\text{KL}}(P \parallel Q)$ is the Kullback-Leibler divergence of $P$ from $Q$, from~\eqref{eq:online concentration eq 1} we have 
    \begin{align}
        0 & \le \bar{\ell}_{k,h}(\thetab^k_h) - \bar{\ell}_{k,h}(\thetab_h^*) \nonumber
        \\
        & \le \bar{\Gb}_{k,h}(\thetab^k_h)^\top (\thetab^k_h - \thetab_h^*)
            - \frac{\kappa}{2} \| \thetab_h^* - \thetab^k_h \|^2_{\Wb_{k,h}} \nonumber
        \\
        & = \Gb_{k,h}(\thetab^k_h)^\top (\thetab^k_h - \thetab_h^*)
            - \frac{\kappa}{2} \| \thetab_h^* - \thetab^k_h \|^2_{\Wb_{k,h}}
            + \left( \bar{\Gb}_{k,h}(\thetab^k_h) - \Gb_{k,h}(\thetab^k_h) \right)^\top (\thetab^k_h - \thetab_h^*) \, . \label{eq:online concentration eq 5}
    \end{align}    
    To get an upper bound of $\Gb_{k,h}(\thetab^k_h)^\top (\thetab^k_h - \thetab_h^*)$, recall that the estimated transition core is given by
    \begin{equation} \label{eq:online concentration eq 2}
        \thetab^{k+1}_h = \argmin_{\thetab \in \Bcal_d(\Pnorm)} \frac{1}{2} \| \thetab - \thetab^k_h \|^2_{\Ab_{k+1, h}} + (\thetab - \thetab^k_h)^\top \Gb_{k,h}(\thetab^k_h) \, .
    \end{equation}
    Since the objective function in~\eqref{eq:online concentration eq 2} is convex, by the first-order optimality condition for any $\thetab \in \Bcal_d(\Pnorm)$, we have
    \begin{equation*}
        \left( \Gb_{k,h}(\thetab^k_h) + \Ab_{k+1, h} ( \thetab^{k+1}_h - \thetab^k_h) \right)^\top (\thetab - \thetab^{k+1}_h) \ge 0, 
    \end{equation*}
    which gives
    \begin{equation} \label{eq:online concentration eq 3}
        \thetab^\top \Ab_{k+1, h} (\thetab^{k+1}_h - \thetab^k_h)
        \ge (\thetab^{k+1}_h)^\top \Ab_{k+1, h} (\thetab^{k+1}_h - \thetab^k_h) - \Gb_{k, h} (\thetab^k_h)^\top (\thetab - \thetab^{k+1}_h) \, .
    \end{equation}

    Then, we have
    \begin{align}
        & \| \thetab^k_h - \thetab^*_h \|^2_{\Ab_{k+1,h}} - \| \thetab^{k+1}_h - \thetab^*_h \|^2_{\Ab_{k+1,h}} \nonumber
        \\
        & = (\thetab^k_h)^\top \Ab_{k+1,h} \thetab^k_h - (\thetab^{k+1}_h)^\top \Ab_{k+1, h} \thetab^{k+1}_h
            + 2 (\thetab^*_h)^\top \Ab_{k+1, h} (\thetab^{k+1}_h - \thetab^k_h) \nonumber
        \\
        & \ge (\thetab^k_h)^\top \Ab_{k+1,h} \thetab^k_h - (\thetab^{k+1}_h)^\top \Ab_{k+1, h} \thetab^{k+1}_h
            + 2 (\thetab^{k+1}_h)^\top \Ab_{k+1, h} (\thetab^{k+1}_h - \thetab^k_h) \nonumber
            \\
            & \quad 
            - 2 \Gb_{k, h} (\thetab^k_h)^\top (\thetab^*_h - \thetab^{k+1}_h) \tag{by~\eqref{eq:online concentration eq 3}}
        \\
        & = (\thetab^k_h)^\top \Ab_{k+1,h} \thetab^k_h + (\thetab^{k+1}_h)^\top \Ab_{k+1, h} \thetab^{k+1}_h
            - 2 (\thetab^{k+1}_h)^\top \Ab_{k+1, h} \thetab^k_h 
            - 2 \Gb_{k, h} (\thetab^k_h)^\top (\thetab^*_h - \thetab^{k+1}_h) \nonumber
        \\
        & = \| \thetab^k_h - \thetab^{k+1}_h \|^2_{\Ab_{k+1,h}}
            - 2 \Gb_{k, h} (\thetab^k_h)^\top (\thetab^*_h - \thetab^{k+1}_h) \nonumber
        \\
        & = \| \thetab^k_h - \thetab^{k+1}_h \|^2_{\Ab_{k+1,h}}
            + 2 \Gb_{k, h} (\thetab^k_h)^\top (\thetab^{k+1}_h - \thetab^k_h) 
            + 2 \Gb_{k, h} (\thetab^k_h)^\top (\thetab^k_h - \thetab^*_h)  \nonumber
        \\
        & \ge - \| \Gb_{k,h}(\thetab^k_h) \|^2_{\Ab_{k+1,h}^{-1}}
            + 2 \Gb_{k, h}(\thetab^k_h)^\top (\thetab^k_h - \thetab^*_h) \, , \label{eq:online concentration eq 4}
    \end{align}
    where the last inequality follows by the fact that
    \begin{align*}
        \| \thetab^k_h - \thetab^{k+1}_h \|^2_{\Ab_{k+1,h}} + 2 \Gb_{k, h}(\thetab^k_h)^\top (\thetab^{k+1}_h - \thetab^k_h) 
        & \ge \min_{\thetab \in \Bcal_d(\Pnorm)} \left\{ \| \thetab \|^2_{\Ab_{k+1,h}} + 2 \Gb_{k,h}(\thetab^k_h)^\top \thetab \right\}
        \\
        & = - \| \Gb_{k,h}(\thetab^k_h) \|^2_{\Ab^{-1}_{k+1,h}} \, .
    \end{align*}
    Therefore, from~\eqref{eq:online concentration eq 4} we have
    \begin{equation} \label{eq:online concentration eq 6}
        \Gb_{k, h}(\thetab^k_h)^\top (\thetab^k_h - \thetab^*_h)
        \le \frac{1}{2} \| \Gb_{k,h}(\thetab^k_h) \|^2_{\Ab^{-1}_{k+1,h}} 
            + \frac{1}{2} \| \thetab^k_h - \thetab^*_h \|^2_{\Ab_{k+1,h}}
            - \frac{1}{2} \| \thetab^{k+1}_h - \thetab^*_h \|^2_{\Ab_{k+1,h}} \, .
    \end{equation}    

    By substituting~\eqref{eq:online concentration eq 6} into~\eqref{eq:online concentration eq 5}, we have
    \begin{align}
        0 & \le \frac{1}{2} \| \Gb_{k,h}(\thetab^k_h) \|^2_{\Ab^{-1}_{k+1,h}} 
            + \frac{1}{2} \| \thetab^k_h - \thetab^*_h \|^2_{\Ab_{k+1,h}}
            - \frac{1}{2} \| \thetab^{k+1}_h - \thetab^*_h \|^2_{\Ab_{k+1,h}}
        \nonumber
        \\
        & \quad - \frac{\kappa}{2} \| \thetab_h^* - \thetab^k_h \|_{\Wb_{k,h}}
            + \left( \bar{\Gb}_{k,h}(\thetab^k_h) - \Gb_{k,h}(\thetab^k_h) \right)^\top (\thetab^k_h - \thetab_h^*) \, . \label{eq:online concentration eq 8} 
    \end{align}
    Note that since we have
    \begin{align}
        & \| \Gb_{k,h}(\thetab^k_h) \|^2_{\Ab^{-1}_{k+1,h}}
        \nonumber
        \\
        & = \sum_{s', \tilde{s} \in \Scal_{k,h}} \left( P_{\thetab^k_h}(s' \mid s^k_h, a^k_h) - y^k_h(s') \right) \left( P_{\thetab^k_h}(\tilde{s} \mid s^k_h, a^k_h) - y^k_h(\tilde{s}) \right) \varphib_{k, h,s'}^\top \Ab^{-1}_{k+1,h} \varphib_{k, h,\tilde{s}} 
        \nonumber        
        \\
        & = \frac{1}{2} \sum_{s', \tilde{s} \in \Scal_{k,h}} \left( P_{\thetab^k_h}(s' \mid s^k_h, a^k_h) - y^k_h(s') \right) \left( P_{\thetab^k_h}(\tilde{s} \mid s^k_h, a^k_h) - y^k_h(\tilde{s}) \right) \nonumber
        \\
        & \phantom{{}={}={}={}={}={}} \cdot ( \varphib_{k, h,s'}^\top \Ab^{-1}_{k+1,h} \varphib_{k, h,\tilde{s}} + \varphib_{k, h,\tilde{s}}^\top \Ab^{-1}_{k+1,h} \varphib_{k, h,s'} )
        \nonumber        
        \\
        & \le \frac{1}{2} \sum_{s', \tilde{s} \in \Scal_{k,h}} \bigg[ \left( P_{\thetab^k_h}(s' \mid s^k_h, a^k_h) - y^k_h(s') \right)^2 \varphib_{k, h,s'}^\top \Ab_{k+1,h}^{-1} \varphib_{k, h,s'} \nonumber
        \\
        &\phantom{{}={}={}={}={}={}}+ \left( P_{\thetab^k_h}(\tilde{s} \mid s^k_h, a^k_h) - y^k_h(\tilde{s}) \right)^2 \varphib_{k, h,\tilde{s}}^\top \Ab_{k+1,h}^{-1} \varphib_{k, h,\tilde{s}} \bigg]
        \nonumber        
        \\
        & = \sum_{s' \in \Scal_{k,h}} \left( P_{\thetab^k_h}(s' \mid s^k_h, a^k_h) - y^k_h(s') \right)^2 \varphib_{k, h,s'}^\top \Ab_{k+1,h}^{-1} \varphib_{k, h,s'}
        \nonumber
        \\
        & \le \sum_{s' \in \Scal_{k,h}} \left| P_{\thetab^k_h}(\tilde{s} \mid s^k_h, a^k_h) - y^k_h(\tilde{s}) \right| \varphib_{k, h, s'}^\top \Ab_{k+1,h}^{-1} \varphib_{k, h,s'}
        \nonumber
        \\
        & \le \sum_{s' \in \Scal_{k,h}} \left( P_{\thetab^k_h}(\tilde{s} \mid s^k_h, a^k_h) + y^k_h(s') \right) \varphib_{k, h, s'}^\top \Ab_{k+1,h}^{-1} \varphib_{k, h,s'}
        \nonumber
        \\
        & = \sum_{s' \in \Scal_{k,h}} P_{\thetab^k_h}(\tilde{s} \mid s^k_h, a^k_h) \varphib_{k, h, s'}^\top \Ab_{k+1,h}^{-1} \varphib_{k, h,s'} 
            + \sum_{s' \in \Scal_{k,h}} y^k_h(s') \varphib_{k, h, s'}^\top \Ab_{k+1,h}^{-1} \varphib_{k, h,s'}
        \nonumber
        \\
        & \le 2 \max_{s' \in \Scal_{k,h}} \| \varphib_{k, h,s'} \|^2_{\Ab_{k+1,h}^{-1}} \, , \label{eq:online concentration eq 7}
    \end{align}
    where the first inequality utilizes the inequality $\xb^\top \Ab \yb + \yb^\top \Ab \xb \le \xb^\top \Ab \xb + \yb^\top \Ab \yb$ for any positive-semidefinite matrix $\Ab$, and the last inequality holds since $0 \le P_{\thetab^k_h}(s' \mid s^k_h, a^k_h) \le 1$ and $\sum_{s'} P_{\thetab^k_h}(s' \mid s^k_h, a^k_h) = 1$.        

    Combining the results of~\eqref{eq:online concentration eq 8} and~\eqref{eq:online concentration eq 7}, we have
    \begin{align}
        0 & \le \max_{s' \in \Scal_{k,h}} \| \varphib_{k, h,s'} \|_{\Ab_{k+1,h}^{-1}}^2
            + \frac{1}{2} \| \thetab^k_h - \thetab^*_h \|^2_{\Ab_{k+1,h}}
            - \frac{1}{2} \| \thetab^{k+1}_h - \thetab^*_h \|^2_{\Ab_{k+1,h}}
        \nonumber
        \\
        & \quad - \frac{\kappa}{2} \| \thetab_h^* - \thetab^k_h \|_{\Wb_{k,h}}
            + \left( \bar{\Gb}_{k,h}(\thetab^k_h) - \Gb_{k,h}(\thetab^k_h) \right)^\top (\thetab^k_h - \thetab_h^*)
        \nonumber
        \\
        & = \max_{s' \in \Scal_{k,h}} \| \varphib_{k, h,s'} \|_{\Ab_{k+1,h}^{-1}}^2
            + \frac{1}{2} \| \thetab^k_h - \thetab^*_h \|^2_{\Ab_{k,h}}
            + \frac{\kappa}{4} \| \thetab^k_h - \thetab^*_h \|^2_{\Wb_{k,h}}
            - \frac{1}{2} \| \thetab^{k+1}_h - \thetab^*_h \|^2_{\Ab_{k+1,h}}
        \nonumber
        \\
        & \quad - \frac{\kappa}{2} \| \thetab_h^* - \thetab^k_h \|_{\Wb_{k,h}}
            + \left( \bar{\Gb}_{k,h}(\thetab^k_h) - \Gb_{k,h}(\thetab^k_h) \right)^\top (\thetab^k_h - \thetab_h^*)
        \nonumber
        \\
        & = \max_{s' \in \Scal_{k,h}} \| \varphib_{k, h,s'} \|_{\Ab_{k+1,h}^{-1}}^2
            + \frac{1}{2} \| \thetab^k_h - \thetab^*_h \|^2_{\Ab_{k,h}}
            - \frac{\kappa}{4} \| \thetab^k_h - \thetab^*_h \|^2_{\Wb_{k,h}}
            - \frac{1}{2} \| \thetab^{k+1}_h - \thetab^*_h \|^2_{\Ab_{k+1,h}}
        \nonumber
        \\
        & \quad + \left( \bar{\Gb}_{k,h}(\thetab^k_h) - \Gb_{k,h}(\thetab^k_h) \right)^\top (\thetab^k_h - \thetab_h^*) \, ,
        \nonumber        
    \end{align}
    where for the first equality we use $\Ab_{k+1, h} = \Ab_{k,h} + \frac{\kappa}{2} \Wb_{k,h}$.
    By rearranging the terms, we have 
    \begin{align*}
        \| \thetab^{k+1}_h - \thetab^*_h \|^2_{\Ab_{k+1,h}}
        & \le \| \thetab^k_h - \thetab^*_h \|^2_{\Ab_{k,h}}
            + 2 \max_{s' \in \Scal_{k,h}} \| \varphib_{k, h,s'} \|_{\Ab_{k+1,h}^{-1}}^2
            - \frac{\kappa}{2} \| \thetab^k_h - \thetab^*_h \|^2_{\Wb_{k,h}}
        \\
        & \quad + 2 \left( \bar{\Gb}_{k,h}(\thetab^k_h) - \Gb_{k,h}(\thetab^k_h) \right)^\top (\thetab^k_h - \thetab_h^*) \, .
    \end{align*}
    Then summing over $k$ gives 
    \begin{align*}
        \| \thetab^{k+1}_h - \thetab^*_h \|^2_{\Ab_{k+1,h}}
        & \le \| \thetab_{1,h} - \thetab^*_h \|^2_{\Ab_{1,h}}
            + 2 \sum_{i=1}^k \max_{s' \in \Scal_{i,h}} \| \varphib_{i,h,s'} \|_{\Ab_{i+1,h}^{-1}}^2
            - \frac{\kappa}{2} \sum_{i=1}^k \| \thetab^{i}_h - \thetab^*_h \|^2_{\Wb_{i,h}}
        \\
        & \quad + 2 \sum_{i=1}^k \left( \bar{\Gb}_{i,h}(\thetab^{i}_h) - \Gb_{i,h}(\thetab^{i}_h) \right)^\top (\thetab^{i}_h - \thetab_h^*)
        \\
        & \le 2 \lambda L^2_{\theta}
            + 2 \sum_{i=1}^k \max_{s' \in \Scal_{i,h}} \| \varphib_{i,h,s'} \|_{\Ab_{i+1,h}^{-1}}^2
            - \frac{\kappa}{2} \sum_{i=1}^k \| \thetab^{i}_h - \thetab^*_h \|^2_{\Wb_{i,h}}
        \\
        & \quad + 2 \sum_{i=1}^k \left( \bar{\Gb}_{i,h}(\thetab^{i}_h) - \Gb_{i,h}(\thetab^{i}_h) \right)^\top (\thetab^{i}_h - \thetab_h^*) \, .
    \end{align*}    

    For the final step, note that $\left( \bar{\Gb}_{i,h}(\thetab^{i}_h) - \Gb_{i,h}(\thetab^{i}_h) \right)^\top (\thetab^{i}_h - \thetab_h^*)$ is a martingale difference sequence. 
    To bound this term, we invoke the following lemmas:

    \begin{lemma} \label{lemma:mds term upper bound}
        For $\delta \in (0,1)$ and $(k,h) \in [K] \times [H]$, with a probability at least $1 - \delta$ we have
        \begin{equation*}
            \begin{split}
                & \sum_{i=1}^k \left( \bar{\Gb}_{i,h} (\thetab^{i}_h) - \Gb_{i, h}(\thetab^{i}_h) \right)^\top (\thetab^{i}_h - \thetab^*)
                \\
                & \le 
                \frac{\kappa}{4} \sum_{i=1}^k \| \thetab^{i}_h - \thetab^*_h \|^2_{\Wb_{i,h}} + \left( \frac{16 \Fnorm \Pnorm}{3} + \frac{8}{\kappa} \right) \log \frac{ \left( 1 + \lceil 2 \log_2 k \Ucal \Fnorm \Pnorm \rceil \right) k^2}{\delta} + \sqrt{2} \, .    
            \end{split}            
        \end{equation*}
    \end{lemma}

    \begin{lemma}[Generalized elliptical potential] \label{lemma:generalized elliptical potential}
        Let $S_t := \{ \xb_{t,1}, \ldots, \xb_{t,K} \} \subset \RR^d$. 
        For any $1 \le t \le T$ and $i \in [K]$, suppose $\| \xb_{t,i} \|_2 \le L$.
        Let $\Vb_t := \lambda \Ib_d + \sum_{\tau=1}^{t-1} \sum_{i \in S_\tau} \xb_{\tau, i} \xb_{\tau, i}^\top$ for some $\lambda > 0$.
        If $\lambda \ge L^2$, then we have
        \begin{align*}
            \sum_{t=1}^T \max_{i \in [K]} \| \xb_{t,i} \|_{\Vb_{t}^{-1}}^2 \le 2 d \log \left( 1 + \frac{T K L}{ d \lambda } \right) \, .
        \end{align*}
    \end{lemma}
    
    By Lemma~\ref{lemma:mds term upper bound}, with probability at least $1 - \delta$, we have
    \begin{align*}
        & \| \thetab^{k+1}_h - \thetab^*_h \|^2_{\Ab_{k+1,h}}
        \\
        & \le 2 \lambda \Pnorm^2
            + 2 \sum_{i=1}^k \max_{s' \in \Scal_{i,h}} \| \varphib_{i,h,s'} \|_{\Ab_{i+1,h}^{-1}}^2
            \\
            &\phantom{{}={}} + \left( \frac{32 \Fnorm \Pnorm}{3} + \frac{16}{\kappa} \right) \log \frac{ \left( 1 + \lceil 2 \log_2 k \Ucal \Fnorm \Pnorm \rceil \right) k^2}{\delta} + 2 \sqrt{2}
        \\
        & \le 2 \lambda \Pnorm^2
            + \frac{8}{\kappa} d \log \left( 1 + \frac{k \Ucal \Fnorm^2}{d \lambda} \right)
            + \left( \frac{32 \Fnorm \Pnorm}{3} + \frac{16}{\kappa} \right) \log \frac{ \left( 1 + \lceil 2 \log_2 k \Ucal \Fnorm \Pnorm \rceil \right) k^2}{\delta}  + 2 \sqrt{2} \, ,
    \end{align*}
    where the second inequality comes from Lemma~\ref{lemma:generalized elliptical potential}. Note that the Gram matrix $\Ab_{k,h}$ in Algorithm~\ref{alg:Algorithm 1} and the Gram matrix $\Vb$ in Lemma~\ref{lemma:generalized elliptical potential} are different by the factor of $\frac{\kappa}{2}$, which results in additional $\frac{2}{\kappa}$ factor for the bound of $\sum_{i=1}^k \max_{s' \in \Scal_{i,h}} \| \varphib_{i, h, s'} \|_{\Ab_{i+1, h}^{-1}}^2$.
\end{proof}

In the following, we provide all the proofs of the lemmas used to prove Lemma~\ref{lemma:concentration of online estimated transition core}.

\subsubsection{Proof of Lemma~\ref{lemma:mds term upper bound}}
\begin{proof}[Proof of Lemma~\ref{lemma:mds term upper bound}]
    Note that $\left( \bar{\Gb}_{i,h} (\thetab^{i}_h) - \Gb_{i, h}(\thetab^{i}_h) \right)^\top (\thetab^{i}_h - \thetab^*_h)$ is a martingale difference sequence, i.e., 
    \begin{align*}
        & \EE \left[ \left( \bar{\Gb}_{i,h} (\thetab^{i}_h) - \Gb_{i, h}(\thetab^{i}_h) \right)^\top (\thetab^{i}_h - \thetab^*_h) \mid \Fcal_{i,h} \right]
        \\
        & = \left( \bar{\Gb}_{i,h} (\thetab^{i}_h) - \EE \left[ \Gb_{i, h}(\thetab^{i}_h) \mid \Fcal_{i,h} \right] \right)^\top (\thetab^{i}_h - \thetab^*_h) 
        \\
        & = 0 \, .
    \end{align*}
    On the other hand, for any $\thetab \in \RR^d$, since we have
    \begin{align*}
        \| \Gb_{i, h} (\thetab) \|_2
         & = \left\| \sum_{s' \in \Scal_{i,h}} \left( P_{\thetab}(s' \mid s^{i}_h, a^{i}_h) - y^{i}_h(s') \right) \varphib_{i, h, s'} \right\|_2
         \\
         & \le \sum_{s' \in \Scal_{i,h}} \left| P_{\thetab}(s' \mid s^{i}_h, a^{i}_h) - y^{i}_h(s') \right| \| \varphib_{i,h,s'} \|_2
         \\
         & \le \Fnorm \left( \sum_{s' \in \Scal_{i,h}} P_{\thetab}(s' \mid s^{i}_h, a^{i}_h) + \sum_{s' \in \Scal_{i,h}}  y^{i}_h(s') \right)
         \\
         & = 2 \Fnorm \, ,
    \end{align*}
    then, it follows by
    \begin{align}
        & \left| \left( \bar{\Gb}_{i,h} (\thetab^{i}_h) - \Gb_{i, h}(\thetab^{i}_h) \right)^\top (\thetab^{i}_h - \thetab^*_h) \right|
        \nonumber
        \\
        & \le \left| \left( \bar{\Gb}_{i,h} (\thetab^{i}_h) \right)^\top  (\thetab^{i}_h - \thetab^*_h) \right|
            + \left| \left( \Gb_{i,h} (\thetab^{i}_h) \right)^\top (\thetab^{i}_h - \thetab^*_h) \right|
        \nonumber
        \\
        & \le \| \bar{\Gb}_{i,h} (\thetab^{i}_h) \|_2 \| \thetab^{i}_h - \thetab^*_h \|_2 
            + \| \Gb_{i,h} (\thetab^{i}_h) \|_2 \| \thetab^{i}_h - \thetab^*_h \|_2 
        \nonumber
        \\
        & \le 4 \Fnorm \| \thetab^{i}_h - \thetab^*_h \|_2 
        \nonumber
        \\
        & \le 8 \Fnorm \Pnorm \, , \label{eq 4 lemma:mds term upper bound}
    \end{align}
    where the last inequality follows by $\| \thetab^{i}_h - \thetab^*_h \|_2 \le \| \thetab^{i}_h \|_2 + \| \thetab^*_h \|_2  \le 2 \Pnorm$. 
    Hence, if we denote $M_{k,h} := \sum_{i=1}^k \left( \bar{\Gb}_{i,h} (\thetab^{i}_h) - \Gb_{i, h}(\thetab^{i}_h) \right)^\top (\thetab^{i}_h - \thetab^*_h)$, then $M_{k,h}$ is a martingale.
    Note that we also have
    \begin{align}
        \Sigma_{k,h}
        & = \sum_{i=1}^k \EE_{y^{i}_h} \left[ \left( \left[ \bar{\Gb}_{i,h} (\thetab^{i}_h) - \Gb_{i, h}(\thetab^{i}_h) \right]^\top (\thetab^{i}_h - \thetab^*_h) \right)^2 \right]
            \nonumber
        \\
        & = \sum_{i=1}^k \EE_{y^{i}_h} \Bigg[ \left( \left[ \Gb_{i,h} (\thetab^{i}_h) \right]^\top (\thetab^{i}_h - \thetab^*_h) \right)^2 \Bigg]
            - \EE_{y^{i}_h} \Bigg[ \left( \left[ \bar{\Gb}_{i,h} (\thetab^{i}_h) \right]^\top (\thetab^{i}_h - \thetab^*_h) \right)^2  \Bigg]
        \nonumber
        \\
        & \le \sum_{i=1}^k \EE_{y^{i}_h} \left[ \left( \left[ \Gb_{i,h} (\thetab^{i}_h) \right]^\top (\thetab^{i}_h - \thetab^*_h) \right)^2 \right]
        \nonumber
        \\
        & = \sum_{i=1}^k \EE_{y^{i}_h} \left[ \left( \sum_{s' \in \Scal_{i,h}} \left( P_{\thetab^{i}_h}(s' \mid s^{i}_h, a^{i}_h) - y^{i}_h(s') \right) \varphib_{i, h, s'}^\top (\thetab^{i}_h - \thetab^*_h) \right)^2 \right]
        \nonumber
        \\
        &\leq \sum_{i=1}^k \EE_{y^{i}_h} \left[ 
        \left( \sum_{s' \in \Scal_{i,h}} 
        \left( P_{\thetab^{i}_h}(s' \mid s^{i}_h, a^{i}_h) - y^{i}_h(s') \right)^2 \right)
        \left( \sum_{s' \in \Scal_{i,h}} 
        \left(\varphib_{i, h, s'}^\top (\thetab^{i}_h - \thetab^*_h) \right)^2 \right) \right] 
        \label{eq 1 lemma:mds term upper bound}
        \\
        & = \sum_{i=1}^k 
        \EE_{y^{i}_h} \left[ \sum_{s' \in \Scal_{i,h}} 
        \left(  P_{\thetab^{i}_h}(s' \mid s^{i}_h, a^{i}_h) - y^{i}_h(s') \right)^2   \right] 
        \left( \sum_{s' \in \Scal_{i,h}} 
        \left(\varphib_{i, h, s'}^\top (\thetab^{i}_h - \thetab^*_h) \right)^2 \right)
        \nonumber
        \\
        & \le 2 \sum_{i=1}^k \sum_{s' \in \Scal_{i,h}} \left( \varphib_{i, h, s'}^\top (\thetab^{i}_h - \thetab^*_h) \right)^2 
        \label{eq 2 lemma:mds term upper bound}
        \\
        & = 2 \sum_{i=1}^k \| \thetab^{i}_h - \thetab^*_h \|^2_{\Wb_{i,h}} =: B_{k,h} \, , 
        \nonumber
    \end{align}
    where~\eqref{eq 1 lemma:mds term upper bound} holds by the Cauchy–Schwarz inequality,~\eqref{eq 2 lemma:mds term upper bound} holds because 
    \begin{align*}
        & \sum_{s' \in \Scal_{i,h}} 
        \left(  P_{\thetab^{i}_h}(s' \mid s^{i}_h, a^{i}_h) - y^{i}_h(s') \right)^2 
        \\
        & = \sum_{s' \in \Scal_{i,h}} \left\{P_{\thetab^{i}_h}(s' \mid s^{i}_h, a^{i}_h)\right\}^2 -2 P_{\thetab^{i}_h}(s' \mid s^{i}_h, a^{i}_h) y^{i}_h(s') + \left\{ y^{i}_h(s')\right\}^2
        \\
        & \leq 2 \, .
    \end{align*}
    However, if we denote $B_{k,h} := 2 \sum_{i=1}^k  \| \thetab^{i}_h - \thetab^*_h \|^2_{\Wb_{i,h}}$, since $B_{k,h}$ is itself a random variable, to apply Freedman's inequality to $M_{k,h}$, we consider two cases depending on the values of $B_{k,h}$.

    \textbf{Case 1 : } $B_{k,h} \le \frac{4}{k \Ucal}$

    Suppose that $B_{k, h} = 2 \sum_{i=1}^k \| \thetab^{i}_h - \thetab^*_h \|^2_{\Wb_{i,h}} \le \frac{4}{k \Ucal}$.
    Then we have
    \begin{align*}
        M_{k,h} 
        & = \sum_{i=1}^k \left( \bar{\Gb}_{i,h} (\thetab^{i}_h) - \Gb_{i, h}(\thetab^{i}_h) \right)^\top (\thetab^{i}_h - \thetab^*_h)
        \\
        & = \sum_{i=1}^k \sum_{s' \in \Scal_{i,h}} \left( y^{i}_h(s') - \EE [y^{i}_h(s')] \right) \varphib_{i, h, s'}^\top   (\thetab^{i}_h - \thetab^*_h)
        \\
        & = \sum_{i=1}^k \sum_{s' \in \Scal_{i,h}} \left( y^{i}_h(s') - P_{\thetab^*_h}(s' \mid s^{i}_h, a^{i}_h ) \right) \varphib_{i, h, s'}^\top   (\thetab^{i}_h - \thetab^*_h)
        \\
        & \le \sum_{i=1}^k \sum_{s' \in \Scal_{i,h}} | \varphib_{i, h, s'}^\top   (\thetab^{i}_h - \thetab^*_h) |
        \\
        & \le \sqrt{ k \Ucal \sum_{i=1}^k \sum_{s' \in \Scal_{i,h}} \left( \varphib_{i, h, s'}^\top   (\thetab^{i}_h - \thetab^*_h) \right)^2 }
        \\
        & = \sqrt{ k \Ucal \frac{B_{k,h}}{2} }
        \\
        & \le \sqrt{2} \, .
    \end{align*}

    \textbf{Case 2 : } $B_{k,h} > \frac{4}{k \Ucal }$

    Suppose that $B_{k, h} = 2 \sum_{i=1}^k \| \thetab^{i}_h - \thetab^*_h \|^2_{\Wb_{i,h}} > \frac{4}{k \Ucal}$. 
    Then, we have both a lower and upper bound for $B_{k,h}$ as follows:
    \begin{equation*}
        \frac{4}{k \Ucal} 
        < B_{k,h} 
        \le 2 \sum_{i=1}^k \sum_{s' \in \Scal_{i,h}}  \| \varphib_{i, h, s'} \|^2_2  \| \thetab^{i}_h - \thetab^*_h \|_2^2
        \le 8 k \Ucal \Fnorm^2 \Pnorm^2  \, .
    \end{equation*}
    Then by the peeling process from~\citet{bartlett2005local}, for any $\eta_k > 0$, we have
    \begin{align}
        & \PP \left(M_{k,h} \ge 2 \sqrt{\eta_k B_{k,h}} + \frac{16 \eta_k \Fnorm \Pnorm}{3} \right)
        \nonumber
        \\
        & = \PP \left(M_{k,h} \ge 2 \sqrt{\eta_k B_{k,h}} + \frac{16 \eta_k \Fnorm \Pnorm}{3}, \frac{4}{k \Ucal } < B_{k,h} \le 8 k \Ucal \Fnorm^2 \Pnorm^2 \right)
        \nonumber
        \\
        & = \PP \left(M_{k,h} \ge 2 \sqrt{\eta_k B_{k,h}} + \frac{16 \eta_k \Fnorm \Pnorm}{3}, \frac{4}{k \Ucal} < B_{k,h} \le 8 k \Ucal \Fnorm^2 \Pnorm^2, \Sigma_{k,h} \le B_{k,h} \right)
        \nonumber
        \\
        & \le \sum_{j=1}^m \PP \left(M_{k,h} \ge 2 \sqrt{\eta_k B_{k,h}} + \frac{16 \eta_k \Fnorm \Pnorm}{3}, \frac{4 \cdot 2^{j-1}}{k \Ucal} < B_{k,h} \le \frac{4 \cdot 2^j}{k \Ucal}, \Sigma_{k,h} \le B_{k,h} \right)
        \nonumber
        \\
        & \le \sum_{j=1}^m \underbrace{\PP \left(M_{k,h} \ge \sqrt{\eta_k \frac{8 \cdot 2^j}{k \Ucal}} + \frac{16 \eta_k \Fnorm \Pnorm}{3}, \Sigma_{k,h} \le \frac{4 \cdot 2^j}{k \Ucal} \right)}_{I_j} \, , \label{eq 3 lemma:mds term upper bound}
    \end{align}
    where $m = 1 + \lceil 2 \log_2 k \Ucal \Fnorm \Pnorm \rceil$.
    For $I_j$, note that from~\eqref{eq 4 lemma:mds term upper bound} we have
    \begin{equation*}
        \left| \left( \bar{\Gb}_{i,h} (\thetab^{i}_h) - \Gb_{i, h}(\thetab^{i}_h) \right)^\top (\thetab^{i}_h - \thetab^*_h) \right| \le 8 \Fnorm \Pnorm \, .
    \end{equation*}
    By Freedman's inequality (\Cref{lemma:freedman ineq}), we have
    \begin{align}
        & \PP \left( M_{k,h} \ge \sqrt{\eta_k \frac{8 \cdot 2^j}{k \Ucal }} + \frac{16 \eta_k \Fnorm \Pnorm}{3}, \Sigma_{k,h} \le \frac{4 \cdot 2^j}{k \Ucal} \right)
        \nonumber
        \\
        & \le \exp \left( \frac{- \left( \sqrt{\eta_k \frac{8 \cdot 2^j}{k \Ucal}} + \frac{16 \eta_k \Fnorm \Pnorm}{3} \right)^2}{\frac{8 \cdot 2^j}{k \Ucal} + \frac{2}{3} \cdot 8 \Fnorm \Pnorm \left( \sqrt{\eta_k \frac{8 \cdot 2^j}{k \Ucal}} + \frac{16 \eta_k \Fnorm \Pnorm}{3} \right)}
        \right)
        \nonumber
        \\
        & = \exp \left( \frac{- \eta_k \left( \sqrt{\frac{8 \cdot 2^j}{k \Ucal }} + \frac{16 \sqrt{\eta_k} \Fnorm \Pnorm}{3} \right)^2}{\frac{8 \cdot 2^j}{k \Ucal } + \frac{16 \Fnorm \Pnorm}{3} \sqrt{\eta_k \frac{8 \cdot 2^j}{k \Ucal }} + \frac{16^2 \eta_k \Fnorm^2 \Pnorm^2}{3^2}}
        \right)
        \nonumber
        \\
        & \le \exp \left( \frac{- \eta_k \left( \sqrt{\frac{8 \cdot 2^j}{k \Ucal}} + \frac{16 \sqrt{\eta_k} \Fnorm \Pnorm}{3} \right)^2}{\frac{8 \cdot 2^j}{k \Ucal } + \frac{32 \Fnorm \Pnorm}{3} \sqrt{\eta_k \frac{8 \cdot 2^j}{k \Ucal }} + \frac{16^2 \eta_k \Fnorm^2 \Pnorm^2}{3^2}}
        \right)
        \nonumber
        \\
        & = \exp(-\eta_k) \, .
        \label{eq 5 lemma:mds term upper bound}
    \end{align}
    By substituting Eq.~\eqref{eq 5 lemma:mds term upper bound} into Eq.~\eqref{eq 3 lemma:mds term upper bound}, we have
    \begin{align*}
        \PP \left( M_{k,h} \ge 2 \sqrt{\eta_k B_{k,h}} + \frac{16 \eta_k \Fnorm \Pnorm}{3} \right) \le m \exp(-\eta_k) \, .
    \end{align*}
    Then, combining with the result of Case 1 \& 2, letting $\eta_k = \log \frac{m}{\delta / k^2} = \log \frac{ \left( 1 + \lceil 2 \log_2 k \Ucal \Fnorm \Pnorm \rceil \right) k^2}{\delta}$ and taking union bound over $k$, with probability at least $1 - \delta$, we have
    \begin{equation} \label{eq 6 lemma:mds term upper bound}
        M_{k,h} \le 2 \sqrt{2 \eta_k \sum_{i=1}^k \| \thetab^{i}_h - \thetab^*_h \|^2_{\Wb_{i,h}}} + \frac{16 \eta_k \Fnorm \Pnorm}{3} + \sqrt{2} \, .
    \end{equation}
    By applying $2 \sqrt{ab} \le a+b$ to the first term on the right hand side, we have
    \begin{equation} \label{eq 7 lemma:mds term upper bound}
        2 \sqrt{2 \eta_k \sum_{i=1}^k \| \thetab^{i}_h - \thetab^*_h \|^2_{\Wb_{i,h}}}
        \le  \frac{8 \eta_k}{\kappa} + \frac{\kappa}{4} \sum_{i=1}^k \| \thetab^{i}_h - \thetab^*_h \|^2_{\Wb_{i,h}} \, .
    \end{equation}
    Combining the results of Eq.~\eqref{eq 6 lemma:mds term upper bound} \& Eq.~\eqref{eq 7 lemma:mds term upper bound}, we have
    \begin{align*}
        M_{k,h}
        & = \sum_{i=1}^k \left( \bar{\Gb}_{i,h} (\thetab^{i}_h) - \Gb_{i, h}(\thetab^{i}_h) \right)^\top (\thetab^{i}_h - \thetab^*_h)
        \\
        & \le \frac{\kappa}{4} \sum_{i=1}^k \| \thetab^{i}_h - \thetab^*_h \|^2_{\Wb_{i,h}} + \left( \frac{16 \Fnorm \Pnorm}{3} + \frac{8}{\kappa} \right) \log \frac{ \left( 1 + \lceil 2 \log_2 k \Ucal \Fnorm \Pnorm \rceil \right) k^2}{\delta}  + \sqrt{2} \, .
    \end{align*}    
\end{proof}

\subsubsection{Proof of Lemma~\ref{lemma:generalized elliptical potential}}
\begin{proof}[Proof of Lemma~\ref{lemma:generalized elliptical potential}]
    By definition of $\Vb_t$, we have
    \begin{align}
        \det (\Vb_{t+1})
        & = \det \left( \Vb_{t} + \sum_{i \in S_t} \xb_{t,i} \xb_{t,i}^\top \right)
        \nonumber
        \\
        &= \det (\Vb_t) \det \left( \Ib_d + \sum_{i \in S_t} \Vb_{t}^{-\frac{1}{2}} \xb_{t,i} \xb_{t,i}^\top \Vb_{t}^{-\frac{1}{2}} \right)
        \nonumber
        \\
        & = \det (\Vb_{t}) \left( 1 + \sum_{i \in S_t} \| \xb_{t,i} \|^2_{\Vb_{t}^{-1}} \right)
        \nonumber
        \\
        & = \det (\lambda \Ib_d) \prod_{\tau=1}^{t} \left( 1 + \sum_{i \in S_\tau} \| \xb_{\tau,i} \|^2_{\Vb_{\tau}^{-1}} \right)
        \nonumber
        \\
        & \ge \det (\lambda \Ib_d) \prod^t_{\tau=1} \left( 1 + \max_{i \in S_\tau} \| \xb_{\tau, i} \|^2_{\Vb_{t}^{-1}} \right) \, .
        \label{eq:generalized elliptical potential 1}
    \end{align}
    
    Since $\lambda \ge L^2$, we have
    \begin{equation*}
        \max_{i \in S_\tau} \| \xb_{\tau, i} \|^2_{\Vb_{\tau}^{-1}} \le \frac{L^2}{\lambda} \le 1 \, .
    \end{equation*}
    
    Since for any $z \in [0,1]$, it follows that $z \le 2 \log(1+z)$. 
    Hence, we have
    \begin{align*}
        \sum^T_{t=1} \max_{i \in S_t} \| \xb_{t,i} \|^2_{\Vb_{t}^{-1}}  
        & \le 2 \sum^T_{t=1} \log \left( 1 + \max_{i \in S_t} \| \xb_{t,i} \|^2_{\Vb_{t}^{-1}} \right)
        \\
        & = 2 \log \prod^T_{t=1} \left( 1 + \max_{i \in S_t} \| \xb_{t,i} \|^2_{\Vb_{t}^{-1}} \right)
        \\
        & \le 2 \log \frac{ \det (\Vb_{T+1}) }{ \det (\lambda \Ib_d )}
        \\
        & \le 2d \log \left( 1 + \frac{T K L^2 }{d \lambda} \right) \, ,
    \end{align*}
    where the second inequality comes from Eq.~\eqref{eq:generalized elliptical potential 1} and the last inequality follows by the determinant-trace inequality (Lemma~\ref{lemma:det-trace ineq}). 
\end{proof}

\subsection{Bound on Prediction Error} 
In this section, we provide the bound on the prediction error induced by estimated transition core $\thetab^k_h$.
\begin{lemma}[Bound on Prediction Error] \label{lemma:bound of prediction error}
    For any $\delta \in (0,1)$, suppose that Lemma~\ref{lemma:concentration of online estimated transition core} holds.
    Then for any $(s,a) \in \Scal \times \Acal$, we have 
    \begin{equation*}
            | \Delta^k_h(s,a) |
            \le H \alpha_k(\delta) \| \hat{\varphib}_{k,h}(s,a) \|_{\Ab_{k,h}^{-1}} \, .
    \end{equation*}
\end{lemma}

\begin{proof}[Proof of Lemma~\ref{lemma:bound of prediction error}]
    Recall that 
    \begin{align*}
        \Delta^k_h(s,a) 
        & = \sum_{s' \in \Scal_{s,a}} \left( P_{\thetab^k_h}(s' \mid s, a) - P_{\thetab^*_h}(s' \mid s, a) \right) V^k_{h+1}(s')
        \\
        & = \sum_{s' \in \Scal_{s,a}} \frac{\exp(\varphib_{s,a,s'}^\top \thetab^k_h) V^k_{h+1}(s')}{\sum_{\tilde{s} \in \Scal_{s,a}} \exp(\varphib_{s,a,\tilde{s}}^\top \, \thetab^k_h)}
        - \sum_{s' \in \Scal_{s,a}} \frac{\exp(\varphib_{s,a,s'}^\top \thetab^*_h) V^k_{h+1}(s')}{\sum_{\tilde{s} \in \Scal_{s,a}} \exp(\varphib_{s,a,\tilde{s}}^\top \, \thetab^*_h)} \, .
    \end{align*}

    Then by the mean value theorem, there exists $\bar{\thetab} = \rho \thetab^k_h + (1 - \rho) \thetab^*_h$ for some $\rho \in [0,1]$ satisfying that
    \begin{align*}
        \Delta^k_h(s,a)
        & = \frac{ \left( \sum_{s' \in \Scal_{s,a}}  \exp( \varphib_{s,a,s'}^\top \bar{\thetab}) V^k_{h+1}(s') \varphib_{s,a,s'}^\top (\thetab^k_h - \thetab^*_h) \right) \left( \sum_{\tilde{s} \in \Scal_{s,a}} \exp( \varphib_{s,a,\tilde{s}}^\top \bar{\thetab}) \right) }{ \left( \sum_{ \tilde{s} \in \Scal_{s,a}} \exp( \varphib_{s,a,\tilde{s}}^\top \, \bar{\thetab}) \right)^2}
            \\
            & \phantom{{}={}}
            - \frac{ \left( \sum_{s' \in \Scal_{s,a}} \exp( \varphib_{s,a,s'}^\top \bar{\thetab}) V^k_{h+1}(s') \right) \left( \sum_{\tilde{s} \in \Scal_{s,a}} \exp( \varphib_{s,a,\tilde{s}}^\top \, \bar{\thetab}) \varphib_{s,a,\tilde{s}}^\top \, (\thetab^k_h - \thetab^*_h) \right)}{ \left( \sum_{ \tilde{s} \in \Scal_{s,a}} \exp( \varphib_{s,a,\tilde{s}}^\top \, \bar{\thetab}) \right)^2}
        \\
        & = \sum_{s' \in \Scal_{s,a}} P_{\bar{\thetab}}(s' \mid s, a) V^k_{h+1}(s') \varphib_{s,a,s'}^\top (\thetab^k_h - \thetab^*_h)
            \\
            & \phantom{{}={}}
            - \left(  \frac{\sum_{s' \in \Scal_{s,a}} \exp( \varphib_{s,a,s'}^\top \bar{\thetab}) V^k_{h+1}(s')}{\sum_{ \tilde{s} \in \Scal_{k,h}} \exp( \varphib_{s,a,\tilde{s}}^\top \, \bar{\thetab})} \right)
            \sum_{s' \in \Scal_{s,a}} P_{\bar{\thetab}}(s' \mid s, a) \varphib_{s,a,s'}^\top (\thetab^k_h - \thetab^*_h)
        \\
        & = \sum_{s' \in \Scal_{s,a}} \left( V^k_{h+1}(s') -  \frac{\sum_{s' \in \Scal_{s,a}} \exp( \varphib_{s,a,s'}^\top \bar{\thetab}) V^k_{h+1}(s')}{\sum_{\tilde{s} \in \Scal_{s,a}} \exp( \varphib_{s,a,\tilde{s}}^\top \, \bar{\thetab})} \right)
        P_{\bar{\thetab}}(s' \mid s, a) \varphib_{s,a,s'}^\top (\thetab^k_h - \thetab^*_h) \, .
    \end{align*}
    
    Since $V^k_h(s') \le H$ for all $s' \in \Scal, k \in [K]$, and $h \in [H]$, we have
        \begin{align*}
            \Delta^k_h(s,a)
            & \le H \sum_{s' \in \Scal_{s,a}} P_{\bar{\thetab}}(s' \mid s, a) \varphib_{s,a,s'}^\top (\thetab^k_h - \thetab^*_h)
            \\
            & \le H \max_{s' \in \Scal_{s,a}} | \varphib_{s,a,s'}^\top (\thetab^k_h - \thetab^*_h) |
            \\
            & \le H \max_{s' \in \Scal_{s,a}} \| \varphib_{s,a,s'}\|_{\Ab_{k,h}^{-1}} \| \thetab^k_h - \thetab^*_h \|_{\Ab_{k,h}}
            \\
            & \le H \alpha_k(\delta) \| \hat{\varphib}_{k,h}(s,a) \|_{\Ab_{k,h}^{-1}} \, ,
        \end{align*}
    where the second inequality comes from the fact that $P_{\bar{\thetab}}(s' \mid s, a) \le 1$ is a multinomial probability, the third inequality holds due to the Cauchy-Schwarz inequality, and the last inequality follows from~\Cref{lemma:concentration of online estimated transition core} and the definition of $\hat{\varphib}_{k,h}$, i.e., $\hat{\varphib}_{k,h}(s,a):= \varphib(s,a,\hat{s})$ for $\hat{s}=\argmax_{s' \in \Scal_{s,a}} \| \varphib(s,a,s') \|_{\Ab_{k,h}^{-1}}$.        
\end{proof}

\subsection{Good Events with High Probability}
\begin{lemma}[Good event probability] \label{lemma:good event prob}
    For any $K \in \NN $ and $\delta \in (0,1)$, the good event $\Gcal(K,\delta')$ holds with probability at least $1 - \delta$ where $\delta' = \delta /(2KH)$.
\end{lemma}

\begin{proof}[Proof of Lemma~\ref{lemma:good event prob}]
    For any $\delta' \in (0,1)$, we have
    \begin{equation*}
        \Gcal(K, \delta')
        = \bigcap_{k \le K} \bigcap_{h \le H} \Gcal_{k,h} (\delta')
        = \bigcap_{k \le K} \bigcap_{h \le H} \left\{ \Gcal_{k,h}^\Delta (\delta') \cap \Gcal_{k,h}^{\xib} (\delta')  \right\} \, .
    \end{equation*}
    On the other hand, for any $(k,h) \in [K] \times [H]$, by Lemma~\ref{auxiliary lemma: gaussian noise concentration}, $\Gcal_{k,h}^{\xib} (\delta') $ holds with probability at least $1 - \delta'$.
    Then, for $\delta' = \delta / (2KH) $ by taking union bound, we have the desired result as follows:
        \begin{equation*}
            \PP (\Gcal(K, \delta')) \ge (1 - \delta')^{2KH} \ge 1 - 2KH \delta' = 1 - \delta \, .
        \end{equation*}
\end{proof}

\subsection{Stochastic Optimism}
\begin{lemma}[Stochastic optimism] \label{lemma:stochastic optimism}
    For any $\delta$ with $0 < \delta < \Phi(-1)/2$, let $\sigma_k = H \alpha_k (\delta) = \BigOTilde(H \sqrt{d})$. 
    If we take multiple sample size $M = \lceil 1 - \frac{\log H}{\log \Phi(1)} \rceil$, then for any $k \in [K]$, we have
        \begin{equation*}
            \PP \left( (V^k_1 - V^*_1)(s^k_1) \ge 0 \mid s^k_1, \Fcal_k \right) \ge \Phi(-1)/2 \, .
        \end{equation*}
\end{lemma}
\begin{proof}[Proof of~\cref{lemma:stochastic optimism}]
    Before presenting the proof, we introduce the following lemmas.

    \begin{lemma} \label{supporting lemma:pessimism decomposition}
        For any $k \in [K]$, it holds
        \begin{equation*}
            V^k_1(s^k_1) - V^*_1 (s^k_1) \ge \EE_{\pi^*} \left[ \sum_{h=1}^H - \BE^k_h(x_h,a_h) \mid x_1 = s^k_1 \right] \, ,
        \end{equation*}
        where $\BE^k_h(s,a):= r(s,a) + P_h V^k_{h+1}(s,a) - Q^k_h(s,a)$.
    \end{lemma}
    
    \begin{lemma} \label{supporting lemma:stochastic optimism for given h and k}
        Let $\delta \in (0,1)$ be given. For any $(k, h) \in [K] \times [H]$, let $\sigma_k = H \alpha_k(\delta)$.
        If we define the event $\Gcal_{k,h}^{\Delta} (\delta)$ as
        \begin{equation*}
            \Gcal_{k,h}^{\Delta}(\delta) :=
                \left\{ \Delta^k_h(s,a) \le H \alpha_k(\delta) \| \hat{\varphib}_{k,h}(s,a) \|_{\Ab_{k,h}^{-1}} \right\} \, ,
        \end{equation*}        
        then conditioned on $\Gcal_{k,h}^{\Delta} (\delta)$, for any $(s,a) \in \Scal \times \Acal$, we have
        \begin{equation*}
            \PP \left( - \BE^k_h(s,a) \ge 0 \mid \Gcal^{\Delta}_{k,h}(\delta) \right) \ge 1 - \Phi(1)^M \, .
        \end{equation*}
    \end{lemma}
    
    \begin{lemma} \label{supporting lemma:stochastic optimism for all h}
        Let $\delta \in (0,1)$ be given. For any $(h, k) \in [H] \times [K]$, let $\sigma_k = H \alpha_k(\delta)$. 
        If we take multiple sample size $M = \lceil 1 - \frac{\log H}{\log \Phi(1)} \rceil$, then conditioned on the event $\Gcal^{\Delta}_{k}(\delta) := \bigcap_{h \in [H]} \Gcal^{\Delta}_{k,h}(\delta)$, we have
        \begin{equation*}
            \PP \left( - \BE^k_h(s_h, a_h) \ge 0, \forall h \in [H] \mid \Gcal^{\Delta}_k (\delta) \right) \ge \Phi(-1) \, .
        \end{equation*}
    \end{lemma}

    Now, we define the event of the estimated value function being optimistic at the start of the $k$-th episode as
    \begin{equation*}
        \Xcal_k := \left\{ (V^k_1 - V^*_1) (s^k_1) \ge 0 \right\} \, .
    \end{equation*}
    Then for the event $\Gcal_k (\delta) =: \Gcal_k$, we have
    \begin{align*}
        \PP (\Xcal_k) 
        & = 1 - \PP (\Xcal_k^\mathsf{c})
        \\
        & = 1 - \PP (\Xcal_k^\mathsf{c} \cap \Gcal_k) - \PP (\Xcal_k^\mathsf{c} \cap \Gcal_k^\mathsf{c})
        \\
        & \ge 1 - \PP (\Xcal_k^\mathsf{c} \cap \Gcal_k) - \PP ( \Gcal_k^\mathsf{c})
        \\
        & \ge 1 - \PP (\Xcal_k^\mathsf{c} \cap \Gcal_k) - \delta \, 
    \end{align*}
    where the last inequality comes from~\cref{lemma:good event prob}.
    
    On the other hand, by Lemma~\ref{supporting lemma:pessimism decomposition}, we have
    \begin{align*}
        V^k_1 (s^k_1) - V^*_1 (s^k_1) 
        & \ge \EE_{\pi^*} \left[ \sum_{h=1}^H - \BE^k_h(x_h,a_h) \mid x_1 = s^k_1 \right]
        \\
        & = \sum_{h=1}^H \EE_{\pi^*} \left[ - \BE^k_h(x_h,a_h) \mid x_1 = s^k_1 \right] \, .
    \end{align*}
    If we define an event
    \begin{equation*}
        \Ycal_k = \left\{ \sum_{h=1}^H \EE_{\pi^*} \left[ - \BE^k_h (x_h,a_h) \mid x_1 = s^k_1 \right] \ge 0 \right\} \, ,
    \end{equation*}    
    then, by Lemma~\ref{supporting lemma:stochastic optimism for all h}, we have 
    \begin{align*}
        \PP(\Ycal_k \mid \Gcal_k) \ge \Phi(-1)
        & \iff \PP(\Ycal_k^\mathsf{c} \mid \Gcal_k) \le 1 - \Phi(-1)
        \\
        & \implies \PP (\Ycal_k^\mathsf{c} \cap \Gcal_k) \le \left( 1 - \Phi(-1) \right) \PP (\Gcal_k) \le 1 - \Phi(-1)
    \end{align*}
Note that since $\Xcal_k^\mathsf{c} \cap \Gcal_k \subset \Ycal_k^\mathsf{c} \cap \Gcal_k$, we can conclude that
    \begin{align*}
        \PP (\Xcal_k) 
        & \ge 1 - \PP (\Xcal_k^\mathsf{c} \cap \Gcal_k) - \delta
        \\
        & \ge 1 - \PP (\Ycal_k^\mathsf{c} \cap \Gcal_k) - \delta
        \\
        & \ge 1 - (1 - \Phi(-1)) - \delta
        \\
        & = \Phi(-1) - \delta
        \\ 
        & \ge \Phi(-1) / 2
    \end{align*}
where the last inequality comes from the choice of $\delta$.   
\end{proof}

In the following, we provide all the proofs of the lemmas used to prove Lemma~\ref{lemma:stochastic optimism}.

\subsubsection{Proof of\texorpdfstring{~\Cref{supporting lemma:pessimism decomposition}}{Lemma 7}}
\begin{proof}[Proof of~\cref{supporting lemma:pessimism decomposition}]
    In this proof, we use $x^k_h$ as the states sampled under the $\pi^*$ to distinguish with $s^k_h$.
    Since we have,
    \begin{align*}
        & V^k_1 (s^k_1) - V^*_1 (s^k_1)
        \\
        & \ge Q^k_1 (s^k_1, \pi^*(s^k_1)) - Q^*_1 (s^k_1, \pi^*(s^k_1) )
        \\
        & = r(s^k_1, \pi^*(s^k_1)) + P_1 V^k_2 (s^k_1, \pi^*(s^k_1)) - \BE^k_1 (s^k_1, \pi^*(s^k_1)) - \left( r(s^k_1, \pi^*(s^k_1)) + P_1 V^*_2(s^k_1, \pi^*(s^k_1)) \right)
        \\
        & = P_1 (V^k_2 - V^*_2)(s^k_1, \pi^*(s^k_1)) - \BE^k_1 (s^k_1, \pi^*(s^k_1))
        \\
        & = \EE_{x \mid s^k_1, \pi^*(s^k_1)} \left[ (V^k_2 - V^*_2)(x) \right] - \BE^k_1 (s^k_1, \pi^*(s^k_1))
        \\
        & \ge \EE_{x^k_2 \mid s^k_1, \pi^*(s^k_1)} \left[ (Q^k_2 - Q^*_2)(x^k_2, \pi^*(x^k_2)) \right] - \BE^k_1 (s^k_1, \pi^*(s^k_1))
        \\
        & = \EE_{x^k_2 \sim s^k_1, \pi^*(s^k_1)} \left[
        \EE_{x \mid x^k_2, \pi^*(x^k_2)} \left[ (V^k_3 - V^*_3)(x) \right] - \BE^k_2 (x^k_2, \pi^*(x^k_2)) 
        \right] 
        - \BE^k_1 (s^k_1, \pi^*(s^k_1))
        \\
        & = \underbrace{\EE_{x^k_2 \sim s^k_1, \pi^*(s^k_1)} \left[
        \EE_{x \mid x^k_2, \pi^*(x^k_2)} \left[ (V^k_3 - V^*_3)(x) \right] \right]}_{\EE_{x^k_3 \sim \pi^* \mid s^k_1} \left[ (V^k_3 - V^*_3)(x^k_3) \right]}
        \\
        & \phantom{{}={}} - \EE_{x^k_2 \sim s^k_1, \pi^*(s^k_1)} \left[ \BE^k_2 (x^k_2, \pi^*(x^k_2)) 
        \right] 
        - \BE^k_1 (s^k_1, \pi^*(s^k_1))
    \end{align*}
    then by applying this argument recursively, we finally have
    \begin{equation*}
        V^k_1 (s^k_1) - V^*_1 (s^k_1) \ge \EE_{\pi^*} \left[ \sum_{h=1}^H - \BE^k_h (x_h,a_h) \mid x_1 = s^k_1 \right] \, .
    \end{equation*}
\end{proof}

\subsubsection{Proof of\texorpdfstring{~\Cref{supporting lemma:stochastic optimism for given h and k}}{Lemma 8}}
\begin{proof}[Proof of~\Cref{supporting lemma:stochastic optimism for given h and k}]
    Since we have
    \begin{align*}
        - \BE^k_h(s,a) 
        & = Q^k_h (s,a) - \left( r(s,a) + P_h V^k_{h+1}(s,a) \right)
        \\
        & = \min \left\{ r(s,a) + \sum_{s' \in \Scal_{s,a}} P_{\thetab^k_h}(s' \mid s, a) V^k_{h+1}(s')
        + \max_{m \in [M]} \hat{\varphib}_{k,h}(s,a)^\top \xi^{(m)}_{k,h}, H \right\}
            \\
            & \phantom{{}={}} - \left( r(s,a) + P_h V^k_{h+1}(s,a) \right)
        \\
        & \ge \min \left\{ \sum_{s' \in \Scal_{s,a}} P_{\thetab^k_h}(s' \mid s,a) V^k_{h+1}(s')
        + \max_{m \in [M]} \hat{\varphib}_{k,h}(s,a)^\top \xi^{(m)}_{k,h}
        -  P_h V^k_{h+1}(s,a), 0 \right\} \, ,
    \end{align*}
it is enough to show that 
    \begin{equation*}
        \sum_{s' \in \Scal_{s,a}} P_{\thetab^k_h} (s' \mid s, a) V^k_{h+1} (s') + \max_{m \in [M]} \hat{\varphib}_{k,h}(s,a)^\top \xi^{(m)}_{k,h} -  P_h V^k_{h+1}(s,a) \ge 0
    \end{equation*}
at least with constant probability.

On the other hand, under the event $\Gcal_{k,h}(\delta)$, by Lemma~\ref{lemma:bound of prediction error} we have
    \begin{align*}
        & \sum_{s' \in \Scal_{s,a}} P_{\thetab^k_h}(s' \mid s, a) V^k_{h+1}(s') + \max_{m \in [M]} \hat{\varphib}_{k,h}(s,a)^\top \xi^{(m)}_{k,h} -  P_h V^k_{h+1}(s,a)
        \\
        & \ge \max_{m \in [M]} \hat{\varphib}_{k,h}(s,a)^\top \xi^{(m)}_{k,h} - H \alpha_k (\delta) \| \hat{\varphib}_{k,h}(s,a) \|_{\Ab_{k,h}^{-1}} \, . 
    \end{align*}
Now, for $\forall m \in [M]$, since $\xi_{k,h}^{(m)} \sim \Ncal(\zero_d, \sigma_k^2 \Ab_{k,h}^{-1})$, we have 
    \begin{equation*}
        \hat{\varphib}_{k,h}(s,a)^\top \xi_{k,h}^{(m)} \sim \Ncal (0, \sigma_k^2 \| \hat{\varphib}_{k,h}(s,a) \|_{\Ab_{k,h}^{-1}}^2) \, ,
    \end{equation*}
which means, 
    \begin{equation*}
        \PP \left( \hat{\varphib}_{k,h}(s,a)^\top \xi_{k,h}^{(m)} \ge H \alpha_k(\delta) \| \hat{\varphib}_{k,h}(s,a) \|_{\Ab_{k,h}^{-1}} \right) \ge \Phi(-1) \, ,
    \end{equation*}
by setting $\sigma_k = H \alpha_k (\delta)$.
Then, finally we have the desired results as follows:
    \begin{align*}
        & \PP \left( - \BE^k_h(s,a) \ge 0 \mid \Gcal^{\Delta}_{k,h}(\delta) \right)
        \\
        & \ge \PP \left( \max_{m \in [M]} \hat{\varphib}_{k,h}(s,a)^\top \xi_{k,h}^{(m)} \ge H \alpha_k(\delta) \| \hat{\varphib}_{k,h}(s,a) \|_{\Ab_{k,h}^{-1}} \mid \Gcal^{\Delta}_{k,h}(\delta) \right)
        \\
        & = 1 - \PP \left( \hat{\varphib}_{k,h}(s,a)^\top \xi_{k,h}^{(m)} < H \alpha_k(\delta) \| \hat{\varphib}_{k,h}(s,a) \|_{\Ab_{k,h}^{-1}}, \forall m \in [M] \mid \Gcal^{\Delta}_{k,h}(\delta) \right)
        \\
        & \ge 1 - (1 - \Phi(-1))^M
        \\
        & = 1 - \Phi(1)^M \, .
    \end{align*}    
\end{proof}

\subsubsection{Proof of\texorpdfstring{~\Cref{supporting lemma:stochastic optimism for all h}}{Lemma 9}}
\begin{proof}[Proof of~\Cref{supporting lemma:stochastic optimism for all h}]
    For each $h \in [H]$ and $k \in [K]$, define an event $\Ecal^k_h:= \{ - \BE^k_h (s_h, a_h) \ge 0 \}$ 
    Then it holds 
        \begin{align*}
            \PP \left( - \BE^k_h (s_h, a_h) \ge 0, \forall h \in [H] \mid \Gcal^{\Delta}_k(\delta) \right)
            & = \PP \left( \bigcap^H_{h=1} \Ecal^k_h \mid \Gcal^{\Delta}_k(\delta) \right)
            \\
            & = 1 - \PP \left( \bigcup^H_{h=1} ( \Ecal^k_h )^{\mathsf{c}} \mid \Gcal^{\Delta}_k(\delta) \right)
            \\
            & \ge 1- \sum_{h=1}^{H} \PP \left( ( \Ecal^k_h )^{\mathsf{c}} \mid \Gcal^{\Delta}_{k,h}(\delta) \right)
            \\
            & \ge 1 - H \Phi(1)^M
            \\
            & \ge \Phi(-1) \, 
        \end{align*}
    where the first inequality uses the union bound, the second inequality comes from the Lemma~\ref{supporting lemma:stochastic optimism for given h and k} and the last inequality holds due to the choice of $M = \lceil 1 - \frac{\log H}{\log \Phi(1)}\rceil$.    
\end{proof}

\subsection{Bound on Estimation Part}
We decompose the regret into the estimation part and the pessimism part as follows:
\begin{equation*}
    \sum_{k=1}^K (V^*_1 - V^{\pi^k}_1)(s^k_1)
    = \sum_{k=1}^K \Big( \underbrace{V^*_1 - V^k_1}_{\text{Pessimism}} + \underbrace{V^k_1 - V^{\pi^k}_1}_{\text{Estimation}} \Big)(s^k_1) \, ,
\end{equation*}
and we bound these two parts in the following sections, respectively.

\begin{lemma}[Bound on estimation part] \label{lemma:bound of estimation part}
    For any $\delta \in (0, 1)$, if $\lambda \ge \Fnorm^2$, then with probability at least $1 - \delta / 2$, we have
    \begin{equation*}
        \sum_{k=1}^{K} (V^k_1 - V^{\pi^k}_1) (s^k_1) = \tilde{\Ocal} \left( \kappa^{-1} d^{\frac{3}{2}} H^{\frac{3}{2}} \sqrt{T}
        \right) \, .
    \end{equation*}
\end{lemma}

\begin{proof}[Proof of~\cref{lemma:bound of estimation part}]
    For any given $k \in [K]$,
    \begin{align}
        (V^k_1 - V^{\pi^k}_1) (s^k_1)
        & = (Q^k_1 - Q^{\pi^k}_1) (s^k_1, a^k_1) + \BE^k_1(s^k_1, a^k_1) - \BE^k_1(s^k_1, a^k_1) \nonumber
        \\
        & = (Q^k_1 - Q^{\pi^k}_1) (s^k_1, a^k_1) + P_1 (V^k_2 - V^{\pi^k}_{2})(s^k_1, a^k_1) 
        \\
        & \phantom{{}={}} + (Q^{\pi^k}_1 - Q^k_1)(s^k_1, a^k_1) - \BE^k_1(s^k_1, a^k_1) \nonumber
        \\
        & =  \underbrace{P_1 (V^k_2 - V^{\pi^k}_{2})(s^k_1, a^k_1) - (V^k_2 - V^{\pi^k}_2) (s^k_2) }_{\dot{\zeta}^k_1} + (V^k_2 - V^{\pi^k}_2) (s^k_2) - \BE^k_1 (s^k_1, a^k_1) \nonumber
    \end{align}
    where the second equality holds due to the variant of $\BE^k_h(s^k_h, a^k_h)$ as follows:
    \begin{align*}
        \BE^k_h (s^k_h, a^k_h)
        & = r(s^k_h, a^k_h) + P_h V^k_{h+1}(s^k_h, a^k_h) - Q^k_h (s^k_h, a^k_h) + Q^{\pi^k}_h (s^k_h, a^k_h) - Q^{\pi^k}_h (s^k_h, a^k_h)
        \\
        & = r(s^k_h, a^k_h) + P_h V^k_{h+1}(s^k_h, a^k_h) - Q^k_h (s^k_h, a^k_h) 
        \\
        &\phantom{{}={}} + Q^{\pi^k}_h (s^k_h, a^k_h) - \left( r(s^k_h, a^k_h) + P_h V^{\pi^k}_{h+1} (s^k_h, a^k_h) \right)
        \\
        & =  P_h (V^k_{h+1} - V^{\pi^k}_{h+1}) (s^k_h, a^k_h) + (Q^{\pi^k}_h - Q^k_h) (s^k_h, a^k_h) \, .
    \end{align*}
    Then, by applying this argument recursively for whole horizon, we have
    \begin{equation} \label{eq:lemma:bound on estimation eq 0}
        (V^k_1 - V^{\pi^k}_1) (s^k_1) = \sum_{h=1}^H - \BE^k_h (s^k_h, a^k_h) + \sum_{h=1}^H \dot{\zeta}^k_h \, ,
    \end{equation}
    where $\dot{\zeta}^k_h:= P_h (V^k_{h+1} - V^{\pi^k}_{h+1})(s^k_h, a^k_h) - (V^k_{h+1} - V^{\pi^k}_{h+1}) (s^k_{h+1})$.

    Let $\delta' = \delta/(8KH)$. 
    By Lemma~\ref{lemma:good event prob}, the good event $\Gcal(K, \delta')$ holds with probability at least $1 - \delta/4$.
    Then under the event $\Gcal(K, \delta')$, for any $h \in [H]$ we have
    \begin{align}
        &- \BE^k_h (s^k_h, a^k_h) \nonumber
        \\
        & = Q^k_h (s^k_h, a^k_h) - \left( r(s^k_h, a^k_h) + P_h V^k_{h+1}(s^k_h, a^k_h) \right) \nonumber
        \\
        & = \min \left\{ r(s^k_h, a^k_h) + \sum_{s' \in \Scal_{k,h}} P_{\thetab^k_h}(s' \mid s^k_h, a^k_h) V^k_{h+1}(s') 
        + \max_{m \in [M]} \hat{\varphib}_{k,h}(s^k_h, a^k_h)^\top \xi^{(m)}_{k,h}, H\right\} \nonumber
            \\
            & \phantom{{}={}} - \left( r(s^k_h, a^k_h) + P_h V^k_{h+1}(s^k_h, a^k_h) \right) \nonumber
        \\
        & \le \sum_{s' \in \Scal_{k,h}} P_{\thetab^k_h}(s' \mid s^k_h, a^k_h) V^k_{h+1}(s')
        + \max_{m \in [M]} \hat{\varphib}_{k,h}(s^k_h, a^k_h)^\top \xi^{(m)}_{k,h}
        - P_h V^k_{h+1}(s^k_h, a^k_h) \nonumber
        \\
        & \le \left| \sum_{s' \in \Scal_{k,h}} P_{\thetab^k_h}(s' \mid s^k_h, a^k_h) V^k_{h+1}(s') - P_h V^k_{h+1}(s^k_h, a^k_h) \right|  
        + \max_{m \in [M]} \left|  \hat{\varphib}_{k,h}(s^k_h, a^k_h)^\top \xi^{(m)}_{k,h} \right| \nonumber
        \\
        & \le |\PE^k_h (s^k_h, a^k_h)| 
        + \max_{m \in [M]} \| \hat{\varphib}_{k,h}(s^k_h, a^k_h) \|_{\Ab_{k,h}^{-1}} \| \xi^{(m)}_{k,h} \|_{\Ab_{k,h}} \label{eq:lemma-bound on estimation eq 1}
        \\
        & \le \left( H \alpha_k(\delta') + \gamma_k (\delta') \right) \| \DomFeat_{k,h}(s^k_h, a^k_h) \|_{\Ab_{k,h}^{-1}} \, , \label{eq:lemma-bound on estimation eq 2}
    \end{align}
    where~\eqref{eq:lemma-bound on estimation eq 1} comes from the Cauchy-Schwarz inequality and~\eqref{eq:lemma-bound on estimation eq 2} holds due the the Lemma~\ref{lemma:bound of prediction error} \& \ref{auxiliary lemma: gaussian noise concentration}.
    Then, with probability at least $1 - \delta/4$, we have
    \begin{equation} \label{eq:lemma-bound on estimation eq 3}
        \sum_{h=1}^H - \BE^k_h (s^k_h, a^k_h) \le \sum_{h=1}^H \left( H \alpha_k(\delta') + \gamma_k (\delta') \right) \| \DomFeat_{k,h}(s^k_h, a^k_h) \|_{\Ab_{k,h}^{-1}} \, . 
    \end{equation}

    On the other hand, for $\dot{\zeta}^k_h$, we have $| \dot{\zeta}^k_{h} | \le 2 H$ and $\EE[ \dot{\zeta}^k_h \mid \Fcal_{k,h} ] = 0 $, which means $\{ \dot{\zeta}^k_h \mid \Fcal_{k,h} \}_{k,h}$ is a martingale difference sequence for any $k \in [K]$ and $h \in [H]$.
    Hence, by applying the Azuma-Hoeffding inequality with probability at least $1 - \delta/4$, we have
    \begin{equation} \label{eq:lemma-bound on estimation eq 4}
        \sum_{k=1}^K \sum_{h=1}^H \dot{\zeta}^k_h \le 2H \sqrt{2 K H \log(4 / \delta)} \, . 
    \end{equation}

    Combining the results of~\eqref{eq:lemma-bound on estimation eq 3} and~\eqref{eq:lemma-bound on estimation eq 4}, with probability at least $1 - \delta/2 $, we have
    \begin{align}
        & (V^k_1 - V^{\pi^k}_1) (s^k_1) 
        \nonumber
        \\
        & \le 2H \sqrt{2 T \log(4 / \delta)} + \sum_{k=1}^K \sum_{h=1}^H \left( H \alpha_k(\delta') + \gamma_k (\delta') \right) \| \DomFeat_{k,h}(s^k_h, a^k_h) \|_{\Ab_{k,h}^{-1}} \nonumber
        \\
        & \le 2H \sqrt{2 T \log(4 / \delta)} + \left( H \alpha_K (\delta') + \gamma_K (\delta') \right) \sum_{k=1}^K \sum_{h=1}^H  \| \DomFeat_{k,h}(s^k_h, a^k_h) \|_{\Ab_{k,h}^{-1}} \label{eq:lemma-bound on estimation eq 5}
        \\
        & \le 2H \sqrt{2 T \log(4 / \delta)} + \left( H \alpha_K (\delta') + \gamma_K (\delta') \right) \sum_{h=1}^H \sqrt{K \sum_{k=1}^K   \| \DomFeat_{k,h}(s^k_h, a^k_h) \|^2_{\Ab_{k,h}^{-1}}} \label{eq:lemma-bound on estimation eq 6}
        \\
        & \le 2H \sqrt{2 T \log(4 / \delta)} + \left( H \alpha_K (\delta') + \gamma_K (\delta') \right) \sum_{h=1}^H \sqrt{4 \kappa^{-1} K d  \log \left( 1 + \frac{K \Ucal \Fnorm^2}{d \lambda} \right) } \label{eq:lemma-bound on estimation eq 7}
        \\
        & = 2H \sqrt{2 T \log(4 / \delta)} + \left( H \alpha_K (\delta') + \gamma_K (\delta') \right) \sqrt{4 \kappa^{-1} T H d \log \left( 1 + \frac{K \Ucal \Fnorm^2 }{d \lambda} \right) } \, , \nonumber
        \\
        & = \tilde{\Ocal} \left( \kappa^{-1} d^{\frac{3}{2}} H^{\frac{3}{2}} \sqrt{T} + H \sqrt{T} \right) \nonumber \, ,
    \end{align}
    where \eqref{eq:lemma-bound on estimation eq 5} follows from the fact that both $\alpha_k(\delta)$ and $\gamma_k(\delta)$ are increasing in $k$, \eqref{eq:lemma-bound on estimation eq 6} comes from Cauchy-Schwarz inequality and \eqref{eq:lemma-bound on estimation eq 7} holds by the generalized elliptical potential lemma (Lemma~\ref{lemma:generalized elliptical potential}).        
\end{proof}

\subsection{Bound on Pessimism Part}
\begin{lemma}[Bound on pessimism] \label{lemma:bound of pessimism part}
    For any $\delta$ with $0 < \delta < \Phi(-1)/2$, let $\sigma_k = H \alpha_k (\delta)$. 
    If $\lambda \ge \Fnorm^2$ and we take multiple sample size $M = \lceil 1 - \frac{\log H}{\log \Phi(1)} \rceil$, then with probability at least $1 - \delta / 2$, we have
    \begin{equation*}
        \sum_{k=1}^{K} (V^*_1 - V^k_1) (s^k_1) = \tilde{\Ocal} \left( \kappa^{-1} d^{\frac{3}{2}} H^{\frac{3}{2}} \sqrt{T}
        \right) \, .
    \end{equation*}
\end{lemma}
\begin{proof}[Proof of~\cref{lemma:bound of pessimism part}]
    Similar to the techniques used in~\cite{zanette2020frequentist}, we show that the difference between the optimal value function $V^*_1$ and the estimated value function $V^k_1$ can be controlled by constructing an upper bound on $V^*_1$ and a lower bound on $V^k_1$.
    In this proof, we consider three kinds of pseudo-noises, $\xib, \bar{\xib}$ and $\underline{\xib}$ that we define later in the proof. 
    Also, for $\delta' = \delta/10 $, we denote $\Gcal(K, \delta'), \bar{\Gcal} (K, \delta')$ and $\underline{\Gcal} (K, \delta') $ as the good events induced by $\xib, \bar{\xib}$ and $\underline{\xib}$ respectively.
    From now on, we denote $G(K, \delta')$ by the event $\Gcal(K, \delta') \cap \bar{\Gcal} (K, \delta') \cap \underline{\Gcal} (K, \delta')$.
    Then, by Lemma~\ref{lemma:good event prob}, the event $G(K, \delta')$ holds with high probability at least $1 - 3\delta/10$.

    First, we construct the lower bound of $V^k_{1}$.
    For any given $k \in [K]$, let $ \tilde{\xib} := \{ \tilde{\xib}_{k,h}^{(m)} \}_{m \in [M]} \subset \RR^d$ be a set of vectors for $h \in [H]$ and $V^k_h(\cdot \, ; {\tilde{\xib}})$ be the value function obtained by the Algorithm~\ref{alg:Algorithm 1} with non-random $\tilde{\xib}^{(m)}_{k,h}$ in place of $\xib^{(m)}_{k,h}$.
    Then consider the following minimization problem:
    \begin{mini*}|s|
        {\{ \tilde{\xib}^{(m)}_{k,h} \}_{h \in [H], m \in [M]}}{V^k_1 (s^k_1 ; \tilde{\xib})}
        {}{}
        \addConstraint{\max_{m \in [M]} \| \tilde{\xib}^{(m)}_{k,h} \|_{\Ab_{k,h}} \le \gamma_k(\delta), \quad \forall h \in [H] }
    \end{mini*}
    And we denote $\underline{\xib} := \{ \underline{\xib}^{(m)}_{k,h} \}_{h \in [H], m \in [M]}$ by a minimizer and $\underline{V}^k_1(s^k_1)$ by the minimum of the above minimization problem, i.e., $\underline{V}^k_h(\cdot) := V^k_h( \cdot \, ; \underline{\xib})$.
    Then, under the event $\Gcal(K,\delta')$, since $\{ \xib^{(m)}_{k,h} \}_{h \in [H], m \in [M]}$ is also a feasible solution of the above optimization problem, and since $V^k_h = V^k_h (\, ; \xib)$, thus we have
    \begin{equation}
        \underline{V}^k_1(s^k_1) \le V^k_1(s^k_1) \, . \label{eq:lemma:bound on pessimism eq 3}
    \end{equation}

    Second, to find an upper bound for $V^*$, considering i.i.d copies $\{ \bar{\xib}^{(m)}_{k,h} \}_{h \in [H], m \in [M]}$ of $\{ \xib^{(m)}_{k,h} \}_{h \in [H], m \in [M]}$ and run Algorithm~\ref{alg:Algorithm 1} to get a corresponding value function $\bar{V}^k_h$ and $\bar{Q}^k_h$ for all $h \in [H]$.
    Define the event that $\bar{V}^k_1(s^k_1)$ is optimistic in the $k$-th episode as
    \begin{equation*}
        \bar{\Xcal}_k = \{ ( \bar{V}^k_1 - V^*_1 ) (s^k_1) \ge 0 \} \, .
    \end{equation*}
    Then by Lemma~\ref{lemma:stochastic optimism}, for given $\delta$, we have 
    \begin{equation*}
        \PP ( \bar{\Xcal}_k \mid s^k_1, \Fcal_k) \ge \Phi(-1) / 2 \, .
    \end{equation*}
    Then by the definition of optimism, under the event $\Gcal(K, \delta')$, we have
    \begin{align}
        (V^*_1 - V^k_1)(s^k_1)
        & \le \EE_{\bar{\xib} \mid \bar{\Xcal}_k} \left[ (\bar{V}^k_1 - V^k_1)(s^k_1) \right] \nonumber
        \\
        & \le \EE_{\bar{\xib} \mid \bar{\Xcal}_k} \left[ (\bar{V}^k_1 - \underline{V}^k_1)(s^k_1) \right] \, , \label{eq:lemma:bound on pessimism eq 2}
    \end{align}
    where the expectations are over the $\bar{\xib}$'s conditioned on the event $\bar{\Xcal}_k$ and the second inequality comes from~\eqref{eq:lemma:bound on pessimism eq 3}.    
    On the other hand, under the event $\bar{\Gcal}(K,\delta')$ by the law of the total expectation, we have
    \begin{align}
        \EE_{\bar{\xib}} \left[ (\bar{V}^k_1 - \underline{V}^k_1)(s^k_1) \right] 
        & = \EE_{\bar{\xib} \mid \bar{\Xcal}_k} \left[ (\bar{V}^k_1 - \underline{V}^k_1)(s^k_1) \right] \PP(\bar{\Xcal}_k)
        + \EE_{\bar{\xib} \mid \bar{\Xcal}^\mathsf{c}_k} \left[ (\bar{V}^k_1 - \underline{V}^k_1)(s^k_1) \right] \PP(\bar{\Xcal}^\mathsf{c}_k) \nonumber
        \\
        & \ge \EE_{\bar{\xib} \mid \bar{\Xcal}_k} \left[ (\bar{V}^k_1 - \underline{V}^k_1)(s^k_1) \right] \PP(\bar{\Xcal}_k) \, , \label{eq:lemma:bound on pessimism eq 1}
    \end{align}
    where~\eqref{eq:lemma:bound on pessimism eq 1} comes from the fact that $\{ \bar{\xib}^{(m)}_{k,h} \}_{h \in [H], m \in [M]}$ is also a feasible solution of the above optimization problem under the event $\bar{\Gcal}(K, \delta')$, i.e., $\bar{V}^k_1 (s^k_1) \ge \underline{V}^k_1 (s^k_1)$.    
    Then, by combining the results of~\eqref{eq:lemma:bound on pessimism eq 1} and~\eqref{eq:lemma:bound on pessimism eq 2}, under the event $G(K, \delta')$, we have
    \begin{align}
        (V^*_1 - V^k_1)(s^k_1)
        & \le \EE_{\bar{\xib} \mid \bar{\Xcal}_k} \left[ (\bar{V}^k_1 - \underline{V}^k_1)(s^k_1) \right]
        \nonumber
        \\
        & \le \EE_{\bar{\xib}} \left[ (\bar{V}^k_1 - \underline{V}^k_1)(s^k_1) \right] / \PP(\bar{\Xcal}_k)
        \nonumber
        \\
        & \le \frac{2}{\Phi(-1)} \EE_{\bar{\xib}} \left[ (\bar{V}^k_1 - V^k_1 + V^k_1 -  \underline{V}^k_1)(s^k_1) \right] 
        \nonumber
        \\
        & = \frac{2}{\Phi(-1)} \left( (V^k_1 -  \underline{V}^k_1)(s^k_1) \right)
        + \ddot{\zeta}_k \, , 
        \label{eq:lemma:bound on pessimism eq 5}
    \end{align}
    where we denote
    \begin{equation*}
        \ddot{\zeta}_k := \frac{2}{\Phi(-1)} \left( \EE_{\bar{\xib}} \left[ \bar{V}^k_1 (s^k_1) \right] - V^k_1 (s^k_1) \right)  \, .
    \end{equation*}
    
    Note that since $\bar{\xib}$ is the i.i.d copy of $\xib$, therefore $\bar{V}_{k,1}$ and $V_{k,1}$ are independent, which means $\{ \ddot{\zeta}_k \mid \Fcal_{k-1} \}_{k=1}^{K}$ is a martingale difference sequence with $|\ddot{\zeta}_k| \le \frac{2H}{\Phi(-1)}$. 
    Therefore by applying Azuma-Hoeffiding inequality under the event $G(K, \delta')$, with probability at least $1 - \delta'$, we have
    \begin{equation}
        \sum_{k=1}^K \ddot{\zeta}_k \le \frac{2 H}{\Phi(-1)} \sqrt{ 2 K \log(1/\delta')} \, . \label{eq:lemma:bound on pessimism eq 4}
    \end{equation}

    On the other hand, by dividing the first term in~\eqref{eq:lemma:bound on pessimism eq 5} into two terms we have
    \begin{equation*}
        (V^k_1 -  \underline{V}^k_1)(s^k_1) = \underbrace{(V^k_1 -  V^{\pi^k}_1)(s^k_1)}_{I_1} + \underbrace{(V^{\pi^k}_1 - \underline{V}^k_1)(s^k_1)}_{I_2} \, .
    \end{equation*}

    For $I_1$, note that since it is related to the estimation error, under the event $G(K, \delta')$ we can bound the sum of $I_1$ for the total episode number using Lemma~\ref{lemma:bound of estimation part} as follows:
    \begin{align} \label{eq:lemma:bound on pessimism eq 6}
        \sum_{k=1}^K (V^k_1 - V^{\pi^k}_1)(s^k_1) &\le \left( H \alpha_K (\delta') + \gamma_K (\delta') \right) \sqrt{4 \kappa^{-1} T H d \log \left( 1 + \frac{K \Ucal \Fnorm^2 }{d \lambda} \right) } \nonumber
        \\ 
        &\phantom{{}={}} + 2 H \sqrt{ 2T \log(1/\delta')} \, .
    \end{align}

    For $I_2$, since we have
    \begin{align}
        I_2 & = Q^{\pi^k}_1(s^k_1, a^k_1) - \underline{V}^k_1 (s^k_1)
        \nonumber
        \\
        & \le Q^{\pi^k}_1 (s^k_1, a^k_1) - \underline{Q}^k_1 (s^k_1, a^k_1) \label{eq:lemma:bound on pessimism eq 7}
        \\
        & = Q^{\pi^k}_1(s^k_1, a^k_1) - \underline{Q}^k_1(s^k_1, a^k_1) - \ULBE^k_1(s^k_1, a^k_1) + \ULBE^k_1 (s^k_1, a^k_1)
        \nonumber
        \\
        & = P_1 (V^{\pi^k}_2 - \underline{V}^k_2)(s^k_1, a^k_1) + \ULBE^k_1 (s^k_1, a^k_1)
        \label{eq:lemma:bound on pessimism eq 8}
        \\
        & = \underbrace{P_1 (V^{\pi^k}_2 - \underline{V}^k_2)(s^k_1, a^k_1) - (V^{\pi^k}_2 - \underline{V}^k_2)(s^k_2)}_{\dddot{\zeta}^k_1} + (V^{\pi^k}_2 - \underline{V}^k_2)(s^k_2) + \ULBE^k_1 (s^k_1, a^k_1) 
        \nonumber
    \end{align}
    where~\eqref{eq:lemma:bound on pessimism eq 7} comes from $a^k_1 = \argmax_a Q^k_1 (s^k_1, a)$ and~\eqref{eq:lemma:bound on pessimism eq 8} holds by the following definition of $\ULBE^k_h (s^k_h, a^k_h)$:
    \begin{align*}
       \ULBE^k_h (s^k_h, a^k_h) & := r(s^k_h, a^k_h) + P_h \underline{V}^k_{h+1}(s^k_h, a^k_h) - \underline{Q}^k_h (s^k_h, a^k_h)
        \\
        & = r(s^k_h, a^k_h) + P_h \underline{V}^k_{h+1}(s^k_h, a^k_h) - \underline{Q}^k_h(s^k_h, a^k_h) + Q^{\pi^k}_h(s^k_h, a^k_h) - Q^{\pi^k}_h(s^k_h, a^k_h)
        \\
        & = P_h(\underline{V}^k_{h+1} - V^{\pi^k}_{h+1})(s^k_h, a^k_h) + (Q^{\pi^k}_h - \underline{Q}^k_h)(s^k_h, a^k_h) \, .
    \end{align*}
    Then by applying the same argument recursively for the whole horizon, we have
    \begin{align}
        I_2 & \le \sum_{h=1}^{H} \ULBE^k_h (s^k_h, a^k_h) + \sum_{h=1}^{H} \dddot{\zeta}^k_h \, ,
        \nonumber 
    \end{align}
    where we denote 
    \begin{equation*}
        \dddot{\zeta}^k_h := P_h (V^{\pi^k}_{h+1} - \underline{V}^k_{h+1})(s^k_h, a^k_h) - (V^{\pi^k}_{h+1} - \underline{V}^k_{h+1})(s^k_{h+1}) \, .
    \end{equation*}

    Note that $\left\{ \dddot{\zeta}^k_h \mid \Fcal_{k,h} \right\}_{k,h}$ is a martingale difference sequence with $|\dddot{\zeta}^k_h | \le 2 H$.
    Then, under the event $G(K, \delta')$ by applying the Azuma-Hoeffding inequality with probability at least $1 - \delta'$, we have
    \begin{equation} \label{eq:lemma:bound on pessimism eq 13}
        \sum_{k=1}^K \sum_{h=1}^H \dddot{\zeta}^k_h \le 2 H \sqrt{2 T \log(1/\delta')} \, .
    \end{equation}
    To bound $\sum_{h=1}^{H} \ULBE^k_h (s^k_h, a^k_h)$, we divide the whole horizon index set into two groups as follows:
    \begin{equation*}
        \begin{split}
            & H^+ 
            \\
            &= \left\{ j \in [H] : r(s^k_j, a^k_j) + \sum_{s' \in \Scal_{k,j}} P_{\thetab^k_h}(s' \mid s^k_j, a^k_j) \underline{V}^k_{j+1}(s') + \max_{m \in [M]} \hat{\varphib}_{k,j}(s^k_j,a^k_j)^\top \underline{\xib}^{(m)}_{k,j} > H \right\}
            \\
            & H^- = [H] \backslash H^+ \, .
        \end{split}
    \end{equation*}

    Then, for $j \in H^+$ since $\underline{Q}^k_j(s^k_j, a^k_j) = H - j + 1$, $\underline{V}^k_{j+1} \le H - j$ and $r(s^k_j, a^k_j) \le 1$, we have
    \begin{equation}
        \ULBE^k_j (s^k_j, a^k_j) = r(s^k_j, a^k_j) + P_j \underline{V}^k_{j+1}(s^k_j, a^k_j) - (H-j+1) \le 0 \, . 
        \label{eq:lemma:bound on pessimism eq 11}
    \end{equation}
    On the other hand, for $j \in H^-$, under the event $G(K, \delta')$ we have
    \begin{align}
        \ULBE^k_j (s^k_j, a^k_j) & = P_j \underline{V}^k_{j+1}(s^k_j, a^k_j) - \sum_{s' \in \Scal_{k,j}}
        P_{\thetab^k_h}(s' \mid s^k_j, a^k_j) \underline{V}^k_{j+1}(s') - \max_{m \in [M]} \hat{\varphib}_{k,j}(s^k_j,a^k_j)^\top \underline{\xib}^{(m)}_{k,j} 
        \nonumber
        \\
        & \le \left| P_j \underline{V}^k_{j+1}(s^k_j, a^k_j) - \sum_{s' \in \Scal_{k,j}} P_{\thetab^k_h}(s' \mid s^k_j, a^k_j)  \underline{V}^k_{j+1}(s') \right|
        + \left| \max_{m \in [M]} \hat{\varphib}_{k,j}(s^k_j,a^k_j)^\top \underline{\xib}^{(m)}_{k,j} \right|
        \nonumber
        \\
        & \le H \alpha_k(\delta') \| \hat{\varphib}_{k,j}(s^k_j,a^k_j) \|_{\Ab_{k,j}^{-1}} +  \max_{m \in [M]} \left| \hat{\varphib}_{k,j}(s^k_j,a^k_j)^\top \underline{\xib}^{(m)}_{k,j} \right| 
        \label{eq:lemma:bound on pessimism eq 9}
        \\
        & \le H \alpha_k(\delta') \| \hat{\varphib}_{k,j}(s^k_j,a^k_j) \|_{\Ab_{k,j}^{-1}} +  \max_{m \in [M]} \| \hat{\varphib}_{k,j}(s^k_j,a^k_j) \|_{\Ab_{k,j}^{-1}}  \| \underline{\xib}^{(m)}_{k,j} \|_{\Ab_{k,j}}
        \nonumber
        \\
        &  \le \left( H \alpha_k(\delta') + \gamma_k (\delta') \right) \| \hat{\varphib}_{k,j}(s^k_j,a^k_j) \|_{\Ab_{k,j}^{-1}} \, ,
        \label{eq:lemma:bound on pessimism eq 10}
    \end{align}
    where~\eqref{eq:lemma:bound on pessimism eq 9} holds by Lemma~\ref{lemma:bound of prediction error}.

    By combining the result of \eqref{eq:lemma:bound on pessimism eq 11} and~\eqref{eq:lemma:bound on pessimism eq 10}, we have
    \begin{align*}
        I_2 & \le \sum_{j \in H^-} \left( H \alpha_k(\delta') + \gamma_k (\delta') \right) \| \hat{\varphib}_{k,j}(s^k_j,a^k_j) \|_{\Ab_{k,j}^{-1}} + \sum_{h=1}^{H} \dddot{\zeta}^k_h
        \\
        & \le \sum_{h=1}^H \left( H \alpha_k(\delta') + \gamma_k (\delta') \right) \| \hat{\varphib}_{k,h}(s^k_h,a^k_h) \|_{\Ab_{k,h}^{-1}} + \sum_{h=1}^{H} \dddot{\zeta}^k_h \, .
    \end{align*}
    
    Then summing $I_2$ over the total number of episodes, under the event $G(K, \delta')$, we have
    \begin{align}
        \sum_{k=1}^K (V^{\pi^k}_1 - \underline{V}^k_1)(s^k_1) 
        & \le \sum_{k=1}^K \sum_{h=1}^H \left( H \alpha_k(\delta') + \gamma_k (\delta') \right) \| \hat{\varphib}_{k,h}(s^k_h,a^k_h) \|_{\Ab_{k,h}^{-1}} 
        + \sum_{k=1}^K \sum_{h=1}^{H} \dddot{\zeta}^k_h
        \nonumber
        \\
        & \le (H \alpha_K (\delta') + \gamma_K(\delta')) \sum_{k=1}^K \sum_{h=1}^{H} \| \hat{\varphib}_{k,h}(s^k_h,a^k_h) \|_{\Ab_{k,h}^{-1}} + \sum_{k=1}^K \sum_{h=1}^{H} \dddot{\zeta}^k_h
        \nonumber
        \\
        & \le \left( H \alpha_K (\delta') + \gamma_K (\delta') \right) \sqrt{4 \kappa^{-1} T H d \log \left( 1 + \frac{K \Ucal \Fnorm^2 }{d \lambda} \right) } 
        \\
        &\phantom{{}={}} + 2 H \sqrt{2 T \log(1/\delta')} \, ,
        \label{eq:lemma:bound on pessimism eq 12}
    \end{align}
    where the last inequality holds due to the Lemma~\ref{lemma:generalized elliptical potential} and~\eqref{eq:lemma:bound on pessimism eq 13}.

    Finally, by summing~\eqref{eq:lemma:bound on pessimism eq 5} over $k$ and plugging the results of~\eqref{eq:lemma:bound on pessimism eq 6}, \eqref{eq:lemma:bound on pessimism eq 12} and~\eqref{eq:lemma:bound on pessimism eq 4} then, we have
    \begin{align*}
        & \sum_{k=1}^K (V^*_1 - V^k_1)(s^k_1) 
        \\
        & \le \frac{4}{\Phi(-1)} \left[ \left( H \alpha_K (\delta') + \gamma_K (\delta') \right) \sqrt{4 \kappa^{-1} T H d \log \left( 1 + \frac{K \Ucal \Fnorm^2 }{d \lambda} \right) } 
        + 2 H \sqrt{2 T \log(1/\delta')} \right]
        \\
        & \phantom{{}={}}
        + \frac{2 H}{\Phi(-1)} \sqrt{ 2 K \log(1/\delta')}
        \\
        & \le \tilde{\Ocal} \left( \kappa^{-1} d^{3/2} H^{3/2} \sqrt{T} + H \sqrt{T} + H \sqrt{K} \right) \, .
    \end{align*}
    
    To conclude the proof, by setting $\delta' = \delta/10$ and we take a union bound over the two applications of Azuma-Hoeffding ($\ddot{\zeta}_k, \dddot{\zeta}^k_h$) and the event $G(K, \delta')$, we get the desired result with probability at least $1 - \delta/2$.    
\end{proof}

\subsection{Regret Bound of \texorpdfstring{$\texttt{RRL-MNL}$}{RRL-MNL}}
\begin{proof}[Proof of~\Cref{thm:alg 1}]
    We can decompose the regret with estimation part and pessimism part as follows:
    \begin{align*}
        \Regret_\pi (K)
        & = \sum_{k=1}^{K} (V^*_1 - V^{\pi^k}_1) (s^k_1)
        \\
        & = \sum_{k=1}^{K} (V^*_1 - V^k_1) (s^k_1)
            + \sum_{k=1}^{K} (V^k_1 - V^{\pi^k}_1) (s^k_1) \, .
    \end{align*}
Since both Lemma~\ref{lemma:bound of estimation part} and Lemma~\ref{lemma:bound of pessimism part} holds with probability at least $1 - \delta / 2$ respectively, by taking the union bound the following holds with probability at least $1 - \delta$:
    \begin{align*}
        \Regret_\pi (K) 
        & = \tilde{\Ocal} \left( \kappa^{-1} d^{\frac{3}{2}} H^{\frac{3}{2}} \sqrt{T} + H \sqrt{T} + H \sqrt{K} \right) 
        + \tilde{\Ocal} \left( \kappa^{-1} d^{\frac{3}{2}} H^{\frac{3}{2}} \sqrt{T} + H \sqrt{T} \right)
        \\
        & = \tilde{\Ocal} \left( \kappa^{-1} d^{\frac{3}{2}} H^{\frac{3}{2}} \sqrt{T} \right) \, .
    \end{align*}    
\end{proof}

\section{Detailed Regret Analysis for \texorpdfstring{$\texttt{ORRL-MNL}$}{ORRL-MNL} (Theorem~\ref{thm:alg 2})} \label{appx:alg2}

\subsection{Concentration of Estimated Transition Core \texorpdfstring{$\omdtheta{k}{h}$}{}}
In this section, we provide the detailed proof of Lemma~\ref{lemma:tight concentration}, which demonstrates the concentration result for $\omdtheta{k}{h}$ independently of $\kappa$ and $\Ucal$.
Note that we adapt the proof provided by~\citet{zhang2023online} in the MNL contextual bandit setting to MNL-MDPs and improve the result, making it independent of $\Ucal$. We provide the lemmas for the concentration of the online transition core for completeness, noting that there are slight differences compared to their work, which stem from the different problem setting.

\begin{lemma}[Concentration of online estimated transition core] \label{lemma:tight concentration}
    Let $\eta = \Ocal(\log \Ucal)$ and $\lambda = \Ocal(d \log \Ucal)$.
    Then, for any $\delta \in (0,1]$ and for any $h \in [H]$, we have
    \begin{equation*}
        \PP \left( \forall k \ge 1, \left\| \omdtheta{k}{h} - \thetab^*_h \right\|_{\Bb_{k,h}} \le \beta_k (\delta) \right) \ge 1 - \delta \, ,
    \end{equation*}
    where $\beta_k (\delta)= \Ocal(\sqrt{d}  \log \Ucal \log (kH))$.
\end{lemma}

\begin{proof}[Proof of Lemma~\ref{lemma:tight concentration}]
    Recall that the transition core updated by the online mirror descent is represented by
    \begin{equation*}
        \omdtheta{k+1}{h} = \argmin_{\thetab \in \Bcal(\Pnorm)} \tilde{\ell}_{k,h}(\thetab) + \frac{1}{2 \eta} \left\| \thetab - \omdtheta{k}{h} \right\|_{\Bb_{k,h}}^2,
    \end{equation*}
    where $\tilde{\ell}_{k,h}(\thetab) = \ell_{k,h} (\omdtheta{k}{h}) + (\thetab - \omdtheta{k}{h})^\top \nabla \ell_{k,h}(\omdtheta{k}{h}) + \frac{1}{2} \left\| \thetab - \omdtheta{k}{h} \right\|_{\nabla^2 \ell_{k,h}(\omdtheta{k}{h})} \, .$
    We introduce the following lemma providing that the estimation error of the online estimator $\omdtheta{k}{h}$ can be bounded by the regret.
    \begin{lemma}[Lemma 12 in~\cite{zhang2023online}] \label{lemma:lemma 12 in zhang2023online}
        Let $\alpha = \log \Ucal + 2 (1 + \Pnorm \Fnorm)$ and $\lambda > 0$.
        If we set the step size $\eta = \alpha / 2 $, then we have
        \begin{equation}
            \begin{split}
                \left\| \omdtheta{k}{h} - \thetab^*_h \right\|_{\Bb_{k,h}}^2 
                & \leq \alpha  \sum_{i=1}^k \left( \ell_{i,h} (\thetab^*_h) - \ell_{i,h}(\omdtheta{i+1}{h}) \right)
                + \lambda \Pnorm^2
                \\
                & \phantom{{}={}}
                + 3 \sqrt{2} \Fnorm^3 \alpha \sum_{i=1}^k \left\| \omdtheta{i+1}{h} - \omdtheta{i}{h} \right\|_2^2
                - \sum_{i=1}^k \left\| \omdtheta{i+1}{h} - \omdtheta{i}{h} \right\|_{\Bb_{i,h}}^2 \, .  
            \end{split}
            \label{eq:online_parameter_gap_bound}
        \end{equation}
    \end{lemma}
    Now, we bound the first term of~\eqref{eq:online_parameter_gap_bound}.
    To simplify the presentation, for all $(k,h) \in [K] \times [H]$, we define the softmax function $\sigmab_{k,h}: \RR^{|\Scal_{k,h}|} \rightarrow [0,1]^{|\Scal_{k,h}|}$ as follows:
    \begin{align*}
        [\sigmab_{k,h}(\zb)]_{s'} = \frac{\exp([\zb]_{s'})}{ \sum_{s'' \in \Scal_{k,h}}   \exp([\zb]_{s''})}
        ,
    \end{align*}
    where $[\cdot]_{s'}$ denote the element corresponding to $s' \in \Scal$ of the input vector.
    We also define the pseudo-inverse of the softmax function $\sigmab_{k,h}$ via $[\sigmab_{k,h}^{+} (\pb)]_{s'} = \log ([\pb]_{s'})$ which has the property that for all $\pb \in \Delta_{|\Scal_{k,h}|}$, we have $\sigmab_{k,h} (\sigmab_{k,h}^{+} (\pb)) = \pb$ and $\sum_{s \in \Scal_{k,h}} \exp \left( [\sigmab_{k,h}^{+} (\pb)]_{s} \right) = 1$.
    
    We denote $\Phib_{k,h} = [\varphib_{k, h, s'} ]_{s' \in \Scal_{k,h}} \in \RR^{ d \times |\Scal_{k,h}|}$ for simplicity. 
    Then, the transition model can also be written as $P_{\thetab}( s' \mid s_h^k,a_h^k) = [\sigmab_{k,h} (  \Phib_{k,h}^\top  \thetab^*_h ) ]_{s'}$.
    We further define $\tilde{\zb}_{i,h} = \sigmab_{i,h}^+\left( \mathbb{E}_{\thetab \sim \mathcal{N} \left( \omdtheta{i}{h}, c \Bb_{i,h}^{-1} \right)} [\sigmab_{i,h}( \Phib_{i,h}^\top \thetab  )  ] \right)$ for our analysis.
    Then, we have
    \begin{align}
        \sum_{i=1}^k \left( \ell_{i,h} (\thetab^*_h) - \ell_{i,h}(\omdtheta{i+1}{h})\right)
        = \sum_{i=1}^k \Big( \ell_{i,h} (\thetab^*_h) - \ell (\tilde{\zb}_{i,h}, y^i_h) \Big)
        + \sum_{i=1}^k \left( \ell (\tilde{\zb}_{i,h}, y^i_h) -  \ell_{i,h}(\omdtheta{i+1}{h})\right).
        \label{eq:loss difference decompostion}
    \end{align}
    We can bound the first term of~\eqref{eq:loss difference decompostion} by the following lemma.
    \begin{lemma} \label{lemma:bound_(A)}
        Let $\delta \in (0,1]$. 
        Then, for all $ (k,h) \in [K]\times [H]$, with probability at least $1-\delta$, we have
        \begin{align*}
            \sum_{i=1}^k   \left(\ell_{i,h} (\thetab^*_h) - \ell (\tilde{\zb}_{i,h}, y^i_h)  \right) \leq \Gamma_{k}^A(\delta),
        \end{align*}
        where $ \Gamma_{k}^A(\delta)  =\frac{5}{4}(3\log (\Ucal k) + L_{\varphib} L_{\thetab}) \lambda + 4(3\log (\Ucal k) + L_{\varphib} L_{\thetab}) \log \left( \frac{H \sqrt{1 + 2k}}{\delta} \right) + 2$.
    \end{lemma}
    Furthermore, we can bound the second term of~\eqref{eq:loss difference decompostion} by the following lemma.
    \begin{lemma} \label{lemma:bound_(B)}
        Let $\lambda \geq 72 L_{\varphib}^2 c d $. Then, for any $c>0$ and all $(k,h) \in [K]\times [H]$, we have
        \begin{align*}
            \sum_{i=1}^k \left( \ell (\tilde{\zb}_{i,h}, y^i_h) -  \ell_{i,h}(\omdtheta{i+1}{h})\right)
            \leq \frac{1}{2c} \sum_{i=1}^k \left\| \omdtheta{i+1}{h} - \omdtheta{i}{h} \right\|_{\Bb_{i,h}}^2 + \Gamma_{k}^B(\delta) .
        \end{align*}
        where $\Gamma_{k}^B(\delta) = \sqrt{6} c d \log \left( 1 + \frac{2 k \Fnorm^2}{d \lambda} \right)$. 
    \end{lemma}
    Combining~\Cref{lemma:lemma 12 in zhang2023online},~\Cref{lemma:bound_(A)}, and~\Cref{lemma:bound_(B)}, and by setting $\eta = \alpha/2, c=2 \alpha / 3$ and $\lambda \geq \max \{12\sqrt{2} L_{\varphib}^3 \alpha, 48 L_{\varphib}^2 d \alpha \}$, we derive that
    \begin{align*}
        &\left\| \omdtheta{k+1}{h} - \thetab^*_h \right\|_{\Bb_{k,h}}^2  \\
        &\leq \alpha\Gamma_{k}^A(\delta) +  \alpha\Gamma_{k}^B(\delta) 
        + \lambda L_{\thetab}^2
        + 3 \sqrt{2} L_{\varphib}^3 \alpha  \sum_{i=1}^k \left\| \omdtheta{i+1}{h} - \omdtheta{i}{h} \right\|_2^2
        + \left( \frac{\alpha}{2c} -1 \right) \sum_{i=1}^k \left\|  \omdtheta{i+1}{h} - \omdtheta{i}{h} \right\|_{\Bb_{i,h}}^2
        \\
        &\leq \alpha\Gamma_{k}^A(\delta) +  \alpha\Gamma_{k}^B(\delta) 
        + \lambda L_{\thetab}^2 
        \\
        & \leq C \log \Ucal \left(  \lambda \log (\Ucal k)
        + \log (\Ucal k)  \log \left( \frac{H\sqrt{1 + 2k}}{\delta} \right) 
        + d \log \left( 1 + \frac{k}{d \lambda}\right) \right)
        + \lambda L_{\thetab}^2 
        \\
        & =: \beta_k (\delta)^2 \numberthis \label{eq:concentration result}
    \end{align*}
    where $C>0$ is an absolute constant.
    In the above, we choose $\lambda = \BigO(d \log \Ucal)$, $\alpha= \BigO(\log\Ucal) $.
    The second inequality of~\eqref{eq:concentration result} is derived from the fact that
    \begin{align*}
        &3 \sqrt{2} L_{\varphib}^3 \alpha  \sum_{i=1}^k \left\| \omdtheta{i+1}{h} - \omdtheta{i}{h} \right\|_2^2
        + \left( \frac{\alpha}{2c} -1 \right) \sum_{i=1}^k \left\|  \omdtheta{i+1}{h} - \omdtheta{i}{h} \right\|_{\Bb_{i,h}}^2 \\
        &= 3 \sqrt{2} L_{\varphib}^3 \alpha  \sum_{i=1}^k \left\| \omdtheta{i+1}{h} - \omdtheta{i}{h} \right\|_2^2
        -\frac{1}{4} \sum_{i=1}^k \left\|  \omdtheta{i+1}{h} - \omdtheta{i}{h} \right\|_{\Bb_{i,h}}^2  \\
        &\le 3 \sqrt{2} L_{\varphib}^3 \alpha  \sum_{i=1}^k \left\| \omdtheta{i+1}{h} - \omdtheta{i}{h} \right\|_2^2
        -\frac{\lambda}{4} \sum_{i=1}^k \left\|  \omdtheta{i+1}{h} - \omdtheta{i}{h} \right\|_2^2 \\
        &\le 0 .
    \end{align*}
    The first inequality holds from $\Bb_{i,h} \succeq \lambda \Ib_d$, and the second inequality is obvious from our setting of $\lambda$.
    Therefore, we can conclude that 
    \begin{align*}
        \left\| \omdtheta{k}{h} - \thetab^*_h \right\|_{\Bb_{k,h}} 
        \leq \beta_k (\delta) = \BigO(\sqrt{d} \log\Ucal \log(kH)) \, .
    \end{align*}
\end{proof}

In the following section, we provide the proofs of the lemmas used in Lemma~\ref{lemma:tight concentration}.
\subsubsection{Proof of Lemma~\ref{lemma:lemma 12 in zhang2023online}}
\begin{proof}[Proof of Lemma~\ref{lemma:lemma 12 in zhang2023online}]
    Let $\tilde{\ell}_{i,h}(\thetab) = \ell_{i,h} (\omdtheta{i}{h}) + \nabla \ell_{i,h}(\omdtheta{i}{h})^\top \left( \thetab - \omdtheta{i}{h} \right) + \frac{1}{2} \left\| \thetab - \omdtheta{i}{h} \right\|^2_{\nabla^2 \ell_{i,h}(\omdtheta{i}{h})}$ be a second-order Taylor expansion of $\ell_{i,h}(\thetab)$ at $\omdtheta{i}{h}$.
    Since we have
    \begin{equation*}
        \omdtheta{k+1}{h} 
        = \argmin_{\thetab \in \Bcal_d(\Pnorm)} \frac{1}{2 \eta} \left\| \thetab - \omdtheta{k}{h} \right\|_{\tilde{\Bb}_{k,h}}^2 + \nabla \ell_{k,h}(\omdtheta{k}{h})^\top \thetab
        = 
        \argmin_{\thetab \in \Bcal(\zero_d, \Pnorm)} \tilde{\ell}_{k,h}(\thetab) + \frac{1}{2 \eta} \left\| \thetab - \omdtheta{k}{h} \right\|_{\Bb_{k,h}}^2 \, ,
    \end{equation*}
    by Lemma~\ref{lemma:loss_firstorder_decomposition}, if we define $\psi (\thetab) = \frac{1}{2} \| \thetab \|^2_{\Bb_{i,h}}$ we obtain
    \begin{align}
        \nabla \tilde{\ell}_{i,h}(\omdtheta{i+1}{h})^\top (\omdtheta{i+1}{h} - \thetab^*_h)
        \leq \frac{1}{2\eta} \left( \left\| \omdtheta{i}{h} - \thetab^*_h \right\|_{\Bb_{i,h}}^2 - \left\| \omdtheta{i+1}{h} - \thetab^*_h \right\|_{\Bb_{i,h}}^2 - \left\| \omdtheta{i+1}{h} - \omdtheta{i}{h} \right\|_{\Bb_{i,h}} \right) .
        \label{eq:concentration_firstorder_bound}
    \end{align}    
    By applying Lemma~\ref{lemma:zhang_lemma1}, we have
    \begin{align}
        \ell_{i,h}(\omdtheta{i+1}{h}) - \ell_{i,h}(\thetab^*_h)
        \leq \left\langle \nabla \ell_{i,h}(\omdtheta{i+1}{h}), \omdtheta{i+1}{h} - \thetab^*_h  \right\rangle
        -\frac{1}{\alpha_{i,h}} \left\| \omdtheta{i+1}{h} - \thetab^*_h \right\|_{\nabla^2 \ell_{i,h}(\omdtheta{i+1}{h})},
        \label{eq:concentration_lossgap_bound}
    \end{align}
    where $\alpha_{i,h} =  \log |\Scal_{i,h}| + 2 (1 + L_{\varphib} L_{\thetab})$.

    By setting $\eta = \alpha_{i,h} / 2$ and merging equations~\eqref{eq:concentration_firstorder_bound} and~\eqref{eq:concentration_lossgap_bound}, we arrive at
    \begin{align*}
        \ell_{i,h}(\omdtheta{i+1}{h}) - \ell_{i,h}(\thetab^*_h)
        &\leq \left\langle \nabla  \ell_{i,h}(\omdtheta{i+1}{h})- \nabla\tilde{\ell}_{i,h}(\omdtheta{i+1}{h}) , \omdtheta{i+1}{h} - \thetab^*_h  \right\rangle
        \\
        &+ \frac{1}{\alpha_{i,h}} \left( \left\| \omdtheta{i}{h} - \thetab^*_h \right\|_{\Bb_{i,h}}^2 
        - \left\| \omdtheta{i+1}{h} - \thetab^*_h \right\|_{\Bb_{i+1,h}}^2 
        - \left\| \omdtheta{i+1}{h} - \omdtheta{i}{h} \right\|_{\Bb_{i,h}} \right).
        \numberthis \label{eq:loss decomposition}
    \end{align*}
    Meanwhile, we obtain
    \begin{equation}
        \label{eq:gradient of tilde ell}
        \nabla \tilde{\ell}_{i,h} (\thetab) = \nabla \ell_{i,h} (\omdtheta{i}{h}) + \nabla^2 \ell_{i,h} (\omdtheta{i}{h}) (\thetab - \omdtheta{i}{h}) \, 
    \end{equation}
    by taking the gradient over both sides of the Taylor approximation of $\ell_{i,h} (\thetab)$.
    Using~\eqref{eq:gradient of tilde ell}, we proceed to bound the first term of~\eqref{eq:loss decomposition} as follows:
    \begin{align*}
        \Big\langle & \nabla \ell_{i,h}(\omdtheta{i+1}{h}) - \nabla\tilde{\ell}_{i,h}(\omdtheta{i+1}{h}), \omdtheta{i+1}{h} - \thetab^*_h \Big\rangle \\
        &= \left\langle \nabla  \ell_{i,h}(\omdtheta{i+1}{h}) 
        - \nabla \ell_{i,h} (\omdtheta{i}{h}) 
        - \nabla^2 \ell_{i,h} (\omdtheta{i}{h}) (\omdtheta{i+1}{h} - \omdtheta{i}{h}) , \omdtheta{i+1}{h} - \thetab^* \right\rangle
        \\
        &= \left\langle D^3 \ell_{i,h}(\bar{\thetab}^{i+1}_h) \left[ \omdtheta{i+1}{h} - \omdtheta{i}{h} \right](\omdtheta{i+1}{h} - \omdtheta{i}{h})
        , \omdtheta{i+1}{h} - \thetab^* \right\rangle
        \\
        &\leq 3 \sqrt{2} L_{\varphib} \left\| \omdtheta{i+1}{h}  - \thetab^*_h \right\|_2 \left\| \omdtheta{i+1}{h} - \omdtheta{i}{h} \right\|_{\nabla^2 \ell_{i,h}(\bar{\thetab}^{i+1}_h )}^2
        \\
        &\leq 3 \sqrt{2} L_{\varphib} \left\| \omdtheta{i+1}{h} - \omdtheta{i}{h} \right\|^2_{\nabla^2 \ell_{i,h}( \bar{\thetab}^{i+1}_h )}
        \\
        &\leq 3 \sqrt{2} L_{\varphib}^3 \left\| \omdtheta{i+1}{h} - \omdtheta{i}{h} \right\|_2^2
    \end{align*}
    where $\bar{\thetab}^{i+1}_h$ is a convex combination of $\omdtheta{i}{h}$ and $\omdtheta{i+1}{h}$. The second equality arises from the Taylor expansion, the first inequality is due to the self-concordant property, and the final inequality is justified by the following:
    \begin{align*}
        &\nabla^2 \ell_{i,h}( \bar{\thetab}^{i+1}_h ) \\
        &= \sum_{s' \in \Scal_{i,h}} P_{ \bar{\thetab}^{i+1}_h }(s' \mid s_h^i,a_h^i) \varphib_{i,h,s'} \varphib_{i,h,s'}^\top 
        \\
        &\phantom{{}={}} - \sum_{s' \in \Scal_{i,h}} \sum_{s'' \in \Scal_{i,h}} P_{ \bar{\thetab}^{i+1}_h }(s' \mid s_h^i,a_h^i) P_{ \bar{\thetab}^{i+1}_h }(s'' \mid s_h^i,a_h^i) \varphib_{i,h,s'} \varphib_{i,h,s''}^\top 
        \\
        &\preceq \sum_{s' \in \Scal_{i,h}} P_{\bar{\thetab}^{i+1}_h}(s' \mid s_h^i,a_h^i) \varphib_{i,h,s'} \varphib_{i,h,s'}^\top \\
        &\preceq L_{\varphib}^2 \Ib_d.
    \end{align*}
    By summing over $i$ and reorganizing the terms, we arrive at the final result as follows:
    \begin{align*}
        &\left\| \omdtheta{k+1}{h} - \thetab^*_h \right\|_{\Bb_{k+1,h}}^2 \\
        &\leq \sum_{i=1}^k \alpha_{i,h} \left( \ell_{i,h} (\thetab^*_h) - \ell_{i,h}(\omdtheta{i+1}{h})\right)
        + \left\| \omdtheta{1}{h} - \thetab^*_h \right\|_{\Bb_{1,h}}^2
        \\
        & \phantom{{}={}} + 3 \sqrt{2} L_{\varphib}^3 \sum_{i=1}^k \alpha_{i,h} \left\| \omdtheta{i+1}{h} - \omdtheta{i}{h} \right\|_2^2 - \sum_{i=1}^k \left\|  \omdtheta{i+1}{h} - \omdtheta{i}{h} \right\|_{\Bb_{i,h}}^2
        \\
        &\leq \alpha \sum_{i=1}^k \left( \ell_{i,h} (\thetab^*_h) - \ell_{i,h}(\omdtheta{i+1}{h})\right)
        + \lambda L_{\thetab}^2
        + 3 \sqrt{2} L_{\varphib}^3 \alpha \sum_{i=1}^k \left\| \omdtheta{i+1}{h} - \omdtheta{i}{h} \right\|_2^2
        - \sum_{i=1}^k \left\|  \omdtheta{i+1}{h} - \omdtheta{i}{h} \right\|_{\Bb_{i,h}}^2 .
    \end{align*}    
    where the first inequality holds by Assumption~\ref{assm:bdd feature & param} and the last inequality holds since $\alpha = \log \Ucal + 2 (1 + L_{\varphib} L_{\thetab}) \ge \alpha_{i,h}$ for all $i \in [k]$.
\end{proof}

\subsubsection{Proof of Lemma~\ref{lemma:bound_(A)}  }
\label{app_subsec:proof_lemma:bound_(A)}
\begin{proof}[Proof of Lemma~\ref{lemma:bound_(A)}]
    The norm of $\tilde{\zb}_{i,h} = \sigmab_{i,h}^+\left( \mathbb{E}_{\thetab \sim \mathcal{N} \left( \omdtheta{i}{h}, c \Bb_{i,h}^{-1} \right)} [\sigmab_{i,h}( \Phib_{i,h}^\top \thetab  )  ] \right)$ is generally unbounded~\citep{foster2018logistic}.
    In this proof, we utilize the smoothed version of $\tilde{\zb}_{i,h}$,  defined as follows:
    \begin{align*}
        \tilde{\zb}^u_{i,h} = \sigmab_{i,h}^+\left( \operatorname{smooth}_{i,h}^u \mathbb{E}_{\thetab \sim \mathcal{N} \left( \omdtheta{i}{h}, c \Bb_{i,h}^{-1} \right)} [\sigmab_{i,h}( \Phib_{i,h}^\top \thetab  )  ] \right)    
    \end{align*}
    where the smooth function $ \operatorname{smooth}_{i,h}^u(\mathbf{p}) = (1-u)\mathbf{p} + (u/\Ucal) \mathbf{1}$ with $u \in [0, 1/2]$, and $\mathbf{1} \in \RR^{|\Scal_{i,h}|}$ is an all-one vector.
    
    Exploiting the property of $\sigmab^+_{i,h}$ such that $\sigmab_{i,h} ( \sigmab_{i,h}^{+}(\mathbf{p})) = \mathbf{p}$ for any $\mathbf{p} \in \Delta_{|\Scal_{i,h}|}$, it is straightforward to show that $\tilde{\zb}_{i,h}^u =\sigmab^+_{i,h} ( \operatorname{smooth}_{i,h}^u( \sigmab_{i,h} (\tilde{\zb}_{i,h})))$.
    Then, by Lemma~\ref{lemma:zhang_lemma17}, we have
    \begin{align*}
        \sum_{i=1}^k \ell (\tilde{\zb}_{i,h}^u, y_{h}^i)
        - \sum_{i=1}^k \ell (\tilde{\zb}_{i,h}, y_{h}^i)
        \leq 2 u k,
        \quad
        \text{and}
        \quad
        \|\tilde{\zb}_{i,h}^u \|_{\infty} \leq \log (\Ucal / u).
        \numberthis \label{eq:I2_A_first_bound}
    \end{align*}
    Given the definition of $\ell_{i,h}$, we know that $\ell(\zb_{i,h}^*, y_{h}^i )  =\ell_{i,h} (\thetab_h^*) $, where $\zb_{i,h}^* = \Phib_{i,h}^\top \thetab_h^*$.
    We can bound the gap between the loss of $\thetab^*_h$ and $\tilde{\zb}_{i,h}^u$ as follows:
    \begin{align*}
        &\sum_{i=1}^k \left(\ell_{i,h} (\thetab^*_h) - \ell(\tilde{\zb}_{i,h}^u, y_{h}^i)  \right) 
        \\
        &= \sum_{i=1}^k \left(\ell(\zb_{i,h}^*, y_{h}^i ) - \ell(\tilde{\zb}_{i,h}^u, y_{h}^i)  \right)
        \\
        &\leq  \sum_{i=1}^k  \langle \nabla_z\ell(\zb_{i,h}^*, y_{h}^i ), \zb_{i,h}^* - \tilde{\zb}_{i,h}^u \rangle
        - \sum_{i=1}^k \frac{1}{M_{i,h}} \| \zb_{i,h}^*- \tilde{\zb}_{i,h}^u\|_{\nabla_z^2 \ell(\zb_{i,h}^*, y_{h}^i ) }^2
        \\
        &= \sum_{i=1}^k  \langle \nabla_z\ell(\zb_{i,h}^*, y_{h}^i ), \zb_{i,h}^* - \tilde{\zb}_{i,h}^u \rangle
        - \sum_{i=1}^k \frac{1}{M_{i,h}} \| \zb_{i,h}^*- \tilde{\zb}_{i,h}^u\|_{\nabla \sigmab_{i,h} (\zb_{i,h}^*) }^2
        ,
        \numberthis  \label{eq:I2_A_intermediate_term}
    \end{align*}
    where $M_{i,h} = \log (|\Scal_{i,h}|) + 2\log (\Ucal/u) $, and the second equality holds by a  direct calculation of the first order and Hessian of the logistic loss.

    Now, we first bound the first term of the right-hand side.
    Let $\mathbf{d}_{i,h} = (\zb_{i,h}^* - \tilde{\zb}_{i,h}^u )/(M + L_{\varphib} L_{\thetab})$, where $M =  \log \Ucal + 2\log (\Ucal/u)$.
    Then, one can check that $\|\mathbf{d}_{i,h} \|_{\infty} \leq 1$ since $\| \zb_{i,h}^* \|_{\infty} \leq \max_{s' \in \Scal_{i,h}} \| \varphib_{i,h, s'} \|_2 \| \thetab_h^* \|_2 \leq L_{\varphib} L_{\thetab}$ and $\| \tilde{\zb}_{i,h}^u \|_{\infty} \leq \log (\Ucal /u)$.
    Moreover, since $\zb_{i,h}^* $ and $ \tilde{\zb}_{i,h}^u $ are independent of $y_{h}^i$, $\mathbf{d}_{i,h}$ is $\mathcal{F}_{i,h}$-measurable.
    Since $\EE [ (\sigmab_{i,h} (\zb_{i,h}^*) - y_{h}^i) (\sigmab_{i,h} (\zb_{i,h}^*) - y_{h}^i)^\top \mid \mathcal{F}_{i,h}] = \nabla \sigmab_{i,h} (\zb_{i,h}^*)$ and $\| \sigmab_{i,h} (\zb_{i,h}^*) - y_{h}^i\|_1 \leq 2$, we can apply~\Cref{lemma:zhang_lemma15}. For any $k$ and $\delta \in (0,1]$, with probability at least $1-\delta / H$, we have
    \begin{align*}
         &\sum_{i=1}^k  \langle \nabla_z\ell(\zb_{i,h}^*, y_{h}^i ), \zb_{i,h}^* - \tilde{\zb}_{i,h}^u \rangle
         \\
         &= (M + L_{\varphib} L_{\thetab}) \sum_{i=1}^k \langle \nabla_z\ell(\zb_{i,h}^*, y_{h}^i ), \mathbf{d}_{i,h} \rangle 
         \\
         &\leq (M + L_{\varphib} L_{\thetab}) \sqrt{\lambda + \sum_{i=1}^k \| \mathbf{d}_{i,h} \|_{\nabla \sigmab_{i,h} (\zb_{i,h}^*)}^2 }
         \\
         &\phantom{{}={}} \cdot \sqrt{\frac{\sqrt{\lambda}}{4} + \frac{4}{\sqrt{\lambda}}\log \left( \frac{ H \sqrt{1 +  \frac{1}{\lambda}\sum_{i=1}^k \| \mathbf{d}_{i,h} \|_{\nabla \sigmab_{i,h} (\zb_{i,h}^*)}^2}}{\delta} \right)  }
         \\
         &\leq  (M + L_{\varphib} L_{\thetab}) \sqrt{\lambda + \sum_{i=1}^k \| \mathbf{d}_{i,h} \|_{\nabla \sigmab_{i,h} (\zb_{i,h}^*)}^2 }
         \sqrt{\frac{\sqrt{\lambda}}{4} + 4 \log \left( \frac{ H \sqrt{1 + 2k}}{\delta} \right)   },
         \numberthis  \label{eq:I2_A_self_normal}
    \end{align*}
    where the second inequality holds since $ \| \mathbf{d}_{i,h} \|_{\nabla \sigmab_{i,h} (\zb_{i,h}^*)}^2 =\mathbf{d}_{i,h}^\top \nabla \sigmab_{i,h} (\zb_{i,h}^*) \mathbf{d}_{i,h} \leq 2$ and  $\lambda \geq 1$.
    Plugging~\eqref{eq:I2_A_self_normal} into~\eqref{eq:I2_A_intermediate_term} and rearranging the term, we get
    \begin{align*}
        &\sum_{i=1}^k \left(\ell_{i,h} (\thetab^*) - \ell(\tilde{\zb}_{i,h}^u, y_{h}^i)  \right)
        \\
        &\leq  (M + L_{\varphib} L_{\thetab}) \sqrt{\lambda + \sum_{i=1}^k \| \mathbf{d}_{i,h} \|_{\nabla \sigmab_{i,h} (\zb_{i,h}^*)}^2 }
         \sqrt{\frac{\sqrt{\lambda}}{4} + 4 \log \left( H \frac{\sqrt{1 + 2k}}{\delta} \right)   }
        \\
        &\phantom{{}={}} - \sum_{i=1}^k \frac{1}{M_{i,h}} \| \zb_{i,h}^*- \tilde{\zb}_{i,h}^u\|_{\nabla \sigmab_{i,h} (\zb_{i,h}^*) }^2
        \\
        &\leq  (M + L_{\varphib} L_{\thetab}) \sqrt{\lambda + \sum_{i=1}^k \| \mathbf{d}_{i,h} \|_{\nabla \sigmab_{i,h} (\zb_{i,h}^*)}^2 }
         \sqrt{\frac{\sqrt{\lambda}}{4} + 4 \log \left( \frac{ H \sqrt{1 + 2k}}{\delta} \right)   }
        \\
        &\phantom{{}={}} - (M + L_{\varphib} L_{\thetab})\sum_{i=1}^k  \| \mathbf{d}_{i,h} \|_{\nabla \sigmab_{i,h} (\zb_{i,h}^*) }^2
         \\
         &\leq  (M + L_{\varphib} L_{\thetab}) \left(  \lambda + \sum_{i=1}^k \| \mathbf{d}_{i,h} \|_{\nabla \sigmab_{i,h} (\zb_{i,h}^*)}^2 \right)
         +  (M + L_{\varphib} L_{\thetab}) \left( \frac{\sqrt{\lambda}}{4} + 4 \log \left( \frac{ H \sqrt{1 + 2k}}{\delta} \right) \right) \\
         &\phantom{{}={}}- (M + L_{\varphib} L_{\thetab})\sum_{i=1}^k  \| \mathbf{d}_{i,h} \|_{\nabla \sigmab_{i,h} (\zb_{i,h}^*) }^2
         \\
         &\leq \frac{5}{4}(M + L_{\varphib} L_{\thetab}) \lambda + 4 (M + L_{\varphib} L_{\thetab})  \log \left( \frac{ H \sqrt{1 + 2k}}{\delta} \right).
         \numberthis \label{eq:I2_A_second_bound}
    \end{align*}
    Finally, combining~\eqref{eq:I2_A_first_bound} and~\eqref{eq:I2_A_second_bound}, by setting $u = 1/k$, we derive that
    \begin{align*}
        &\sum_{i=1}^k \left(\ell_{i,h} (\thetab^*_h) - \ell (\tilde{\zb}_{i,h}, y^i_h)  \right)
        \\
        &\leq \frac{5}{4}(M + L_{\varphib} L_{\thetab}) \lambda + 4 (M + L_{\varphib} L_{\thetab})  \log \left( \frac{ H \sqrt{1 + 2k}}{\delta} \right) + 2uk
        \\
        &\leq \frac{5}{4}(3\log (\Ucal k) + L_{\varphib} L_{\thetab}) \lambda 
        + 4(3\log (\Ucal k) + L_{\varphib} L_{\thetab}) \log \left( \frac{ H \sqrt{1 + 2k}}{\delta} \right) + 2
    \end{align*}
    where the last inequality holds by the definition of $M =  \log \Ucal + 2\log (\Ucal/u)$.
    Taking the union bound over $h\in [H]$, we conclude the proof. 
\end{proof}
\subsubsection{Proof of Lemma~\ref{lemma:bound_(B)}  }
\label{app_subsec:proof_lemma:bound_(B)}
\begin{proof}[Proof of Lemma~\ref{lemma:bound_(B)}]
    We start the proof from the observation of Proposition 2 in~\citet{foster2018logistic}, stating that $\tilde{\zb}_{i,h}$ represents the mixed prediction, which adheres to the following property:
    \begin{align}
        \ell(\tilde{\zb}_{i,h}, y^i_h)
        \leq -\log \left( \EE_{\thetab \sim  \mathcal{N} \left(\omdtheta{i}{h}, c \Bb_{i,h}^{-1} \right)}\left[ \exp \left( -\ell_{i,h}(\thetab) \right) \right] \right)
        = -\log \left( \frac{1}{Z_{i,h}} \int_{\RR^d} \exp \left( -L_{i,h}(\thetab) \right) \diff \thetab \right),
        \label{eq:ztilde_loss_upper_ln}
    \end{align}
    where $L_{i,h}(\thetab) := \ell_{i,h}(\thetab) + \frac{1}{2c} \left\| \thetab - \omdtheta{i}{h} \right\|_{\Bb_{i,h}}^2$ and $Z_{i,h} := \sqrt{(2\pi)^d c |\Bb_{i,h}^{-1}|} \,$.

    Consider the quadratic approximation
    \begin{align*}
        \tilde{L}_{i,h}(\thetab) = L_{i,h}(\omdtheta{i+1}{h}) 
        + \left\langle \nabla L_{i,h}(\omdtheta{i+1}{h}) , \thetab - \omdtheta{i+1}{h} \right\rangle
        + \frac{1}{2c} \left\| \thetab - \omdtheta{i+1}{h} \right\|_{\Bb_{i,h}}^2.
    \end{align*}
    Using the property that $\ell_{i,h}$ is $3 \sqrt{2} L_{\varphib}$-self-concordant-like function as asserted by Proposition B.1 in~\citep{lee2024nearly}, and applying~\Cref{lemma:zhang_lemma18}, we obtain
    \begin{align*}
          L_{i,h}(\thetab) \leq \tilde{L}_{i,h}(\thetab) + \exp \left( 18 L_{\varphib}^2 \left\| \thetab - \omdtheta{i+1}{h} 
          \right\|_2^2 \right) \left\| \thetab - \omdtheta{i+1}{h} \right\|_{\nabla \ell_{i,h}(\omdtheta{i+1}{h})}^2.   
    \end{align*}
    Also, we have
    \begin{align*}
        & \frac{1}{Z_{i,h}} \int_{\RR^d} \exp(-L_{i,h}(\thetab)) \diff \thetab \\
        &\geq \frac{1}{Z_{i,h}} \int_{\RR^d} \exp \left( -\tilde{L}_{i,h}(\thetab) - \exp \left( 18 L_{\varphib}^2 \left\| \thetab - \omdtheta{i+1}{h} \right\|_2^2 \right) \left\| \thetab - \omdtheta{i+1}{h} \right\|_{\nabla \ell_{i,h}(\omdtheta{i+1}{h})}^2 \right) \diff \thetab \\
        &= \frac{\exp\left( -L_{i,h}(\omdtheta{i+1}{h}) \right)}{Z_{i,h}} \int_{\RR^d} \tilde{f}_{i+1,h}(\thetab) \cdot \exp \left( -\left\langle \nabla L_{i,h} (\omdtheta{i+1}{h}), \thetab - \omdtheta{i+1}{h} \right\rangle \right) \diff \thetab,
        \numberthis \label{eq:exp_loss_lowerbound}
    \end{align*}
    where we define the function $\tilde{f}_{i,h}: \mathcal{B}(\mathbf{0}_d, 1) \rightarrow \RR$ as
    \begin{align*}
        \tilde{f}_{i+1,h} (\thetab)
        = \exp\left( -\frac{1}{2c} \left\| \thetab - \omdtheta{i+1}{h} \right\|_{\Bb_{i,h}}^2 -  \exp \left( 18 L_{\varphib}^2 \left\| \thetab - \omdtheta{i+1}{h} \right\|_2^2 \right) \left\| \thetab - \omdtheta{i+1}{h} \right\|_{\nabla^2 \ell_{i,h}(\omdtheta{i+1}{h})}^2   \right).
    \end{align*}
    We denote $\tilde{Z}_{i+1,h} = \int_{\RR^d} \tilde{f}_{i+1,h} (\thetab)  \diff \thetab \leq + \infty$ and define $\tilde{\Theta}_{i+1,h}$ as the distribution whose density function is $\tilde{f}_{i+1,h} (\thetab)/\tilde{Z}_{i+1,h} $.
    Then, we can rewrite~\eqref{eq:exp_loss_lowerbound} as follows:
    \begin{align*}
        &\frac{1}{Z_{i,h}} \int_{\RR^d} \exp(-L_{i,h}(\thetab)) \diff \thetab \\
        &\geq \frac{\exp\left( -L_{i,h}(\omdtheta{i+1}{h}) \right) \tilde{Z}_{i+1,h}}{Z_{i,h}} \EE_{\thetab \sim \tilde{\Theta}_{i+1,h}} \left[ \exp\left( -\left\langle \nabla L_{i,h} (\omdtheta{i+1}{h}), \thetab - \omdtheta{i+1}{h} \right\rangle \right) \right]
        \\
        &\geq \frac{\exp\left( -L_{i,h}(\omdtheta{i+1}{h}) \right) \tilde{Z}_{i+1,h}}{Z_{i,h}}
        \exp \left( - \EE_{\thetab \sim \tilde{\Theta}_{i+1,h}} \left[\left\langle \nabla L_{i,h} (\omdtheta{i+1}{h}), \thetab - \omdtheta{i+1}{h} \right\rangle \right]   \right)
        \\
        &= \frac{\exp\left( -L_{i,h}(\omdtheta{i+1}{h}) \right) \tilde{Z}_{i+1,h}}{Z_{i,h}},
        \numberthis \label{eq:exp_loss_lowerbound2}
    \end{align*}
    where the second inequality is by Jensen's inequality and the last inequality holds because $\tilde{\Theta}_{i+1,h}$ is symmetric around $\omdtheta{i+1}{h}$ and thus $\EE_{\thetab \sim \tilde{\Theta}_{i+1,h}} \left[\left\langle \nabla L_{i,h} (\omdtheta{i+1}{h}), \thetab - \omdtheta{i+1}{h} \right\rangle \right] = 0$.

    Combining~\eqref{eq:ztilde_loss_upper_ln} and~\eqref{eq:exp_loss_lowerbound2}, we get
    \begin{align}
        \ell_{i,h}(\tilde{\zb})
        \leq L_{i,h}(\omdtheta{i+1}{h}) + \log Z_{i,h} - \log \tilde{Z}_{i+1,h}.
\label{eq:ztilde_loss_upper_ln_intermediate}
    \end{align}
    Moreover, we have
    \begin{align*}
        &- \log \tilde{Z}_{i+1,h} \\
        &= -\log \left( \int_{\RR^d}  \exp\left( -\frac{1}{2c} \left\| \thetab - \omdtheta{i+1}{h} \right\|_{\Bb_{i,h}}^2 \right. \right.
        \\
        &\phantom{{}={}={}={}} \left. \left. - \exp \left(18 L_{\varphib}^2 \left\| \thetab - \omdtheta{i+1}{h} \right\|_2^2 \right) \left\| \thetab - \omdtheta{i+1}{h} \right\|_{\nabla^2 \ell_{i,h}(\omdtheta{i+1}{h})}^2 \right) \diff \thetab \right)
        \\
        &= -\log \left( \widehat{Z}_{i+1,h} \cdot \EE_{\thetab \sim \widehat{\Theta}_{i+1,h} } \left[ \exp \left( - \exp \left(18 L_{\varphib}^2 \left\| \thetab - \omdtheta{i+1}{h} \right\|_2^2 \right) \left\| \thetab - \omdtheta{i+1}{h} \right\|_{\nabla^2 \ell_{i,h}(\omdtheta{i+1}{h})}^2  \right)  \right]  \right)
        \\
        &\leq -\log  \widehat{Z}_{i+1,h} + \EE_{\thetab \sim \widehat{\Theta}_{i+1,h} }  \left[ \exp \left(18 L_{\varphib}^2 \left\| \thetab - \omdtheta{i+1}{h} \right\|_2^2 \right) \left\| \thetab - \omdtheta{i+1}{h} \right\|_{\nabla^2 \ell_{i,h}(\omdtheta{i+1}{h})}^2  \right]
        \\
        &=  -\log  Z_{i,h} + \EE_{\thetab \sim \widehat{\Theta}_{i+1,h} }  \left[  \exp \left(18 L_{\varphib}^2 \left\| \thetab - \omdtheta{i+1}{h} \right\|_2^2 \right) \left\| \thetab - \omdtheta{i+1}{h} \right\|_{\nabla^2 \ell_{i,h}(\omdtheta{i+1}{h})}^2  \right],
        \numberthis \label{eq:ztilde_loss_upper_ln_intermediate2}
    \end{align*}
    where $ \widehat{\Theta}_{i+1,h}  = \mathcal{N}(\omdtheta{i+1}{h}, c \Bb_{i,h}^{-1})$ and $\widehat{Z}_{i+1,h} = \bigintss_{\RR^d} \exp\left(  -\frac{1}{2c} \left\| \thetab - \omdtheta{i+1}{h} \right\|_{\Bb_{i,h}}^2 \right) \diff \thetab$, and the last inequality holds because $\widehat{Z}_{i+1,h}$ and $Z_{i,h}$ are identical normalizing factors.
    Integrating~\eqref{eq:ztilde_loss_upper_ln_intermediate} and~\eqref{eq:ztilde_loss_upper_ln_intermediate2} and summing over $k$, yields
    \begin{align*}
        &\sum_{i=1}^k \ell(\tilde{\zb}_{i,h}, y^i_h)
        \\
        & = \sum_{i=1}^k L_{i,h}(\omdtheta{i+1}{h})
        + \sum_{i=1}^k \EE_{\thetab \sim \widehat{\Theta}_{i+1,h} }  \left[  \exp \left( 18 L_{\varphib}^2 \left\| \thetab - \omdtheta{i+1}{h} \right\|_2^2 \right) \left\| \thetab - \omdtheta{i+1}{h} \right\|_{\nabla^2 \ell_{i,h}(\omdtheta{i+1}{h})}^2  \right] \,.
    \end{align*}
    Moreover, we can further bound the second term on the right-hand side of~\eqref{eq:ztilde_loss_upper_ln_intermediate2}.
    By Cauchy-Schwarz inequality, we get
    \begin{align*}
        &\EE_{\thetab \sim \widehat{\Theta}_{i+1,h} }  \left[  \exp \left( 18 L_{\varphib}^2 \left\| \thetab - \omdtheta{i+1}{h} \right\|_2^2 \right) \left\| \thetab - \omdtheta{i+1}{h} \right\|_{\nabla^2 \ell_{i,h}(\omdtheta{i+1}{h})}^2  \right] \\
        \leq &\underbrace{\sqrt{ \EE_{\thetab \sim \widehat{\Theta}_{i+1,h} }  \left[  \exp \left( 36 L_{\varphib}^2 \left\| \thetab - \omdtheta{i+1}{h} \right\|_2^2 \right) \right] }}_{\mathrm{(I)}}
        \underbrace{\sqrt{  \EE_{\thetab \sim \widehat{\Theta}_{i+1,h} }  \left[  \left\| \thetab - \omdtheta{i+1}{h} \right\|_{\nabla^2 \ell_{i,h}(\omdtheta{i+1}{h})}^4  \right] }}_{\mathrm{(II)}}.
    \end{align*}
    Since $\widehat{\Theta}_{i+1,h} = \mathcal{N} \left(\omdtheta{i+1}{h}, c \Bb_{i,h}^{-1} \right)$, $\thetab - \omdtheta{i+1}{h}$ follows the same distribution as 
    \begin{align}
        \sum_{j=1}^d \sqrt{c \lambda_j \left( \Bb_{i,h}^{-1} \right)} X_j \eb_j,
        \quad
        \text{where }
        X_j \stackrel{i.i.d.}{\sim} \mathcal{N}(0,1), 
        \forall j \in [d],
        \label{eq:basis_expression}
    \end{align}
    where $\lambda_j \left( \Bb_{i,h}^{-1} \right)$  denotes the $j$-th largest eigenvalue of $\Bb_{i,h}^{-1}$ and $\{\eb_1, \dots, \eb_d\}$ are orthogonal basis of $\RR^d$.
    Furthermore, since we know that $\Bb_{i,h}^{-1} \leq \lambda^{-1} \Ib_d$, we can bound the term $\mathrm{(I)}$ by
    \begin{align*}
        \mathrm{(I)}
        &\leq \sqrt{\EE_{X_j} \left[  \prod_{j=1}^d \exp \left(36 L_{\varphib}^2 c\lambda^{-1} X_j^2 \right) \right] }
        = \sqrt{ \prod_{j=1}^d\EE_{X_j} \left[ \exp \left( 36 L_{\varphib}^2 c\lambda^{-1} X_j^2 \right)  \right] } \\
        &\leq \left( \EE_{W \sim \chi^2} \left[ \exp \left( 36 L_{\varphib}^2 c \lambda^{-1} W \right) \right] \right)^{\frac{d}{2}}
        \leq  \EE_{W \sim \chi^2} \left[ \exp \left( 18 L_{\varphib}^2 c \lambda^{-1} W d \right) \right] 
    \end{align*}
    where $\chi^2$ is the chi-square distribution and the last inequality holds due to Jensen's inequality.
    By choosing $\lambda \geq 72 L_{\varphib}^2 c d$, we arrive that
    \begin{align}
        \mathrm{(I)}
        \leq \EE_{W \sim \chi^2} \left[ \exp \left( \frac{W}{4} \right) \right] 
        \leq \sqrt{2},
        \label{eq:term (a-1)}
    \end{align}
    where the last inequality holds because the moment-generating function for $\chi^2$-distribution is bounded by $\EE_{W \sim \chi^2} [ \exp (tW) ] \leq 1/\sqrt{1-2t} $ for all $t \leq 1/2$.
    Now, we bound the term $\mathrm{(II)}$.
    \begin{align*}
        \mathrm{(II)}
        &=\sqrt{  \EE_{\thetab \sim \widehat{\Theta}_{i+1,h} }  \left[  \left\| \thetab - \omdtheta{i+1}{h} \right\|_{\nabla^2 \ell_{i,h}(\omdtheta{i+1}{h})}^4  \right] }
        = \sqrt{  \EE_{\thetab \sim  \mathcal{N} \left( 0, c \Bb_{i,h}^{-1} \right) }  \left[  \| \thetab  \|_{\nabla^2 \ell_{i,h}(\omdtheta{i+1}{h})}^4  \right] }
        \\ &= \sqrt{  \EE_{\thetab \sim  \mathcal{N} \left( 0, c \bar{\Bb}_{i,h}^{-1} \right) } \left[ \| \thetab \|_2^4  \right] },
    \end{align*}
    where $\bar{\Bb}_{i,h} = \left( \nabla^2 \ell_{i,h}(\omdtheta{i+1}{h}) \right)^{-1/2} \Bb_{i,h} \left( \nabla^2 \ell_{i,h}(\omdtheta{i+1}{h}) \right)^{-1/2}$.
    Let $\bar{\lambda}_j := \lambda_j \left( c \bar{\Bb}^{-1}_{i,h} \right)$ be the $j$-th largest eigenvalue of the matrix.
    Then, a similar analysis as~\eqref{eq:basis_expression} gives that
    \begin{align*}
        \mathrm{(II)}
        &= \sqrt{\EE_{X_j \sim \mathcal{N}(0,1)} \left[ \left\| \sum_{j=1}^d \sqrt{\bar{\lambda}_j} X_j \eb_j  \right\|_2^4 \right] }
        = \sqrt{\EE_{X_j \sim \mathcal{N}(0,1)} \left[ \left( \sum_{j=1}^d \bar{\lambda}_j X_j^2\right)^2 \right] }
        \\
        &= \sqrt{ \sum_{j=1}^d \sum_{j'=1}^d \bar{\lambda}_j \bar{\lambda}_{j'} \EE_{X_j, X_{j'} \sim \mathcal{N}(0,1)} [X_j^2 X_{j'}^2] }
        \leq \sqrt{3 \sum_{j=1}^d \sum_{j'=1}^d \bar{\lambda}_j \bar{\lambda}_{j'}}
        = \sqrt{3}c \tr \left( \bar{\Bb}_{i,h}^{-1} \right),
    \end{align*}
    where the last inequality holds due to $\EE_{X_j, X_{j'} \sim \mathcal{N}(0,1)} [X_j^2 X_{j'}^2] \leq 3$ when considering the case where $j=j'$ and the last equality is derived from the fact that $\left( \sum_{j=1}^d \bar{\lambda}_j \right)^2 = \tr \left(c \bar{\Bb}_{i,h}^{-1} \right)$.
    Here, we denote $\tr(A)$ as the trace of the matrix $A$.

    We define matrix $\Rb_{i+1,h} := \lambda \Ib_d/2 + \sum_{\tau=1}^i \nabla^2 \ell_{\tau,h} (\thetab_{\tau+1,h})$.
    Under the condition $\lambda \geq 2 L_{\varphib}^2$, we have $\nabla^2 \ell_{i,h} (\thetab_{i+1, h}) \preceq L_{\varphib}^2 \Ib_d \leq \frac{\lambda}{2} \Ib_d $.
    Then, we have $\Bb_{i,h} \succeq \Rb_{i+1,h}$.
    Therefore, we can bound the trace by
    \begin{align*}
         \tr \left( \bar{\Bb}_{i,h}^{-1} \right)
         &= \tr\left( \Bb_{i,h}^{-1} \nabla^2 \ell_{i,h}(\thetab_{i+1,h}) \right)
         \leq \tr \left( \Rb_{i+1,h}^{-1} \nabla^2 \ell_{i,h}(\thetab_{i+1,h}) \right) 
         \\
         &= \tr  \left( \Rb_{i+1,h}^{-1} (\Rb_{i+1,h}- \Rb_{i,h}) \right)
         \leq \log \frac{\operatorname{det}(\Rb_{i+1,h})}{\operatorname{det}(\Rb_{i,h})},
    \end{align*}
    where the last inequality holds due to Lemma 4.7 of~\citet{hazan2016introduction}.
    Therefore we can bound the term $\mathrm{(II)}$ as
    \begin{align}
        \mathrm{(II)} \leq \sqrt{3}c \log \frac{\operatorname{det}(\Rb_{i+1,h})}{\operatorname{det}(\Rb_{i,h})}.
        \label{eq:term (a-2)}
    \end{align}
    Combining~\eqref{eq:term (a-1)} and~\eqref{eq:term (a-2)}, we get
    \begin{align}
        \EE_{\thetab \sim \widehat{\Theta}_{i+1,h} }  \left[  \exp\left( {6\left\| \thetab - \omdtheta{i+1}{h} \right\|_2^2 } \right) \left\| \thetab - \omdtheta{i+1}{h} \right\|_{\nabla^2 \ell_{i,h}(\omdtheta{i+1}{h})}^2  \right]
        \leq \sqrt{6}c \log \frac{\operatorname{det}(\Rb_{i+1,h})}{\operatorname{det}(\Rb_{i,h})}.
        \label{eq:term (a-1) and (a-2)}
    \end{align}
    Plugging~\eqref{eq:ztilde_loss_upper_ln_intermediate2} and~\eqref{eq:term (a-1) and (a-2)} into~\eqref{eq:ztilde_loss_upper_ln_intermediate}, and taking summation over $k$, we derive that
    \begin{align*}
        \sum_{i=1}^k \ell (\tilde{\zb}_{i,h}, y^i_h)
        &\leq \sum_{i=1}^k L_{i,h}(\omdtheta{i+1}{h}) +  \sqrt{6}c \sum_{i=1}^k \log \frac{\operatorname{det}(\Rb_{i+1,h})}{\operatorname{det}(\Rb_{i,h})}
        \\
        &=\sum_{i=1}^k \left( \ell_{i,h}(\omdtheta{i+1}{h}) + \frac{1}{2c} \left\| \omdtheta{i+1}{h} - \omdtheta{i}{h} \right\|_{\Bb_{i,h}}^2 \right) +  \sqrt{6}c \sum_{i=1}^k \log \frac{\operatorname{det}(\Rb_{i+1,h})}{\operatorname{det}(\Rb_{i,h})}
        \\
        &\leq \sum_{i=1}^k \left( \ell_{i,h}(\omdtheta{i+1}{h}) + \frac{1}{2c} \left\| \omdtheta{i+1}{h} - \omdtheta{i}{h} \right\|_{\Bb_{i,h}}^2 \right) + \sqrt{6}cd \log \left( 1 + \frac{2 k \Fnorm^2}{d \lambda} \right),
    \end{align*}
    where the last inequality holds because $ \sum_{i=1}^k \log \frac{\operatorname{det}(\Rb_{i+1,h})}{\operatorname{det}(\Rb_{i,h})} = \log ( \operatorname{det}(\Rb_{k+1,h}) /\operatorname{det}(\lambda/2 \Ib_d) ) \leq d \log \left( 1 + \frac{2 k \Fnorm^2}{d \lambda} \right)$.
    By rearranging the terms, we conclude the proof.
\end{proof}


\subsection{Bound on Prediction Error} \label{appx:prediction error of alg 2}
In this section, we present the bound on the prediction error of parameters updated by $\texttt{ORRL-MNL}$.
First, we compare the problem setting of MNL contextual bandits with ours and introduce the challenges of applying their analysis to our setting.

\paragraph{MNL dynamic assortment optimization (single-parameter \& uniform reward)~\cite{perivier2022dynamic}}
~\citet{perivier2022dynamic} consider an assortment selection problem where the user choice is given by a MNL choice model with the single-parameter. 
At each time $t$, the agent observes context features $\{ \xb_{t,i} \}_{i=1}^M \subset \RR^d$. 
Then the agent decides on the set $S_t \subset [M]$ to offer to a user, with $|S_t| \le N$. 
Without loss of generality, we may assume $|S_t| = N$.
Then the user purchases one single product $j \in S_t \cup \{0\}$ and the probability of each product $j$ is purchased by a user follows the MNL model parametrized by a unknown fixed parameter $\thetab^* \in \RR^d$,
\begin{equation*}
    q_{t,j}(S_t, \thetab^*) := 
    \begin{cases}
        \frac{\exp(\xb_{t,j}^\top \thetab^*)}{1 + \sum_{k \in S_t} \exp(\xb_k^\top \thetab^*)} & \quad \text{ if } j \in S_t
        \\
        \frac{1}{1 + \sum_{k \in S_t} \exp(\xb_k^\top \thetab^*)} & \quad \text{ if } j =0 \, .
    \end{cases}    
\end{equation*}
Then the difference between the revenue induced by $\thetab^*$ and that by an estimator $\thetab$ in~\citet{perivier2022dynamic} is expressed as follows:
\begin{equation} \label{eq:goyal revenue difference}
    \sum_{j \in S_t} q_{t,j}(S_t, \thetab^*) - \sum_{j \in S_t} q_{t,j}(S_t, \thetab) \, .
\end{equation}
If we define $Q: \RR^N \rightarrow \RR$, such that for all $\ub = (u_1, \ldots, u_N) \in \RR^N$, $Q(\ub) := \sum_{i=1}^N \frac{\exp(u_i)}{1 + \sum_{j=1}^N \exp(u_j)}$ and let $\vb^* = (\xb_{t,i_1}^\top \thetab^*, \ldots, \xb_{t,i_N}^\top \thetab^*)$ and $\vb = (\xb_{t,i_1}^\top \thetab, \ldots, \xb_{t,i_N}^\top \thetab)$, then Eq.~\eqref{eq:goyal revenue difference} can be expressed as follows:
\begin{align}
    \sum_{j \in S_t} q_{t,j}(S_t, \thetab^*) - \sum_{j \in S_t} q_{t,j}(S_t, \thetab)
    & = Q(\vb^*) - Q(\vb)
    \nonumber
    \\
    & = \nabla Q(\vb^*)^\top (\vb^* - \vb) + \frac{1}{2} (\vb^* - \vb)^\top \nabla^2 Q(\bar{\vb}) (\vb^* - \vb) \, ,
    \label{eq:goyal revenue taylor}
\end{align}
where $\bar{\vb}$ is a convex combination of $\vb^*$ and $\vb$. 
For the first term in Eq.~\eqref{eq:goyal revenue taylor}, we have
\begin{align}
    &\nabla Q(\vb^*)^\top (\vb^* - \vb) \nonumber
    \\
    & = \frac{\sum_{i \in S_t} \exp(\xb_{t,j}^\top \thetab^*) (v_j - v_j^*)}{ 1 + \sum_{j \in S_t} \exp(\xb_{t,j}^\top \thetab^*)}
        - \frac{\sum_{i \in S_t} \exp(\xb_{t,j}^\top \thetab^*) \sum_{i \in S_t} \exp(\xb_{t,i}^\top \thetab^*)(v_j - v_j^*)}{ \left( 1 + \sum_{j \in S_t} \exp(\xb_{t,j}^\top \thetab^*) \right)^2}
    \nonumber
    \\
    & = \sum_{j \in S_t} q_{t,j}(S_t, \thetab^*) \xb_{t,j}^\top (\thetab^* - \thetab)
        - \sum_{j \in S_t} \sum_{i \in S_t} q_{t,j}(S_t, \thetab^*) q_{t,j}(S_t, \thetab^*) \xb_{t,i}^\top (\thetab^* - \thetab)
    \nonumber
    \\
    & = \sum_{j \in S_t} q_{t,j}(S_t, \thetab^*) \left( 1 - \sum_{i \in S_t} q_{t,i}(S_t, \thetab^*) \right) \xb_{t,j}^\top (\thetab^* - \thetab)
    \nonumber
    \\
    & = \sum_{j \in S_t} q_{t,j}(S_t, \thetab^*) q_{t,0}(S_t, \thetab^*) \xb_{t,j}^\top (\thetab^* - \thetab)
    \nonumber
    \\
    & \le \sum_{j \in S_t} q_{t,j}(S_t, \thetab^*) q_{t,0}(S_t, \thetab^*) \| \xb_{t,j} \|_{\Hb^{-1}_t(\thetab^*)} \| \thetab^* - \thetab \|_{\Hb_t(\thetab^*)} \, ,
    \label{eq:goyal frist order term}
\end{align}
where $\Hb_t(\thetab)$ is the Gram matrix used in~\cite{perivier2022dynamic} defined by
\begin{equation*}
    \Hb_t (\thetab^*) :=  \sum_{\tau=1}^{t-1} \sum_{j \in S_\tau} q_{\tau,j}(S_\tau, \thetab^*) \xb_{\tau,j} \xb_{\tau,j}^\top
        - \sum_{j \in S_\tau} \sum_{i \in S_\tau} q_{\tau,j}(S_\tau, \thetab^*) q_{\tau, i}(S_\tau, \thetab^*) \xb_{\tau,j} \xb_{\tau,i}^\top \, .    
\end{equation*}
Note that the term $\| \thetab^* - \thetab \|_{\Hb_t(\thetab^*)}$ can be bounded by the concentration result of the estimated parameter. 
On the other hand, to apply the elliptical potential lemma to the term $ \sum_{j \in S_t} q_{t,j}(S_t, \thetab^*) q_{t,0}(S_t, \thetab^*) \| \xb_{t,j} \|_{\Hb^{-1}_t(\thetab^*)}$, note that 
$\Hb_t(\thetab^*)$ can be bounded as follows:
\begin{align}
    & \Hb_t (\thetab^*) \nonumber
    \\
    & = \Hb_{t-1} (\thetab^*) 
        + \sum_{j \in S_t} q_{t,j}(S_t, \thetab^*) \xb_{t,j} \xb_{t,j}^\top
        - \frac{1}{2}  \sum_{i,j \in S_t} q_{t,j}(S_t, \thetab^*) q_{t, i}(S_t, \thetab^*) \left( \xb_{t,j} \xb_{t,i}^\top + \xb_{t,i} \xb_{t,j}^\top \right)        
    \nonumber
    \\
    & \succeq
        \Hb_{t-1} (\thetab^*) 
        + \sum_{j \in S_t} q_{t,j}(S_t, \thetab^*) \xb_{t,j} \xb_{t,j}^\top
        - \frac{1}{2} \sum_{i, j \in S_t} q_{t,j}(S_t, \thetab^*) q_{t, i}(S_t, \thetab^*) \left( \xb_{t,j} \xb_{t,j}^\top + \xb_{t,i} \xb_{t,i}^\top \right)
    \nonumber
    \\
    & = \Hb_{t-1} (\thetab^*) 
        + \sum_{j \in S_t} q_{t,j}(S_t, \thetab^*) \left( 1 - \sum_{i \in S_t} q_{t,i}(S_t, \thetab^*) \right) \xb_{t,j} \xb_{t,j}^\top
    \nonumber
    \\
    & = \Hb_{t-1} (\thetab^*) 
        + \sum_{j \in S_t} q_{t,j}(S_t, \thetab^*) q_{t,0}(S_t, \thetab^*) \xb_{t,j} \xb_{t,j}^\top \, .
    \label{eq:goyal lower bound of hessian}
\end{align}
Now since the coefficient $ q_{t,j}(S_t, \thetab^*) q_{t,0}(S_t, \thetab^*)$ of $\| \xb \|_{\Hb_t^{-1}(\thetab^*)}$ in Eq.~\eqref{eq:goyal frist order term} aligns with the coefficients of the lower bound of $\Hb_t(\thetab^*)$ in Eq.~\eqref{eq:goyal lower bound of hessian}, the elliptical potential lemma can be applied. 
Note that such a lower bound in Eq.~\eqref{eq:goyal lower bound of hessian} holds since \citet{perivier2022dynamic} deals with the uniform reward, i.e., $1 - \sum_{i \in S_t} q_{t,i}(S_t, \thetab^*) = q_{t,0}(S_t, \thetab^*)$.

\paragraph{Mulitinomial logistic bandit problem~\cite{zhang2023online}}
~\citet{zhang2023online} address the multiple-parameter MNL contextual bandit problem where at each time step $t$ the agent selects an action $\xb_t \in \RR^d$ and receives response feedback $y_t \in \{ 0 \} \cup [N]$ with $N+1$ possible outcomes. 
Each outcome $i \in [N]$ is associated with a ground-truth parameter $\thetab_i^* \in \RR^d$, and the probability of the outcome $\PP(y_t = i \mid \xb_t)$ follows the MNL model,
\begin{equation*}
    \PP(y_t = i \mid \xb_t) = \frac{\exp(\xb_t^\top \thetab^*_i)}{1 + \sum_{j=1}^N \exp(\xb_t^\top \thetab^*_j)} \, ,
    \quad
    \PP(y_t = 0 \mid \xb_t) = 1 - \sum_{j=1}^N \PP(y_t = j \mid \xb_t)  \, .    
\end{equation*}
In this model, there are $N$ unknown choice parameter $\Thetab^* := [\thetab^*_1, \ldots, \thetab^*_N] \in \RR^{d \times N}$ and the agent chooses one context feature $\xb_t$, that is why we call multiple-parameter MNL model. 
Then, the expected revenue of an action $\xb_t$ in~\cite{zhang2023online} is given by
\begin{equation*}
    \sum_{i = 1}^N  \frac{\exp(\xb_t^\top \thetab_i^*) \rho_i }{1 + \sum_{j=1}^N \exp(\xb_t^\top \thetab_j^*)} 
    := \rhob^\top \sigmab(\Thetab^* \xb_t) \, ,
\end{equation*}
where we define the softmax function $\sigmab: \RR^N \rightarrow [0,1]^{N}$ by
\begin{equation*}
    [\sigmab(\zb)]_k = \frac{\exp( [\zb]_i) }{1 + \sum_{j=1}^N \exp([\zb]_j)} \, \quad \forall k \in [N] 
    \, \quad \text{and} \quad
    [\sigmab(\zb)]_0 = \frac{1}{1 + \sum_{j=1}^N \exp([\zb]_j)} \, \quad \forall k \in [N] \, ,
\end{equation*}
and $\rhob := [\rho_1, \ldots, \rho_N] \in \RR_+^{N+1}$ represents the reward for each outcome $j \in [N]$ with $\rho_0 = 0$.
Then, the difference between the revenue induced by $\Thetab^*$ and that by an estimator $\hat{\Thetab}$ in~\cite{zhang2023online} is expressed by
\begin{align}
    & \rhob^\top \left( \sigmab(\Thetab^* \xb_t) - \sigmab(\hat{\Thetab} \xb_t) \right)
    \nonumber
    \\
    & = \sum_{k=1}^N \rho_k \left( [\sigmab(\Thetab^* \xb_t)]_k - [\sigmab(\hat{\Thetab} \xb_t)]_k \right)
    \nonumber
    \\
    & = \sum_{k=1}^N \rho_k \left( \nabla [\sigmab(\hat{\Thetab} \xb_t)]_k \right)^\top (\Thetab^* - \hat{\Thetab}) \xb_t
        + \sum_{k=1}^N \rho_k \| (\Thetab^* - \Thetab) \xb_t \|_{\Xib_k} \, ,
    \label{eq:zhang first order term}
\end{align}
where $\Xib_k = \int_0^1 (1 - \nu) \nabla^2 [\sigmab ( \hat{\Thetab} \xb_t + \nu (\Thetab^* - \hat{\Thetab})\xb_t)]_k d \nu$.
Then for the first term in Eq.~\eqref{eq:zhang first order term}, we have
\begin{align}
    & \sum_{k=1}^N \rho_k \left( \nabla [\sigmab(\hat{\Thetab} \xb_t)]_k \right)^\top (\Thetab^* - \hat{\Thetab}) \xb_t
    \nonumber
    \\
    & \le \left| \rhob^\top \nabla \sigmab( \hat{\Thetab} \xb_t) (\Thetab^* - \hat{\Thetab}) \xb_t \right|
    \nonumber
    \\
    & = \left| \rhob^\top \nabla \sigmab( \hat{\Thetab} \xb_t) (\Ib_{N} \otimes \xb_t^\top) (\textvec(\Thetab^*) - \textvec(\hat{\Thetab})) \right|
    \nonumber
    \\
    & \le \| \textvec(\Thetab^*) - \textvec(\hat{\Thetab}) \|_{\Hb_t}
        \| \Hb_t^{-\frac{1}{2}} (\Ib_{N} \otimes \xb_t^\top) \nabla \sigmab( \hat{\Thetab} \xb_t) \rhob \|_2 \, 
    \label{eq:zhang bound term}
\end{align}
where $\Hb_t$ is the Gram matrix used in~\cite{zhang2023online} defined by
\begin{equation*}
    \Hb_t := \lambda \Ib_{N} + \sum_{s=1}^{t-1} \nabla \sigmab (\hat{\Thetab}_{s+1} \xb_s) \otimes \xb_s \xb_s^\top \, .
\end{equation*}
Note that the term $\| \textvec(\Thetab^*) - \textvec(\hat{\Thetab}) \|_{\Hb_t}$ in Eq.~\eqref{eq:zhang bound term} can be bounded by the concentration result of the estimated parameter, and the term $\| \Hb_t^{-\frac{1}{2}} (\Ib_{N} \otimes \xb_t^\top) \nabla \sigmab( \hat{\Thetab} \xb_t) \rhob \|_2$ also can be bounded as follows:
\begin{equation*}
    \| \Hb_t^{-\frac{1}{2}} (\Ib_{N} \otimes \xb_t^\top) \nabla \sigmab( \hat{\Thetab} \xb_t) \rhob \|_2
    \le \| \rhob \|_2 \| \Hb_t^{-\frac{1}{2}} (\Ib_{N} \otimes \xb_t^\top) \nabla \sigmab( \hat{\Thetab} \xb_t) \|_2 \, .
\end{equation*}
Here~\citet{zhang2023online} bound the term $\| \Hb_t^{-\frac{1}{2}} (\Ib_{N} \otimes \xb_t^\top) \nabla \sigmab( \hat{\Thetab} \xb_t) \|_2$ using a matrix version of elliptical lemma.
However, they assume $\| \rhob \|_2 \le R$ (Assumption 2 in~\cite{zhang2023online}).

Now, regarding the prediction error in our setting, the estimated values ($\tilde{V}^k_{h+1}(\cdot)$) for each reachable state are typically distinct, and we do not assume a constant upper bound on the $\ell_2$-norm of the estimated value vector for all reachable states.
Instead, we can bound the $\ell_2$-norm of the estimated value vector for all reachable states as follows:
\begin{equation*}
    \| \tilde{\Vb}^k_{h+1} (s,a) \|_2 
    \le \max_{s' \in \Scal_{s,a}} \left| \tilde{V}^k_{h+1}(s') \right| \sqrt{| \Scal_{s,a} |}
    \le H \sqrt{\Ucal} \, ,
\end{equation*}
where $\tilde{\Vb}^k_{h+1} (s,a) := \left[ \tilde{V}^k_{h+1}(s') \right]_{s' \in \Scal_{s,a}} \in \RR^{|\Scal_{s,a}|}$.
However, such a bound leads to a looser regret by a factor of $\sqrt{\Ucal}$.
To address, we adapt the \textit{feature centralization technique}~\cite{lee2024nearly} to bound the prediction error independently of $\Ucal$, without making any additional assumptions.
The key point is that the Hessian of per-round loss $\ell_{k,h}(\thetab)$ is expressed in terms of the centralized feature as follows:
\begin{equation*}
    \nabla^2 \ell_{k,h} (\thetab) = \sum_{s' \in \Scal_{k,h}} P_{\thetab}(s' \mid s^k_h,a^k_h) \bar{\varphib}(s^k_h,a^k_h,s'; \thetab) \bar{\varphib}(s^k_h,a^k_h,s';\thetab)^\top \, .
\end{equation*}
where $\bar{\varphib}(s,a,s'; \thetab) := \varphib(s,a,s') - \EE_{\tilde{s} \sim P_{\thetab}(\cdot \mid s,a)}[\varphib(s, a, \tilde{s})]$ is the centralized feature by $\thetab$.
Now, we provide the bound on prediction error of the estimated parameter updated by $\texttt{ORRL-MNL}$.

\begin{lemma}[Bound on the prediction error] \label{lemma:prediction error bound (alg2)}
    For any $\delta \in (0,1)$, suppose that Lemma~\ref{lemma:tight concentration} holds.
    Let us denote the prediction error about $\omdtheta{k}{h}$ by 
    \begin{equation*}
        \PE^k_h(s,a) := \sum_{s' \in \Scal_{s,a}} \left( P_{\omdtheta{k}{h}}(s' \mid s,a) - P_{\thetab^*_h}(s' \mid s,a)
        \right) \tilde{V}^k_{h+1}(s') \, .
    \end{equation*}
    Then, for any $(s,a) \in \Scal \times \Acal$, we have
    \begin{align*}
        |\Delta^k_h (s, a)| 
        &\le H \beta_k (\delta) \sum_{s' \in \Scal_{s,a}} P_{\omdtheta{k}{h}} (s' \mid s, a) \left\| \bar{\varphib}_{s,a,s'} (\omdtheta{k}{h}) \right\|_{\Bb^{-1}_{k,h}} 
        + 3H \beta_k (\delta)^2 \max_{s' \in \Scal_{s,a}} \| \varphib_{s,a,s'} \|^2_{\Bb^{-1}_{k,h}} \, .
        \label{eq:first bound on prediction error}  
    \end{align*} 
\end{lemma}

\begin{proof}[Proof of Lemma~\ref{lemma:prediction error bound (alg2)}]
    Let us define $F(\thetab) := \sum_{s' \in \Scal_{s,a}} P_{\thetab} (s' \mid s,a) \tilde{V}^k_{h+1} (s')$.
    Then, by Taylor expansion we have
    \begin{equation*}
        F(\thetab^*_h)
        = F(\omdtheta{k}{h})  + \nabla F(\omdtheta{k}{h})^\top (\thetab^*_h - \omdtheta{k}{h})
        + \frac{1}{2} ( \thetab^*_h - \omdtheta{k}{h} )^\top \nabla^2 F(\bar{\thetab}) (\thetab^*_h - \omdtheta{k}{h}) \, ,
    \end{equation*}
    where $\bar{\thetab} = (1-v) \thetab^*_h + v \omdtheta{k}{h} \,$ for some $v \in (0,1)$.
    By Proposition~\ref{prop:derivative of transition prob}, we have
    \begin{align*}
        \nabla F( \thetab ) 
        & = \sum_{s' \in \Scal_{s,a}} \nabla P_{\thetab} (s' \mid s,a) \tilde{V}^k_{h+1} (s')
        \\
        & = \sum_{s' \in \Scal_{s,a}} P_{\thetab} (s' \mid s, a) \left( \varphib_{s, a, s'} - \sum_{\tilde{s} \in \Scal_{s,a}} P_{\thetab} (\tilde{s} \mid s, a) \varphib_{s, a, \tilde{s}} \right) \tilde{V}^k_{h+1} (s')
        \\
        & = \sum_{s' \in \Scal_{s,a}} P_{\thetab} (s' \mid s, a) \bar{\varphib}_{s,a,s'} (\thetab) \tilde{V}^k_{h+1} (s') \, , 
    \end{align*}
    and
    \begin{align*}
        &\nabla^2 F(\thetab) = \sum_{s' \in \Scal_{s,a}} \nabla^2 P_{\thetab} (s' \mid s, a) \tilde{V}^k_{h+1} (s') \\
        &\, = \sum_{s' \in \Scal_{s,a}} P_{\thetab} (s' \mid s, a) \tilde{V}^k_{h+1} (s') \varphib_{s,a,s'} \varphib_{s,a,s'}^\top \\
        &\phantom{{}={}} - \sum_{s' \in \Scal_{s,a}} P_{\thetab} (s' \mid s, a) \tilde{V}^k_{h+1} (s') 
        \\
        &\phantom{{}={}={}} \cdot \sum_{s'' \in \Scal_{s,a}} P_{\thetab} (s'' \mid s, a) \left( \varphib_{s,a,s'} \varphib_{s,a,s''}^\top + \varphib_{s,a,s''} \varphib_{s,a,s'}^\top + \varphib_{s,a,s''} \varphib_{s,a,s''}^\top \right) \\
        &\phantom{{}={}} + 2 \sum_{s' \in \Scal_{s,a}} P_{\thetab} (s' \mid s, a) \tilde{V}^k_{h+1} (s') 
        \\
        &\phantom{{}={}={}} \cdot \left( \sum_{s'' \in \Scal_{s,a}}  P_{\thetab}(s'' \mid s, a) \varphib_{s,a,s''}  \right) \left(\sum_{s'' \in \Scal_{s,a}} P_{\thetab}(s'' \mid s, a)  \varphib_{s,a,s''} \right)^\top \, . 
    \end{align*}    

    Then, the prediction error can be bounded as follows:
    \begin{align*} 
        |\Delta^k_h (s, a)| 
        & = | F(\thetab^*_h) - F(\omdtheta{k}{h})| 
        \\
        & \le \left| \nabla F(\omdtheta{k}{h})^\top (\omdtheta{k}{h} - \thetab^*_h) \right|
        + \frac{1}{2} \left| (\omdtheta{k}{h} - \thetab^*_h)^\top \nabla^2 F(\bar{\thetab}) (\omdtheta{k}{h} - \thetab^*_h) \right| \numberthis \label{eq:prediction error upper bound} \, .
    \end{align*}

    For the first term in Eq.~\eqref{eq:prediction error upper bound}, 
    \begin{align*}
        \left| \nabla F(\omdtheta{k}{h})^\top (\omdtheta{k}{h} - \thetab^*_h) \right|
        & = \left| \sum_{s' \in \Scal_{s,a}} P_{\omdtheta{k}{h}} (s' \mid s, a) \bar{\varphib}_{s,a,s'} (\omdtheta{k}{h})^\top (\omdtheta{k}{h} - \thetab^*_h) \tilde{V}^k_{h+1} (s') \right|
        \\
        & \le H \sum_{s' \in \Scal_{s,a}} P_{\omdtheta{k}{h}} (s' \mid s, a) \left\| \bar{\varphib}_{s,a,s'} (\omdtheta{k}{h}) \right\|_{\Bb^{-1}_{k,h}} \left\| \omdtheta{k}{h} - \thetab^*_h \right\|_{\Bb_{k,h}}
        \\
        &\le H \beta_k (\delta) \sum_{s' \in \Scal_{s,a}} P_{\omdtheta{k}{h}} (s' \mid s, a) \left\| \bar{\varphib}_{s,a,s'} (\omdtheta{k}{h}) \right\|_{\Bb^{-1}_{k,h}} \numberthis \, ,
        \label{eq:prediction error eq 1}
    \end{align*}
    where in the first inequality we use $\tilde{V}^k_{h+1}(s') \le H$ and Cauchy-Scharwz inequality, and the second inequality follows by the concentration result of Lemma~\ref{lemma:tight concentration}.

    For the second term in Eq.~\eqref{eq:prediction error upper bound}, since $0 \le \tilde{V}^k_{h+1}(s') \le H$, 
    \begin{align*}
        & \left| (\omdtheta{k}{h} - \thetab^*_h)^\top \nabla^2 F(\bar{\thetab}) (\omdtheta{k}{h} - \thetab^*_h) \right| 
        \\
        & \le H \sum_{s' \in \Scal_{s,a}} P_{\bar{\thetab}} (s' \mid s, a) \left( ( \omdtheta{k}{h} - \thetab^*_h )^\top \varphib_{s,a,s'} \right)^2
            \\
            & \phantom{-}
            + H \sum_{s' \in \Scal_{s,a}} P_{\bar{\thetab}} (s' \mid s, a)
            \\
            & \phantom{--}
            \cdot \sum_{s'' \in \Scal_{s,a}} P_{\bar{\thetab}} (s'' \mid s, a)
            \left| ( \omdtheta{k}{h} - \thetab^*_h )^\top \left( \varphib_{s,a,s'} \varphib_{s,a,s''}^\top + \varphib_{s,a,s''} \varphib_{s,a,s'}^\top \right) ( \omdtheta{k}{h} - \thetab^*_h ) \right|
            \\
            & \phantom{-}
            + H \sum_{s' \in \Scal_{s,a}} P_{\bar{\thetab}} (s' \mid s, a)
            \sum_{s'' \in \Scal_{s,a}} P_{\bar{\thetab}} (s'' \mid s, a) 
            \left(( \omdtheta{k}{h} - \thetab^*_h)^\top \varphib_{s,a,s''} \right)^2
            \\ 
            & \phantom{-}
            + 2 H \left( ( \omdtheta{k}{h} - \thetab^*_h )^\top  \bigg( \sum_{s'' \in \Scal_{s,a}}  P_{\bar{\thetab}}(s'' \mid s, a) \varphib_{s,a,s''}  \bigg) \right)^2
            \\
        & \le H \sum_{s' \in \Scal_{s,a}} P_{\bar{\thetab}} (s' \mid s, a) \| \varphib_{s,a,s'} \|^2_{\Bb^{-1}_{k,h}} \left\| \omdtheta{k}{h} - \thetab^*_h \right\|^2_{\Bb_{k,h}}
            \\
            & \phantom{-}
            + H \sum_{s' \in \Scal_{s,a}} P_{\bar{\thetab}} (s' \mid s, a)
            \\
            & \phantom{--} 
            \cdot \sum_{s'' \in \Scal_{s,a}} P_{\bar{\thetab}} (s'' \mid s, a)
            \left|
                ( \omdtheta{k}{h} - \thetab^*_h )^\top \left( \varphib_{s,a,s'} \varphib_{s,a,s'}^\top + \varphib_{s,a,s''} \varphib_{s,a,s''}^\top \right) ( \omdtheta{k}{h} - \thetab^*_h )
            \right|
            \\
            & \phantom{-}
            + H \sum_{s'' \in \Scal_{s,a}} P_{\bar{\thetab}} (s'' \mid s, a) \| \varphib_{s,a,s''} \|^2_{\Bb^{-1}_{k,h}} \left\| \omdtheta{k}{h} - \thetab^*_h \right\|^2_{\Bb_{k,h}}
            \\ 
            &\phantom{-}
            + 2 H \bigg( \sum_{s'' \in \Scal_{s,a}}  P_{\bar{\thetab}}(s'' \mid s, a) \| \varphib_{s,a,s''} \|_{\Bb^{-1}_{k,h}} \left\| \omdtheta{k}{h} - \thetab^*_h \right\|_{\Bb_{k,h}} \bigg)^2 \, , \numberthis \label{eq:prediction error eq 2-1}
    \end{align*}    
    where for the second inequality we use Cauchy-Schwarz inequality, $\xb \xb^\top + \yb \yb^\top \succeq \xb \yb^\top + \yb \xb^\top$ for any $\xb, \yb \in \RR^d$, and triangle inequality.
    Note that 
    \begin{align*}
        & H \sum_{s' \in \Scal_{s,a}} P_{\bar{\thetab}} (s' \mid s, a)
        \\
        & \phantom{-} \cdot \sum_{s'' \in \Scal_{s,a}} P_{\bar{\thetab}} (s'' \mid s, a)
        \left|
            ( \omdtheta{k}{h} - \thetab^*_h )^\top \left( \varphib_{s,a,s'} \varphib_{s,a,s'}^\top + \varphib_{s,a,s''} \varphib_{s,a,s''}^\top \right) ( \omdtheta{k}{h} - \thetab^*_h )
        \right|
        \\
        & = H \sum_{s' \in \Scal_{s,a}} P_{\bar{\thetab}} (s' \mid s, a)  \left( (\omdtheta{k}{h} - \thetab^*_h) ^\top  \varphib_{s,a,s'}\right)^2
        \\
        & \phantom{-} + H \sum_{s'' \in \Scal_{s,a}} P_{\bar{\thetab}} (s'' \mid s, a) \left( (\omdtheta{k}{h} - \thetab^*_h )^\top \varphib_{s,a,s''}\right)^2
        \\
        & \le 2 H \sum_{s' \in \Scal_{s,a}} P_{\bar{\thetab}} (s' \mid s, a) \| \varphib_{s,a,s'} \|^2_{\Bb^{-1}_{k,h}} \left\| \omdtheta{k}{h} - \thetab^*_h \right\|^2_{\Bb_{k,h}} \, . \numberthis \label{eq:prediction error eq 2-2}
    \end{align*}
    By substituting Eq.~\eqref{eq:prediction error eq 2-2} into Eq.~\eqref{eq:prediction error eq 2-1} we have
    \begin{align*}
        &\left| (\omdtheta{k}{h} - \thetab^*_h)^\top \nabla^2 F(\bar{\thetab}) (\omdtheta{k}{h} - \thetab^*_h) \right| 
        \\
        & \le  4 H \sum_{s' \in \Scal_{s,a}} P_{\bar{\thetab}} (s' \mid s, a) \| \varphib_{s,a,s'} \|^2_{\Bb^{-1}_{k,h}} \left\| \omdtheta{k}{h} - \thetab^*_h \right\|^2_{\Bb_{k,h}} 
            \\
            & \phantom{{}={}}
            +  2 H \bigg( \sum_{s'' \in \Scal_{s,a}}  P_{\bar{\thetab}}(s'' \mid s, a) \| \varphib_{s,a,s''} \|_{\Bb^{-1}_{k,h}} \left\| \omdtheta{k}{h} - \thetab^*_h \right\|_{\Bb_{k,h}} \bigg)^2 
        \\
        & \le 4 H \beta_k^2 \max_{s' \in \Scal_{s,a}} \| \varphib_{s,a,s'} \|^2_{\Bb^{-1}_{k,h}} 
            + 2 H \left( \beta_k \max_{s' \in \Scal_{s,a}} \| \varphib_{s,a,s'} \|_{\Bb^{-1}_{k,h}} \right)^2
        \\
        & \le 6 H \beta_k^2 \max_{s' \in \Scal_{s,a}} \| \varphib_{s,a,s'} \|^2_{\Bb^{-1}_{k,h}} \, ,
        \numberthis \label{eq:prediction error eq 2-3}
    \end{align*}
    where for the second inequality follows by Lemma~\ref{lemma:tight concentration} and $\sum_{s' \in \Scal_{s,a}} P_{\bar{\thetab}}(s' \mid s,a) =1$.
    Combining the results of Eq.~\eqref{eq:prediction error eq 1} and Eq.~\eqref{eq:prediction error eq 2-3} and , we conclude the proof.
\end{proof}

\subsection{Good Events with High Probability}
In this section, we introduce the good events used to prove Theorem~\ref{thm:alg 2} and show that the good events happen with high probability. 
\begin{lemma}[Good event probability] \label{lemma:good event prob_2}
    For any $K \in \NN $ and $\delta \in (0,1)$, the good event $\Gfrak(K,\delta')$ holds with probability at least $1 - \delta$ where $\delta' = \delta /(2KH)$.
\end{lemma}

\begin{proof}[Proof of Lemma~\ref{lemma:good event prob_2}]
    For any $\delta' \in (0,1)$, we have
    \begin{equation*}
        \Gfrak(K, \delta')
        = \bigcap_{k \le K} \bigcap_{h \le H} \Gfrak_{k,h} (\delta')
        = \bigcap_{k \le K} \bigcap_{h \le H} \left\{ \Gfrak_{k,h}^\Delta (\delta') \cap \Gfrak_{k,h}^{\xib} (\delta')  \right\} \, .
    \end{equation*}
    On the other hand, for any $(k,h) \in [K] \times [H]$, by Lemma~\ref{auxiliary lemma: gaussian noise concentration} $\Gfrak_{k,h}^{\xib} (\delta') $ holds with probability at least $1 - \delta'$.
    Then, for $\delta' = \delta / (2KH) $ by taking union bound, we have the desired result as follows:
        \begin{equation*}
            \PP (\Gfrak(K, \delta')) \ge (1 - \delta')^{2KH} \ge 1 - 2KH \delta' = 1 - \delta \, .
        \end{equation*}
\end{proof}

\subsection{Stochastic Optimism}

\begin{lemma}[Stochastic optimism] \label{lemma:stochastic optimism of alg2}
    For any $\delta$ with $0 < \delta < \Phi(-1)/2$, let $\sigma_k = H \beta_k (\delta)$. 
    If we take multiple sample size $M = \lceil 1 - \frac{\log (H \Ucal)}{\log \Phi(1)} \rceil$, then for any $k \in [K]$, we have
    \begin{equation*}
        \PP \left( (\tilde{V}^k_1 - V^*_1)(s^k_1) \ge 0 \mid s^k_1, \Fcal_k \right) \ge \Phi(-1)/2 \, .
    \end{equation*}
\end{lemma}

\begin{proof}[Proof of Lemma~\ref{lemma:stochastic optimism of alg2}]
    First, we introduce the following lemmas.
    \begin{lemma} \label{lemma:stochastic optimism for given h and k (alg2)}
        Let $\delta \in (0,1)$ be given.
        For any $(k, h) \in [K] \times [H]$, let $\sigma_k = H \beta_k(\delta)$.
        If we define the event $\Gfrak_{k,h}^{\Delta} (\delta)$ as
        \begin{equation*}
            \begin{split}
                \Gfrak_{k,h}^\Delta (\delta) 
                & := \bigg\{
                    | \Delta^k_h(s,a) | \le H \beta_k (\delta) \sum_{s' \in \Scal_{s,a}} P_{\omdtheta{k}{h}} (s' \mid s, a) \left\| \bar{\varphib}_{s,a,s'} (\omdtheta{k}{h}) \right\|_{\Bb^{-1}_{k,h}}
                    \\
                    & \phantom{{}={}}
                    + 3H \beta_k (\delta)^2 \max_{s' \in \Scal_{s,a}} \| \varphib_{s,a,s'} \|^2_{\Bb^{-1}_{k,h}}
                    \bigg\} \, ,       
            \end{split}  
        \end{equation*}
        then conditioned on $\Gfrak_{k,h}^{\Delta} (\delta)$, for any $(s,a) \in \Scal \times \Acal$, we have
        \begin{equation*}
            \PP \left( - \BE^k_h(s,a) \ge 0 \mid \Gfrak^{\Delta}_{k,h}(\delta) \right) \ge 1 - \Phi(1)^M \, .
        \end{equation*}
    \end{lemma}

    \begin{lemma} \label{lemma:stochastic optimism for all h (alg2)}
        Let $\delta \in (0,1)$ be given. 
        For any $(h, k) \in [H] \times [K]$, let $\sigma_k = H \beta_k(\delta)$. 
        If we take multiple sample size $M = \lceil 1 - \frac{\log (H \Ucal)}{\log \Phi(1)} \rceil$, then conditioned on the event $\Gfrak^{\Delta}_{k}(\delta) := \cap_{h \in [H]} \Gfrak^{\Delta}_{k,h}(\delta)$, we have
        \begin{equation*}
            \PP \left( - \BE^k_h(s_h, a_h) \ge 0, \forall h \in [H] \mid \Gfrak^{\Delta}_k (\delta) \right) \ge \Phi(-1) \, .
        \end{equation*}
    \end{lemma}
    Based on the result of Lemma~\ref{lemma:stochastic optimism for all h (alg2)}, using the same argument as in Lemma~\ref{lemma:stochastic optimism} we obtain the desired result. 
\end{proof}
In the following section, we provide the proofs of the lemmas used in Lemma~\ref{lemma:stochastic optimism of alg2}.

\subsubsection{Proof of Lemma~\ref{lemma:stochastic optimism for given h and k (alg2)}}
\begin{proof}[Proof of Lemma~\ref{lemma:stochastic optimism for given h and k (alg2)}]
    Recall the definition of Bellman error (Definition~\ref{def:prediction error & bellman error}), we have
    \begin{align*}
        & -\BE^k_h (s, a)
        \\
        & = \tilde{Q}^k_h (s,a) - \left( r(s,a) + P_h \tilde{V}^k_{h+1} (s, a) \right)
        \\
        & = \min \bigg\{ r(s, a) + \sum_{s' \in \Scal_{s,a}} P_{\omdtheta{k}{h}} (s' \mid s, a) \tilde{V}^k_{h+1} (s') + \rbonus_{k,h}(s,a) \bigg\}
            - \left( r(s,a) + P_h \tilde{V}^k_{h+1} (s, a) \right)
        \\
        & \ge \min \bigg\{ \sum_{s' \in \Scal_{s,a}} P_{\omdtheta{k}{h}} (s' \mid s, a) \tilde{V}^k_{h+1} (s') - P_h \tilde{V}^k_{h+1} (s, a) + \rbonus_{k,h}(s,a), 0 \bigg\} \, .
    \end{align*}
    Then, it is enough to show that
    \begin{equation*}
        \sum_{s' \in \Scal_{s,a}} P_{\omdtheta{k}{h}} (s' \mid s, a) \tilde{V}^k_{h+1} (s') - P_h \tilde{V}^k_{h+1} (s, a) + \rbonus_{k,h}(s,a) \ge 0
    \end{equation*}
    at least with constant probability. 
    On the other hand, under the event $\Gfrak_{k,h}^{\Delta} (\delta)$, by Lemma~\ref{lemma:prediction error bound (alg2)} we have
    \begin{align*}
        & \sum_{s' \in \Scal_{s,a}} P_{\omdtheta{k}{h}} (s' \mid s, a) \tilde{V}^k_{h+1} (s') - P_h \tilde{V}^k_{h+1} (s, a) + \rbonus_{k,h}(s,a)
        \\
        & = \PE^k_h(s,a) + \rbonus_{k,h}(s,a)
        \\
        & \ge - H \beta_k (\delta) \sum_{s' \in \Scal_{s,a}} P_{\omdtheta{k}{h}} (s' \mid s, a) \left\| \bar{\varphib}_{s,a,s'} (\omdtheta{k}{h}) \right\|_{\Bb^{-1}_{k,h}}
        \\
        & \phantom{--} - 3 H \beta_k(\delta)^2 \max_{s' \in \Scal_{s,a}} \| \varphib_{s,a,s'} \|_{\Bb_{k,h}^{-1}}^2
            + \rbonus_{k,h}(s,a)
        \\
        & = \sum_{s' \in \Scal_{s,a}} P_{\omdtheta{k}{h}} (s' \mid s, a) \bar{\varphib}_{s, a, s'} (\omdtheta{k}{h})^\top \xib^{s'}_{k,h} 
        - H \beta_k (\delta) \sum_{s' \in \Scal_{s,a}} P_{\omdtheta{k}{h}} (s' \mid s, a) \left\| \bar{\varphib}_{s,a,s'} (\omdtheta{k}{h}) \right\|_{\Bb^{-1}_{k,h}} \, .
    \end{align*}
    Note that since $\xib^{(m)}_{k,h} \sim \Ncal (\zero, \sigma_k^2 \Bb^{-1}_{k,h})$, it follows that
    \begin{equation*}
        \bar{\varphib}_{s, a, s'} (\omdtheta{k}{h})^\top \xib^{(m)}_{k,h} \sim \Ncal \left( 0, \sigma_k^2 \left\| \bar{\varphib}_{s,a,s'} (\omdtheta{k}{h}) \right\|^2_{\Bb_{k,h}^{-1}} \right) \, , \quad \forall m \in [M] \, .
    \end{equation*}
    Therefore, by setting $\sigma_k = H \beta_k(\delta)$, we have for $m \in [M]$ and $s' \in \Scal_{s,a}$,
    \begin{equation*}
         \PP \left( \bar{\varphib}_{s, a, s'} (\omdtheta{k}{h})^\top \xib^{(m)}_{k,h} \ge H \beta_k(\delta) \left\| \bar{\varphib}_{s,a,s'} (\omdtheta{k}{h}) \right\|_{\Bb_{k,h}^{-1}} \right)
         = \Phi(-1) \, .
    \end{equation*}
    Recall that $\xib^{s'}_{k,h} := \xib^{m(s')}_{k,h}$ where $m(s') := \argmax_{m \in [M]} \bar{\varphib}_{s,a,s'}(\omdtheta{k}{h})^\top \xib^{(m)}_{k,h}$. 
    Then, we can deduce
    \begin{align}
        & \PP \left( \bar{\varphib}_{s, a, s'} (\omdtheta{k}{h})^\top \xib^{s'}_{k,h} \ge H \beta_k (\delta) \left\| \bar{\varphib}_{s,a,s'} (\omdtheta{k}{h}) \right\|_{\Bb^{-1}_{k,h}} \right)
        \nonumber
        \\
        & = \PP \left( \max_{m \in [M]} \bar{\varphib}_{s, a, s'} (\omdtheta{k}{h})^\top \xib^{(m)}_{k,h} \ge H \beta_k (\delta) \left\| \bar{\varphib}_{s,a,s'} (\omdtheta{k}{h}) \right\|_{\Bb^{-1}_{k,h}} \right)
        \nonumber
        \\
        & = 1 - \PP \left( \bar{\varphib}_{s, a, s'} (\omdtheta{k}{h})^\top \xib^{(m)}_{k,h} < H \beta_k (\delta) \left\| \bar{\varphib}_{s,a,s'} (\omdtheta{k}{h}) \right\|_{\Bb^{-1}_{k,h}}, \forall m \in [M] \right)
        \nonumber
        \\
        & \ge 1 - (1 - \Phi (-1))^M
        \nonumber
        \\
        & = 1 - \Phi(1)^M \, .
        \label{eq:stochastic optimism for given h and k (alg2) 1}
    \end{align}
    Consequently, we arrive at the conclusion as follows:
    \begin{align}
        & \PP (-\iota^k_h (s, a) \ge 0 \mid \Gfrak_{k,h}^\Delta (\delta))
        \nonumber
        \\
        & \ge \PP \left( \sum_{s' \in \Scal_{s,a}} P_{\omdtheta{k}{h}} (s' \mid s, a) \tilde{V}^k_{h+1} (s') - P_h \tilde{V}^k_{h+1} (s, a) + \rbonus_{k,h}(s,a) \ge 0 \mid \Gfrak_{k,h}^\Delta (\delta) \right)
        \nonumber
        \\
        & \ge \PP \left( \sum_{s' \in \Scal_{s,a}} P_{\omdtheta{k}{h}} (s' \mid s, a) \bar{\varphib}_{s, a, s'} (\omdtheta{k}{h})^\top \xib^{s'}_{k,h} \right.
        \\
        & \phantom{----} \left.
        \ge H \beta_k (\delta) \sum_{s' \in \Scal_{s,a}} P_{\omdtheta{k}{h}} (s' \mid s, a) \left\| \bar{\varphib}_{s,a,s'} (\omdtheta{k}{h}) \right\|_{\Bb^{-1}_{k,h}} \mid \Gfrak_{k,h}^\Delta (\delta) \right)
        \nonumber
        \\
        & \ge \PP \left( \bar{\varphib}_{s, a, s'} (\omdtheta{k}{h})^\top \xib^{s'}_{k,h} \ge H \beta_k (\delta) \left\| \bar{\varphib}_{s,a,s'} (\omdtheta{k}{h}) \right\|_{\Bb^{-1}_{k,h}} , \, \forall s' \in \Scal_{s,a} \mid \Gfrak_{k,h}^\Delta (\delta) \right)
        \nonumber
        \\
        & = 1 - \PP \left( \exists s' \in \Scal_{s,a} \,\, \text{s.t.} \,\, \bar{\varphib}_{s, a, s'} (\omdtheta{k}{h})^\top \xib^{s'}_{k,h} < H \beta_k (\delta) \left\| \bar{\varphib}_{s,a,s'} (\omdtheta{k}{h}) \right\|_{\Bb^{-1}_{k,h}} \mid \Gfrak_{k,h}^\Delta (\delta) \right) 
        \nonumber
        \\
        & \ge 1 - \Ucal \PP \left( \bar{\varphib}_{s, a, s'} (\omdtheta{k}{h})^\top \xib^{s'}_{k,h} < H \beta_k (\delta) \left\| \bar{\varphib}_{s,a,s'} (\omdtheta{k}{h}) \right\|_{\Bb^{-1}_{k,h}} \mid \Gfrak_{k,h}^\Delta (\delta) \right) 
        \label{eq:stochastic optimism for given h and k (alg2) 2}
        \\
        &\ge 1- \Ucal \Phi (1)^M \, ,
        \label{eq:stochastic optimism for given h and k (alg2) 3}
    \end{align}
    where~\eqref{eq:stochastic optimism for given h and k (alg2) 2} comes from the fact that $\max_{s,a} |\Scal_{s,a}| = \Ucal$ and the union bound, and~\eqref{eq:stochastic optimism for given h and k (alg2) 3} follows by~\eqref{eq:stochastic optimism for given h and k (alg2) 1}.    
\end{proof}

\subsubsection{Proof of Lemma~\ref{lemma:stochastic optimism for all h (alg2)}}

\begin{proof}[Proof of Lemma~\ref{lemma:stochastic optimism for all h (alg2)}]
    It holds 
    \begin{align*}
        \PP \left( -\BE^k_h (s_h, a_h) \ge 0, \forall h \in [H] \right)
        & = 1 - \PP \left( \exists h \in [H] \,\, \text{s.t.} \,\, -\BE^k_h (s_h, a_h) < 0 \right)
        \\
        & \ge 1 - H \PP \left( -\BE^k_h (s_h, a_h) < 0 \right)
        \\
        & \ge 1 - H \Ucal \Phi(1)^M
        \\
        & \ge \Phi (-1)
    \end{align*}
    where the first inequality uses the Bernoulli's inequality, the second inequality follows by Lemma~\ref{lemma:stochastic optimism for given h and k (alg2)}, and the last inequality holds due to the choice of $M = \lceil 1- \frac{\log(\Ucal H )}{ \log \Phi(1) } \rceil $.
\end{proof}

\subsection{Bound on Estimation Part}
In this section, we provide the upper bound on the estimation part of the regret: $\sum_{k=1}^K (\tilde{V}^k_1 - V^*_1)(s^k_1)$.
\begin{lemma}[Bound on estimation] \label{lemma:estimation bound for alg2}
    For any $\delta \in (0, 1)$, if $\lambda = \Ocal(\Fnorm^2 d \log \Ucal)$,
    then with probability at least $1 - \delta/2$, we have
    \begin{align*}
        \sum^K_{k=1} (\tilde{V}^k_1 - V^{\pi^k}_1) (s^k_1) = \BigOTilde \left( d^{3/2} H^{3/2} \sqrt{T} + \kappa^{-1} d^2 H^2 \right) \, .
    \end{align*}
\end{lemma}

\begin{proof}[Proof of Lemma~\ref{lemma:estimation bound for alg2}]
    With the same argument in Lemma~\ref{lemma:bound of estimation part}, we have
    \begin{equation}
        (\tilde{V}^k_1 - V^{\pi^k}_1) (s^k_1) = \sum_{h=1}^H - \iota^k_h (s^k_h, a^k_h) + \sum_{h=1}^H \dot{\zeta}^k_h \, ,
    \end{equation}
    where $\dot{\zeta}^k_h:= P_h (\tilde{V}^k_{h+1} - V^{\pi^k}_{h+1})(s^k_h, a^k_h) - (\tilde{V}^k_{h+1} - V^{\pi^k}_{h+1}) (s^k_{h+1})$.
    Note that
    \begin{align*}
        - \BE^k_h (s^k_h, a^k_h)
        & = \tilde{Q}^k_h (s^k_h, a^k_h) - \left( r(s^k_h, a^k_h) + P_h \tilde{V}^k_{h+1}(s^k_h, a^k_h) \right) 
        \\
        & \le \sum_{s' \in \Scal_{k,h}} P_{\omdtheta{k}{h}}(s' \mid s^k_h, a^k_h) \tilde{V}^k_{h+1}(s')  - P_h \tilde{V}^k_{h+1}(s^k_h, a^k_h) 
            + \rbonus_{k,h}(s^k_h, a^k_h)
        \\
        & \le \left| \PE^k_h (s^k_h, a^k_h) \right| 
            + \rbonus_{k,h}(s^k_h, a^k_h)
        \\
        & \le H \beta_k \sum_{s' \in \Scal_{k,h}} P_{\omdtheta{k}{h}} (s' \mid s^k_h, a^k_h) \left\| \bar{\varphib}_{k,h,s'} (\omdtheta{k}{h}) \right\|_{\Bb^{-1}_{k,h}}
            + 3H \beta_k^2 \max_{s' \in \Scal_{k,h}} \| \varphib_{k,h,s'} \|^2_{\Bb^{-1}_{k,h}}
            \\
            & \phantom{{}={}}
            + \rbonus_{k,h}(s^k_h, a^k_h) \, ,
            \numberthis \label{eq:estimation bound for alg2 1}
    \end{align*}
    where the last inequality follows by Lemma~\ref{lemma:prediction error bound (alg2)}.
    Now we introduce the following lemma.
    \begin{lemma} \label{lemma:auxliary bound for prediction error}
        For any $(k,h) \in [K] \times [H]$ and $(s,a) \in \Scal \times \Acal$, it holds
        \begin{align*}
            & \sum_{s' \in \Scal_{s,a}} P_{\omdtheta{k}{h}} (s' \mid s, a) \left\| \bar{\varphib}_{s,a,s'} (\omdtheta{k}{h}) \right\|_{\Bb^{-1}_{k,h}} 
            \\
            & \le \sum_{s' \in \Scal_{s,a}} P_{\omdtheta{k+1}{h}} (s' \mid s, a) \left\| \bar{\varphib}_{s,a,s'} (\omdtheta{k+1}{h}) \right\|_{\Bb^{-1}_{k,h}}
            + \frac{16 \eta \Fnorm}{\sqrt{\lambda}} \max_{s' \in \Scal_{s,a}} \left\| \bar{\varphib}_{s,a,s'} (\omdtheta{k+1}{h}) \right\|^2_{\Bb^{-1}_{k,h}} \, .
        \end{align*}
    \end{lemma}
    By plugging the result of Lemma~\ref{lemma:auxliary bound for prediction error} into Eq.~\eqref{eq:estimation bound for alg2 1}, we have
    \begin{align*}
        &- \BE^k_h (s^k_h, a^k_h)
        \\
        & \le H \beta_k \sum_{s' \in \Scal_{k,h}} P_{\omdtheta{k+1}{h}} (s' \mid s^k_h, a^k_h) \left\| \bar{\varphib}_{k,h,s'} (\omdtheta{k+1}{h}) \right\|_{\Bb^{-1}_{k,h}}
            \\
            & \phantom{{}={}}
            + H \beta_k \frac{16 \eta \Fnorm}{\sqrt{\lambda}} \max_{s' \in \Scal_{k,h}} \left\| \bar{\varphib}_{k,h,s'} (\omdtheta{k+1}{h}) \right\|^2_{\Bb^{-1}_{k,h}}
            + 3H \beta_k^2 \max_{s' \in \Scal_{k,h}} \| \varphib_{k,h,s'} \|^2_{\Bb^{-1}_{k,h}}
            + \rbonus_{k,h}(s^k_h, a^k_h)
        \\
        & \le H \beta_k \sum_{s' \in \Scal_{k,h}} P_{\omdtheta{k+1}{h}} (s' \mid s^k_h, a^k_h) \left\| \bar{\varphib}_{k,h,s'} (\omdtheta{k+1}{h}) \right\|_{\Bb^{-1}_{k,h}}
            \\
            & \phantom{{}={}}
            + \sum_{s' \in \Scal_{k,h}} P_{\omdtheta{k}{h}} (s' \mid s^k_h, a^k_h) \bar{\varphib}_{k,h, s'}(\omdtheta{k}{h})^\top \xib^{s'}_{k,h}
            \\
            & \phantom{{}={}}
            + H \beta_k \frac{16 \eta \Fnorm}{\sqrt{\lambda}} \max_{s' \in \Scal_{k,h}} \left\| \bar{\varphib}_{k,h,s'} (\omdtheta{k+1}{h}) \right\|^2_{\Bb^{-1}_{k,h}}
            + 6 H \beta_k^2 \max_{s' \in \Scal_{k,h}} \| \varphib_{k,h,s'} \|^2_{\Bb^{-1}_{k,h}} \, .
    \end{align*}
    By letting us denote 
    \begin{equation}
        \Upsilon^k_h(s, a)
        := H \beta_k \frac{16 \eta \Fnorm}{\sqrt{\lambda}} \max_{s' \in \Scal_{s,a}} \left\| \bar{\varphib}_{s,a,s'} (\omdtheta{k+1}{h}) \right\|^2_{\Bb^{-1}_{k,h}}
        + 6 H \beta_k^2 \max_{s' \in \Scal_{s,a}} \| \varphib_{s,a,s'} \|^2_{\Bb^{-1}_{k,h}} \, ,
    \end{equation}
    and summing over all episodes, we have
    \begin{align*}
        \sum_{k=1}^K (\tilde{V}^k_1 - V^{\pi^k}_1) (s^k_1)
        & = \sum_{k=1}^K \sum_{h=1}^H - \iota^k_h (s^k_h, a^k_h) + \sum_{k=1}^K \sum_{h=1}^H \dot{\zeta}^k_h
        \\
        & \le \underbrace{H \beta_K \sum_{k=1}^K \sum_{h=1}^H \sum_{s' \in \Scal_{k,h}} P_{\omdtheta{k+1}{h}} (s' \mid s^k_h, a^k_h) \left\| \bar{\varphib}_{k,h,s'} (\omdtheta{k+1}{h}) \right\|_{\Bb^{-1}_{k,h}}}_{(\mathrm{i})}
            \\
            & \phantom{{}={}}
            + \underbrace{\sum_{k=1}^K \sum_{h=1}^H \sum_{s' \in \Scal_{k,h}} P_{\omdtheta{k}{h}} (s' \mid s^k_h, a^k_h) \bar{\varphib}_{k,h, s'}(\omdtheta{k}{h})^\top \xib^{s'}_{k,h}}_{(\mathrm{ii})}
            \\
            & \phantom{{}={}}
            + \underbrace{\sum_{k=1}^K \sum_{h=1}^H \Upsilon^k_h(s^k_h,a^k_h)}_{(\mathrm{iii})}
            + \underbrace{\sum_{k=1}^K \sum_{h=1}^H \dot{\zeta}^k_h}_{(\mathrm{iv})} \, . 
            \numberthis
    \end{align*}

    Note that $\sum_{k=1}^K \sum_{h=1}^H \sum_{s' \in \Scal_{k,h}}$ is hereafter abbreviated as $\sum_{k,h,s'}$.

    For term $(\mathrm{i})$, we have
    \begin{align*}
        & \sum_{k,h,s'} P_{\omdtheta{k+1}{h}} (s' \mid s^k_h, a^k_h) \left\| \bar{\varphib}_{k,h,s'} (\omdtheta{k+1}{h}) \right\|_{\Bb^{-1}_{k,h}}
        \\
        & \le \sqrt{ \sum_{k,h,s'} P_{\omdtheta{k+1}{h}} (s' \mid s^k_h, a^k_h)}
            \sqrt{\sum_{k,h,s'} P_{\omdtheta{k+1}{h}} (s' \mid s^k_h, a^k_h) \left\| \bar{\varphib}_{k,h,s'} (\omdtheta{k+1}{h}) \right\|^2_{\Bb^{-1}_{k,h}}}
            \nonumber
        \\
        & = \sqrt{T} \sqrt{\sum_{h=1}^H \sum_{k=1}^K  \sum_{s' \in \Scal_{k,h}} P_{\omdtheta{k+1}{h}} (s' \mid s^k_h, a^k_h) \left\| \bar{\varphib}_{k,h,s'} (\omdtheta{k+1}{h}) \right\|^2_{\Bb^{-1}_{k,h}}}
        \nonumber
        \\
        & \le \sqrt{T} \sqrt{2 H d \log \left( 1 + \frac{K \Ucal L^2_{\varphib}}{ d \lambda } \right)} \, ,
        \numberthis \label{eq:bound on estimation for alg 2 final 1}
    \end{align*}
    where the last inequality follows by the following lemma:
    \begin{lemma} \label{lemma:elliptical potential for centralized feature}
        For each $ h \in [H]$, if $\lambda \ge \Fnorm^2$, then we have
        \begin{align*}
            \sum^{K}_{k=1} \sum_{s' \in \Scal_{k,h}} P_{\omdtheta{k+1}{h}} (s' \mid s^k_h, a^k_h) \left\| \bar{\varphib}_{k,h,s'} (\omdtheta{k+1}{h}) \right\|^2_{\Bb^{-1}_{k,h}}
            \le 2d \log \left( 1 + \frac{K \Ucal \Fnorm^2}{ d \lambda } \right) \, .
        \end{align*}
    \end{lemma}
    Then, term $(\mathrm{i})$ can be bounded as follows:
    \begin{align*} 
        (\mathrm{i})
        & = H \beta_K \sum_{k=1}^K \sum_{h=1}^H \sum_{s' \in \Scal_{k,h}} P_{\omdtheta{k+1}{h}} (s' \mid s^k_h, a^k_h) \left\| \bar{\varphib}_{k,h,s'} (\omdtheta{k+1}{h}) \right\|_{\Bb^{-1}_{k,h}}
        \\
        & \le H \beta_K \sqrt{T} \sqrt{2 H d \log \left( 1 + \frac{K \Ucal L^2_{\varphib}}{ d \lambda } \right)}
        \\
        & = \tilde{\Ocal}( d H^{3/2} \sqrt{T})  \, .
        \numberthis 
        \label{eq:bound on estimation for alg 2 final 2}
    \end{align*}

    For term $(\mathrm{ii})$, we introduce the following lemma:
    \begin{lemma} \label{lemma:auxliary bound for estimation bound (alg2)}
        Let $\delta \in (0,1)$ be given. 
        For any $(k,h) \in [K] \times [H]$ and $(s,a) \in \Scal \times \Acal$, with probability at least $1-\delta$, it holds
        \begin{align*}
            & \sum_{s' \in \Scal_{s,a}} P_{\omdtheta{k}{h}} (s' \mid s, a) \bar{\varphib}_{s,a,s'} (\omdtheta{k}{h})^\top \xib^{s'}_{k,h} \\
            & \le \gamma_k(\delta)
                \bigg(
                \sum_{s' \in \Scal_{k,h}} P_{\omdtheta{k+1}{h}} (s' \mid s^k_h, a^k_h) \left\| \bar{\varphib}_{k,h,s'} (\omdtheta{k+1}{h}) \right\|_{\Bb^{-1}_{k,h}}
            \\    
            & \phantom{-----} + \frac{16 \eta \Fnorm}{\sqrt{\lambda}} \max_{s' \in \Scal_{k,h}} \left\| \bar{\varphib}_{k,h,s'} (\omdtheta{k+1}{h}) \right\|^2_{\Bb^{-1}_{k,h}}
                \bigg) \, , 
        \end{align*}
        where $\gamma_k(\delta) := C_{\xib} \sigma_k \sqrt{d \log(Md / \delta)}$ for an absolute constant $C_{\xib} > 0$.
    \end{lemma}
    
    By Lemma~\ref{lemma:auxliary bound for estimation bound (alg2)}, we have
    \begin{align*}
        & \sum_{k,h,s'} P_{\omdtheta{k}{h}} (s' \mid s^k_h, a^k_h) \bar{\varphib}_{k,h, s'}(\omdtheta{k}{h})^\top \xib^{s'}_{k,h}
        \\
        & \le \gamma_{K}(\delta) \bigg( \sum_{k,h,s'} P_{\omdtheta{k+1}{h}} (s' \mid s^k_h, a^k_h) \left\| \bar{\varphib}_{k,h,s'} (\omdtheta{k+1}{h}) \right\|_{\Bb^{-1}_{k,h}}
        \\    
        &\phantom{-----}
        + \frac{16 \eta \Fnorm}{\sqrt{\lambda}}  \sum_{k=1}^K \sum_{h=1}^H  \max_{s' \in \Scal_{k,h}} \left\| \bar{\varphib}_{k,h,s'} (\omdtheta{k+1}{h}) \right\|^2_{\Bb^{-1}_{k,h}} \bigg)
        \\
        & \le \gamma_{K}(\delta) \bigg( 
            \sqrt{T} \sqrt{2 H d \log \left( 1 + \frac{K \Ucal L^2_{\varphib}}{ d \lambda } \right)}
            + \frac{16 \eta \Fnorm}{\sqrt{\lambda}} \sum_{k=1}^K \sum_{h=1}^H \max_{s' \in \Scal_{k,h}} \left\| \bar{\varphib}_{k,h,s'} (\omdtheta{k+1}{h}) \right\|^2_{\Bb^{-1}_{k,h}} \bigg) \, ,
        \numberthis
        \label{eq:bound on estimation for alg 2 final 3}
    \end{align*}
    where the last inequality follows by Eq.~\eqref{eq:bound on estimation for alg 2 final 1}.
    Note that
    \begin{align*}
        & \sum_{k=1}^K \sum_{h=1}^H \max_{s' \in \Scal_{k,h}} \left\| \bar{\varphib}_{k,h,s'} (\omdtheta{k+1}{h}) \right\|^2_{\Bb^{-1}_{k,h}} 
        \\
        & \le \sum_{k=1}^K \sum_{h=1}^H \max_{s' \in \Scal_{k,h}} \left\| \bar{\varphib}_{k,h,s'} (\omdtheta{k+1}{h}) \right\|^2_{\Ab^{-1}_{k,h}} 
        \\
        & = \sum_{k=1}^K \sum_{h=1}^H \max_{s' \in \Scal_{k,h}} \left\| \varphib_{k,h,s'}  - \sum_{\tilde{s} \in \Scal_{k,h}} P_{\omdtheta{k+1}{h}}(\tilde{s} \mid s^k_h, a^k_h) \varphib_{k,h,\tilde{s}} \right\|^2_{\Ab^{-1}_{k,h}} 
        \\
        & \le \sum_{k=1}^K \sum_{h=1}^H \max_{s' \in \Scal_{k,h}}  \left( 
            2 \| \varphib_{k,h,s'} \|^2_{\Ab_{k,h}^{-1}} + 2 \left\| \sum_{\tilde{s} \in \Scal_{k,h}} P_{\omdtheta{k+1}{h}}(\tilde{s} \mid s^k_h, a^k_h) \varphib_{k,h,\tilde{s}} \right\|^2_{\Ab^{-1}_{k,h}}
        \right)
        \\
        & \le 2 \sum_{k=1}^K \sum_{h=1}^H \max_{s' \in \Scal_{k,h}} \| \varphib_{k,h,s'} \|^2_{\Ab_{k,h}^{-1}}
            + 2 \sum_{k=1}^K \sum_{h=1}^H  \sum_{\tilde{s} \in \Scal_{k,h}} P_{\omdtheta{k+1}{h}}(\tilde{s} \mid s^k_h, a^k_h) \|  \varphib_{k,h,\tilde{s}} \|^2_{\Ab^{-1}_{k,h}}
        \\ 
        & \le 4 \sum_{k=1}^K \sum_{h=1}^H \max_{s' \in \Scal_{k,h}} \| \varphib_{k,h,s'} \|^2_{\Ab_{k,h}^{-1}} 
        \\
        & \le 16 \kappa^{-1} d H \log \left( 1 + \frac{K \Ucal \Fnorm^2}{d \lambda} \right) \, ,
        \numberthis
        \label{eq:eq:bound on estimation for alg 2 final 4}
    \end{align*}
    where the first inequality holds since $\Bb_{k,h}^{-1} \preceq \Ab_{k,h}^{-1}$, the second inequality follows from $(x+y)^2 \le 2x^2 + 2y^2$, and the third inequality uses the triangle inequality, and the fourth inequality uses $\sum_{\tilde{s} \in \Scal_{k,h}} P_{\omdtheta{k+1}{h}}(\tilde{s} \mid s^k_h, a^k_h)=1$, and the last inequality follows by Lemma~\ref{lemma:generalized elliptical potential}.
    By substituting Eq.~\eqref{eq:eq:bound on estimation for alg 2 final 4} into Eq.~\eqref{eq:bound on estimation for alg 2 final 3}, we have
    \begin{align*}
        (\mathrm{ii})
        & \le \gamma_{K}(\delta) \bigg( 
            \sqrt{T} \sqrt{2 H d \log \left( 1 + K \Ucal \Fnorm^2 /(d \lambda) \right)}
            + \frac{256 \eta \Fnorm}{\sqrt{\lambda}} \kappa^{-1} d H \log \left( 1 + K \Ucal \Fnorm^2/(d \lambda) \right) \bigg) 
        \\
        & = \tilde{\Ocal}(d^{3/2} H^{3/2} \sqrt{T} + \kappa^{-1} d^{3/2} H^2) \, .
        \numberthis \label{eq:eq:bound on estimation for alg 2 final 5}
    \end{align*}

    For term $(\mathrm{iii})$, 
    \begin{align*}
        & \sum_{k=1}^K \sum_{h=1}^H \Upsilon^k_h(s^k_h, a^k_h)
        \\
        & = \sum_{k=1}^K \sum_{h=1}^H \bigg(
            H \beta_k \frac{16 \eta \Fnorm}{\sqrt{\lambda}} \max_{s' \in \Scal_{k,h}} \left\| \bar{\varphib}_{k,h,s'} (\omdtheta{k+1}{h}) \right\|^2_{\Bb^{-1}_{k,h}}
            + 6 H \beta_k^2 \max_{s' \in \Scal_{k,h}} \| \varphib_{k,h,s'} \|^2_{\Bb^{-1}_{k,h}}
            \bigg)
        \\
        & \le H \beta_K \frac{16 \eta \Fnorm}{\sqrt{\lambda}} \sum_{k=1}^K \sum_{h=1}^H \max_{s' \in \Scal_{k,h}} \left\| \bar{\varphib}_{k,h,s'} (\omdtheta{k+1}{h}) \right\|^2_{\Bb^{-1}_{k,h}}
            + 6 H \beta_K^2 \sum_{k=1}^K \sum_{h=1}^H \max_{s' \in \Scal_{k,h}} \| \varphib_{k,h,s'} \|^2_{\Ab^{-1}_{k,h}}
        \\
        & \le \beta_K \frac{256 \eta \Fnorm }{\sqrt{\lambda}} \kappa^{-1} d H^2 \log \left( 1 + K \Ucal \Fnorm^2/(d \lambda) \right)
            + 24 \kappa^{-1} d H^2 \beta_K^2 \log \left( 1 + K \Ucal \Fnorm^2/(d \lambda) \right)
        \\
        & = \tilde{\Ocal}(\kappa^{-1} d^2 H^2) \, ,
        \numberthis
        \label{eq:eq:bound on estimation for alg 2 final 6}
    \end{align*}
    where for the second inequality we use the same argument used to derive Eq.~\eqref{eq:eq:bound on estimation for alg 2 final 4} and Lemma~\ref{lemma:generalized elliptical potential}.

    For term $(\mathrm{iv})$, since we have $| \dot{\zeta}^k_{h} | \le 2 H$ and $\EE[ \dot{\zeta}^k_h \mid \Fcal_{k,h} ] = 0 $, which means $\{ \dot{\zeta}^k_h \mid \Fcal_{k,h} \}_{k,h}$ is a martingale difference sequence for any $k \in [K]$ and $h \in [H]$.
    Hence, by applying the Azuma-Hoeffding inequality with probability at least $1 - \delta/4$, we have
    \begin{equation} \label{eq:eq:bound on estimation for alg 2 final 7}
        \sum_{k=1}^K \sum_{h=1}^H \dot{\zeta}^k_h \le 2H \sqrt{2 K H \log(4 / \delta)} \, . 
    \end{equation}

    Combining all results of Eq.~\eqref{eq:bound on estimation for alg 2 final 2},~\eqref{eq:eq:bound on estimation for alg 2 final 5}, 
    ~\eqref{eq:eq:bound on estimation for alg 2 final 6}, and~\eqref{eq:eq:bound on estimation for alg 2 final 7}, we have the desired result.
    \begin{align*}
        \sum_{k=1}^K (\tilde{V}^k_1 - V^{\pi^k}_1) (s^k_1)
        & = \tilde{\Ocal}(d H^{3/2} \sqrt{T} + d^{3/2} H^{3/2} \sqrt{T} + \kappa^{-1}d^{3/2}H^2 + \kappa^{-1} d^2 H^2 + H \sqrt{T})
        \\
        & = \tilde{\Ocal}(d^{3/2} H^{3/2} \sqrt{T} + \kappa^{-1} d^2 H^2) \, .
    \end{align*}
\end{proof}

In the following, we provide the proof of the lemmas used in Lemma~\ref{lemma:estimation bound for alg2}.
\subsubsection{Proof of Lemma~\ref{lemma:auxliary bound for prediction error}}

\begin{proof}[Proof of Lemma~\ref{lemma:auxliary bound for prediction error}]
    Note that
    \begin{align*}
        & \sum_{s' \in \Scal_{s,a}} P_{\omdtheta{k}{h}} (s' \mid s, a) \left\| \bar{\varphib}_{s,a,s'} (\omdtheta{k}{h}) \right\|_{\Bb^{-1}_{k,h}}
        \\ 
        & \le \sum_{s' \in \Scal_{s,a}} P_{\omdtheta{k}{h}} (s' \mid s, a) \left\| \bar{\varphib}_{s,a,s'} (\omdtheta{k+1}{h}) \right\|_{\Bb^{-1}_{k,h}}
        \\
        &\phantom{-}
        + \sum_{s' \in \Scal_{s,a}} P_{\omdtheta{k}{h}} (s' \mid s, a) \left\| \bar{\varphib}_{s,a,s'} (\omdtheta{k}{h}) - \bar{\varphib}_{s,a,s'} (\omdtheta{k+1}{h}) \right\|_{\Bb^{-1}_{k,h}}
        \\
        & \le \sum_{s' \in \Scal_{s,a}} P_{\omdtheta{k+1}{h}} (s' \mid s, a) \left\| \bar{\varphib}_{s,a,s'} (\omdtheta{k+1}{h}) \right\|_{\Bb^{-1}_{k,h}}
            \\ 
            & \phantom{{}={}}
            + \underbrace{ \sum_{s' \in \Scal_{s,a}} \left( P_{\omdtheta{k}{h}} (s' \mid s, a) - P_{\omdtheta{k+1}{h}} (s' \mid s, a) \right) \left\| \bar{\varphib}_{s,a,s'} (\omdtheta{k+1}{h}) \right\|_{\Bb^{-1}_{k,h}} }_{(\mathrm{i})}
            \\
            & \phantom{{}={}}
            + \underbrace{ \sum_{s' \in \Scal_{s,a}} P_{\omdtheta{k}{h}} (s' \mid s, a) \left\| \bar{\varphib}_{s,a,s'} (\omdtheta{k}{h}) - \bar{\varphib}_{s,a,s'} (\omdtheta{k+1}{h}) \right\|_{\Bb^{-1}_{k,h}} }_{(\mathrm{ii})} \, ,
    \end{align*}
    where the first inequality holds by triangle inequality.

    For $(\mathrm{i})$, we have
    \begin{align}
        (\mathrm{i})
        & = \sum_{s' \in \Scal_{s,a}} \nabla P_{\varthetab^k_h} (s' \mid s,a)^\top (\omdtheta{k}{h} - \omdtheta{k+1}{h}) \left\| \bar{\varphib}_{s,a,s'} (\omdtheta{k+1}{h}) \right\|_{\Bb^{-1}_{k,h}}
        \nonumber
        \\
        & \le \sum_{s' \in \Scal_{s,a}} \| \nabla P_{\varthetab^k_h} (s' \mid s,a) \|_{\Bb^{-1}_{k,h}}
        \left\| \omdtheta{k}{h} - \omdtheta{k+1}{h} \right\|_{\Bb_{k,h}}
        \left\| \bar{\varphib}_{s,a,s'} (\omdtheta{k+1}{h}) \right\|_{\Bb^{-1}_{k,h}} 
        \label{eq:auxliary bound for prediction error 1}
    \end{align}
    where in the equality we apply the mean value theorem with $\varthetab^k_h = v \omdtheta{k}{h} + (1-v) \omdtheta{k+1}{h}$ for some $v \in [0,1]$, and the inequality follows by Cauchy-Schwarz inequality.
    Meanwhile, since we have
    \begin{align}
        & P_{\varthetab^k_h} (s' \mid s, a) \bigg( \bar{{\varphib}}_{s, a, s'}(\omdtheta{k+1}{h}) - \sum_{s'' \in \Scal_{s,a}} P_{\varthetab^k_h} (s'' \mid s, a) \bar{\varphib}_{s, a, s''}(\omdtheta{k+1}{h}) \bigg)
        \label{eq:auxliary bound for prediction error 2}
        \\
        & = P_{\varthetab^k_h} (s' \mid s, a)
            \Bigg( \varphib_{s, a, s'} - \sum_{\tilde{s} \in \Scal_{s,a}} P_{\omdtheta{k+1}{h}}(\tilde{s} \mid s,a) \varphib_{s,a,\tilde{s}}
            \nonumber
            \\
            & \phantom{{}={}}
            - \sum_{s'' \in \Scal_{s,a}} P_{\varthetab^k_h} (s'' \mid s, a)  \left[\varphib_{s, a, s''} - \sum_{\tilde{s}} P_{\omdtheta{k+1}{h}}(\tilde{s} \mid s,a) \varphib_{s,a,\tilde{s}} \right]
            \Bigg)
        \nonumber
        \\
        & = P_{\varthetab^k_h} (s' \mid s, a) \varphib_{s, a, s'}
            - P_{\varthetab^k_h} (s' \mid s, a) \sum_{\tilde{s} \in \Scal_{s,a}} P_{\omdtheta{k+1}{h}}(\tilde{s} \mid s,a) \varphib_{s,a,\tilde{s}}
            \nonumber
            \\
            & \phantom{{}={}}
            - P_{\varthetab^k_h} (s' \mid s, a) \sum_{s'' \in \Scal_{s,a}} P_{\varthetab^k_h} (s'' \mid s, a) \varphib_{s, a, s''}
            \nonumber
            \\
            & \phantom{{}={}}
            + P_{\varthetab^k_h} (s' \mid s, a) \bigg( \underbrace{\sum_{s'' \in \Scal_{s,a}} P_{\varthetab^k_h} (s'' \mid s, a)}_{1} \bigg) \sum_{\tilde{s}} P_{\omdtheta{k+1}{h}}(\tilde{s} \mid s,a) \varphib_{s,a,\tilde{s}}
        \nonumber
        \\
        & = P_{\varthetab^k_h} (s' \mid s, a) \varphib_{s, a, s'}
            - P_{\varthetab^k_h} (s' \mid s, a) \sum_{s'' \in \Scal_{s,a}} P_{\varthetab^k_h} (s'' \mid s, a) \varphib_{s, a, s''}
        \nonumber
        \\
        & = \nabla P_{\varthetab^k_h} (s' \mid s,a) \, ,
        \nonumber
    \end{align}
    by substituting~\eqref{eq:auxliary bound for prediction error 2} into~\eqref{eq:auxliary bound for prediction error 1} we have 
    \begin{align}
        &(\mathrm{i}) \nonumber
        \\
        & \le \sum_{s' \in \Scal_{s,a}}
            \left\{
            \left\| P_{\varthetab^k_h} (s' \mid s, a) \bar{{\varphib}}_{s, a, s'}(\omdtheta{k+1}{h}) \right. \right. \nonumber
        \\
        & \phantom{-----} 
        - P_{\varthetab^k_h} (s' \mid s, a) \sum_{s'' \in \Scal_{s,a}} P_{\varthetab^k_h} (s'' \mid s, a) \bar{\varphib}_{s, a, s''}(\omdtheta{k+1}{h}) \Big\|_{\Bb^{-1}_{k,h}}
            \nonumber
            \\
            & \phantom{-----}
            \cdot \left\| \omdtheta{k}{h} - \omdtheta{k+1}{h} \right\|_{\Bb_{k,h}}
            \left\| \bar{\varphib}_{s,a,s'} (\omdtheta{k+1}{h}) \right\|_{\Bb^{-1}_{k,h}} 
            \Bigg\}
            \nonumber
        \\
        & \le 
            \sum_{s' \in \Scal_{s,a}} P_{\varthetab^k_h}(s' \mid s, a) \left\| \bar{\varphib}_{s,a,s'} (\omdtheta{k+1}{h}) \right\|^2_{\Bb^{-1}_{k,h}} \left\| \omdtheta{k}{h} - \omdtheta{k+1}{h} \right\|_{\Bb_{k,h}} 
            \nonumber
            \\
            & \phantom{{}={}}
            + \bigg( \sum_{s' \in \Scal_{s,a}} P_{\varthetab^k_h} (s' \mid s,a) \left\| \bar{\varphib}_{s,a,s'} (\omdtheta{k+1}{h}) \right\|_{\Bb^{-1}_{k,h}} \bigg)^2 
            \left\| \omdtheta{k}{h} - \omdtheta{k+1}{h} \right\|_{\Bb_{k,h}} \, .
            \label{eq:auxliary bound for prediction error 3}
    \end{align}
    Note that by Jensen's inequality, we have
    \begin{align}
        \bigg( \sum_{s' \in \Scal_{s,a}} P_{\varthetab^k_h} (s' \mid s,a) \left\| \bar{\varphib}_{s,a,s'} (\omdtheta{k+1}{h}) \right\|_{\Bb^{-1}_{k,h}} \bigg)^2 
        & = \left( \EE_{s' \sim P_{\varthetab^k_h} (\cdot \mid s,a)} \left[ \left\| \bar{\varphib}_{s,a,s'} (\omdtheta{k+1}{h}) \right\|_{\Bb^{-1}_{k,h}} \right] \right)^2
        \nonumber
        \\
        & \le \EE_{s' \sim P_{\varthetab^k_h} (\cdot \mid s,a)} \left[ \left\| \bar{\varphib}_{s,a,s'} (\omdtheta{k+1}{h}) \right\|^2_{\Bb^{-1}_{k,h}} \right]
        \nonumber
        \\
        & = \sum_{s' \in \Scal_{s,a}} P_{\varthetab^k_h}(s' \mid s, a) \left\| \bar{\varphib}_{s,a,s'} (\omdtheta{k+1}{h}) \right\|^2_{\Bb^{-1}_{k,h}} \, .
        \label{eq:auxliary bound for prediction error 4}
    \end{align}

    Also, we introduce the following lemma:
    \begin{lemma} \label{lemma:zhang_lemma20}
        For any $k \in [K]$ and $h \in [H]$, the following holds:
        \begin{equation*}
           \left\| \omdtheta{k+1}{h} - \omdtheta{k}{h} \right\|_{\Bb_{k,h}} \le \frac{4 \eta \Fnorm}{\sqrt{\lambda}} \, .
        \end{equation*}
    \end{lemma}

    Then, substituting~\eqref{eq:auxliary bound for prediction error 4} into~\eqref{eq:auxliary bound for prediction error 3}, we have
    \begin{align}
        (\mathrm{i})
        & \le 2 \sum_{s' \in \Scal_{s,a}} P_{\varthetab^k_h}(s' \mid s, a) \left\| \bar{\varphib}_{s,a,s'} (\omdtheta{k+1}{h}) \right\|^2_{\Bb^{-1}_{k,h}} \left\| \omdtheta{k}{h} - \omdtheta{k+1}{h} \right\|_{\Bb_{k,h}}
        \nonumber
        \\
        & \le \frac{8 \eta \Fnorm}{\sqrt{\lambda}} \sum_{s' \in \Scal_{s,a}} P_{\varthetab^k_h}(s' \mid s, a) \left\| \bar{\varphib}_{s,a,s'} (\omdtheta{k+1}{h}) \right\|^2_{\Bb^{-1}_{k,h}}
        \nonumber
        \\
        & \le \frac{8 \eta \Fnorm}{\sqrt{\lambda}} \max_{s' \in \Scal_{s,a}} \left\| \bar{\varphib}_{s,a,s'} (\omdtheta{k+1}{h}) \right\|^2_{\Bb^{-1}_{k,h}} \, ,
        \label{eq:auxliary bound for prediction error 5}
    \end{align}
    where the second inequality comes from Lemma~\ref{lemma:zhang_lemma20}, and the last inequality holds due to $\sum_{s' \in \Scal_{s,a}} P_{\varthetab^k_h}(s' \mid s, a) = 1$.

    For $(\mathrm{ii})$, we have
    \begin{align}
        (\mathrm{ii})
        & = \sum_{s' \in \Scal_{s,a}} P_{\omdtheta{k}{h}} (s' \mid s, a) \left\| \bar{\varphib}_{s,a,s'} (\omdtheta{k}{h}) - \bar{\varphib}_{s,a,s'} (\omdtheta{k+1}{h}) \right\|_{\Bb^{-1}_{k,h}}
        \nonumber
        \\
        & = \sum_{s' \in \Scal_{s,a}} P_{\omdtheta{k}{h}} (s' \mid s, a) \left\| \EE_{\tilde{s} \sim P_{\omdtheta{k}{h}} (\cdot \mid s, a) } \left[ \varphib_{s,a,\tilde{s}} \right] - \EE_{\tilde{s} \sim P_{\omdtheta{k+1}{h}} (\cdot \mid s, a) } \left[ \varphib_{s,a,\tilde{s}} \right] \right\|_{\Bb^{-1}_{k,h}}
        \nonumber
        \\
        & =  \left\| \sum_{\tilde{s} \in \Scal_{s,a}} \left( P_{\omdtheta{k}{h}}(\tilde{s} \mid s,a) - P_{\omdtheta{k+1}{h}}(\tilde{s} \mid s,a) \right) \varphib_{s,a,\tilde{s}} \right\|_{\Bb^{-1}_{k,h}}
        \nonumber
        \\
        & = \left\| \sum_{\tilde{s} \in \Scal_{s,a}} \left( P_{\omdtheta{k}{h}}(\tilde{s} \mid s,a) - P_{\omdtheta{k+1}{h}}(\tilde{s} \mid s,a) \right)
            \left( \varphib_{s,a,\tilde{s}} - \EE_{s' \sim P_{ \omdtheta{k+1}{h} } (\cdot \mid s, a) } \left[ \varphib_{s,a,s'} \right]  \right) \right\|_{\Bb^{-1}_{k,h}}
        \nonumber
        \\ 
        & = \left\| \sum_{\tilde{s} \in \Scal_{s,a}} \left( P_{\omdtheta{k}{h}}(\tilde{s} \mid s,a) - P_{\omdtheta{k+1}{h}}(\tilde{s} \mid s,a) \right) \bar{\varphib}_{s,a,\tilde{s}} (\omdtheta{k+1}{h}) \right\|_{\Bb^{-1}_{k,h}}
        \nonumber
        \\ 
        & \le \frac{8 \eta \Fnorm}{\sqrt{\lambda}}  \max_{s' \in \Scal_{s,a}}  \left\| \bar{\varphib}_{s,a,s'} (\omdtheta{k+1}{h}) \right\|^2_{\Bb^{-1}_{k,h}} \, ,
        \label{eq:auxliary bound for prediction error 6}
    \end{align}
    where the last inequality is obtained through the same argument as used to bound $(\mathrm{i})$.
    Combining the results of Eq.~\eqref{eq:auxliary bound for prediction error 5} and Eq.~\eqref{eq:auxliary bound for prediction error 6}, we have
    \begin{align*}
        & \sum_{s' \in \Scal_{s,a}} P_{\omdtheta{k}{h}} (s' \mid s, a) \left\| \bar{\varphib}_{s,a,s'} (\omdtheta{k}{h}) \right\|_{\Bb^{-1}_{k,h}}
        \\
        & \le \sum_{s' \in \Scal_{s,a}} P_{\omdtheta{k+1}{h}} (s' \mid s, a) \left\| \bar{\varphib}_{s,a,s'} (\omdtheta{k+1}{h}) \right\|_{\Bb^{-1}_{k,h}}
        + \frac{16 \eta \Fnorm}{\sqrt{\lambda}}  \max_{s' \in \Scal_{s,a}} \left\| \bar{\varphib}_{s,a,s'} (\omdtheta{k+1}{h}) \right\|^2_{\Bb^{-1}_{k,h}}
    \end{align*}
\end{proof}

\subsubsection{Proof of Lemma~\ref{lemma:elliptical potential for centralized feature}}
\begin{proof}[Proof of Lemma~\ref{lemma:elliptical potential for centralized feature}]
    Note that
    \begin{align*}
        \Bb_{k+1, h} 
        & = \Bb_{k,h} + \sum_{s' \in \Scal_{k,h}} P_{\omdtheta{k+1}{h}} (s' \mid s^k_h, a^k_h) \bar{\varphib}_{k,h,s'} (\omdtheta{k+1}{h}) \bar{\varphib}_{k,h,s'} (\omdtheta{k+1}{h})^\top
        \\
        & = \Bb_{k,h} + \sum_{s' \in \Scal_{k,h}} \tilde{\varphib}_{k,h,s'} (\omdtheta{k+1}{h}) \tilde{\varphib}_{k,h,s'} (\omdtheta{k+1}{h})^\top \, ,
    \end{align*}
    where we define $\tilde{\varphib}_{k,h,s'} (\omdtheta{k+1}{h}) := \sqrt{P_{\omdtheta{k+1}{h}} (s' \mid s^k_h, a^k_h)} \bar{\varphib}_{k,h,s'} (\omdtheta{k+1}{h})$.
    Then, we have
    \begin{align*}
        \det (\Bb_{k+1, h}) 
        & = \det (\Bb_{k, h}) 
            \det \left(\Ib_d + \Bb_{k, h}^{-\frac{1}{2}} \sum_{s' \in \Scal_{k,h}} \tilde{\varphib}_{k,h,s'} (\omdtheta{k+1}{h}) \tilde{\varphib}_{k,h,s'} (\omdtheta{k+1}{h})^\top \Bb_{k, h}^{-\frac{1}{2}} \right)
        \\
        & = \det (\Bb_{k, h})
            \left( 1 + \sum_{s' \in \Scal_{k,h}} \left\| \tilde{\varphib}_{k,h,s'} (\omdtheta{k+1}{h}) \right\|^2_{\Bb^{-1}_{k,h}} \right)
        \\
        & = \det ( \lambda \Ib_d ) \, \prod^K_{k=1} \left( 1 + \sum_{s' \in \Scal_{k,h}} \left\| \tilde{\varphib}_{k,h,s'} (\omdtheta{k+1}{h}) \right\|^2_{\Bb^{-1}_{k,h}} \right) \, .
    \end{align*}
    Taking the logarithm on both sides yields
    \begin{align*}
        \log \frac{\det (\Bb_{k+1, h}) }{ \det (\lambda \Ib_d) }
        = \sum^K_{k=1} \log \left( 1 + \sum_{s' \in \Scal_{k,h}} \left\| \tilde{\varphib}_{k,h,s'} (\omdtheta{k+1}{h}) \right\|^2_{\Bb^{-1}_{k,h}} \right) \, .
    \end{align*}
    On the other hand, since $\lambda \ge \Fnorm^2$, 
    \begin{align*}
        \sum_{s' \in \Scal_{k,h}} \left\| \tilde{\varphib}_{k,h,s'} (\omdtheta{k+1}{h}) \right\|^2_{\Bb^{-1}_{k,h}} 
        & \le \sum_{s' \in \Scal_{k,h}} \frac{1}{\lambda} \left\| \tilde{\varphib}_{k,h,s'} (\omdtheta{k+1}{h}) \right\|^2_2
        \\
        & = \sum_{s' \in \Scal_{k,h}} \frac{1}{\lambda} P_{\omdtheta{k+1}{h}} (s' \mid s^k_h, a^k_h) \left\| \bar{\varphib}_{k,h,s'} (\omdtheta{k+1}{h}) \right\|^2_2
        \\
        & \le \frac{ \Fnorm^2 }{ \lambda } \sum_{s' \in \Scal_{k,h}} P_{\omdtheta{k+1}{h}} (s' \mid s^k_h, a^k_h)
        \\
        & \le 1 \, ,
    \end{align*}
    where the last inequality uses $\sum_{s' \in \Scal_{k,h}} P_{\omdtheta{k+1}{h}} (s' \mid s^k_h, a^k_h) = 1$.
    From the fact that $z \le 2 \log (1 + z)$ for any $z \in [0, 1]$, it follows that
    \begin{align*}
        \sum^K_{k=1} \log \left( 1 + \sum_{s' \in \Scal_{k,h}} \left\| \tilde{\varphib}_{k,h,s'} (\omdtheta{k+1}{h}) \right\|^2_{\Bb^{-1}_{k,h}} \right) 
        \ge \sum^K_{k=1} \frac{1}{2} \sum_{s' \in \Scal_{k,h}} \left\| \tilde{\varphib}_{k,h,s'} (\omdtheta{k+1}{h}) \right\|^2_{\Bb^{-1}_{k,h}} \, .
    \end{align*}
    Finally, we obtain
    \begin{align*}
        \sum^K_{k=1} \sum_{s' \in \Scal_{k,h}} \left\| \tilde{\varphib}_{k,h,s'} (\omdtheta{k+1}{h}) \right\|^2_{\Bb^{-1}_{k,h}} 
        & \le 2 \sum^K_{k=1} \log \left( 1 + \sum_{s' \in \Scal_{k,h}} \left\| \tilde{\varphib}_{k,h,s'} (\omdtheta{k+1}{h}) \right\|^2_{\Bb^{-1}_{k,h}} \right)
        \\
        & = 2 \log \frac{\det (\Bb_{K+1, h}) }{ \det (\lambda \Ib_d) }
        \\
        & \le 2 d \log \left( 1 + \frac{K \Ucal \Fnorm^2}{d \lambda} \right) \, ,
    \end{align*}
    where the last inequality follows by the determinant-trace inequality (Lemma~\ref{lemma:det-trace ineq}).    
\end{proof}

\subsubsection{Proof of Lemma~\ref{lemma:auxliary bound for estimation bound (alg2)}}
\begin{proof}[Proof of Lemma~\ref{lemma:auxliary bound for estimation bound (alg2)}]
    Since $\xib^{(m)}_{k,h} \sim \Ncal (\zero, \sigma_k^2 \Bb^{-1}_{k,h})$, by Lemma~\ref{auxiliary lemma: gaussian noise concentration} for each $m \in [M]$, we have
    \begin{align*}
        \| \xib^{(m)}_{k,h} \|_{\Bb_{k,h}} \le C_{\xib} \sigma_k \sqrt{d \log (Md / \delta)} \, .
    \end{align*}
    Following the result of Lemma~\ref{lemma:auxliary bound for prediction error}, we have 
    \begin{align*}
        \sum_{s' \in \Scal_{k,h}} P_{\omdtheta{k}{h}} (s' \mid s^k_h, a^k_h) \left\| \bar{\varphib}_{k,h,s'} (\omdtheta{k}{h}) \right\|_{\Bb^{-1}_{k,h}} 
        & \le \sum_{s' \in \Scal_{k,h}} P_{\omdtheta{k+1}{h}} (s' \mid s^k_h, a^k_h) \left\| \bar{\varphib}_{k,h,s'} (\omdtheta{k+1}{h}) \right\|_{\Bb^{-1}_{k,h}}
            \\
            & \phantom{{}={}}
            + \frac{16 \eta \Fnorm}{\sqrt{\lambda}} \max_{s' \in \Scal_{k,h}} \left\| \bar{\varphib}_{k,h,s'} (\omdtheta{k+1}{h}) \right\|^2_{\Bb^{-1}_{k,h}}  \, .
    \end{align*}
    Then, we obtain
    \begin{align*}
        & \sum_{s' \in \Scal_{k,h}} P_{\omdtheta{k}{h}} (s' \mid s^k_h, a^k_h) \bar{\varphib}_{k,h,s'} (\omdtheta{k}{h})^\top \xib^{s'}_{k,h}
        \\
        & \le \sum_{s' \in \Scal_{k,h}} P_{\omdtheta{k}{h}} (s' \mid s^k_h, a^k_h) \left\| \bar{\varphib}_{k,h,s'} (\omdtheta{k}{h}) \right\|_{\Bb^{-1}_{k,h}} \| \xib^{s'}_{k,h} \|_{\Bb_{k,h}}
        \\
        & \le C_{\xib} \sigma_k \sqrt{d \log (Md / \delta)} \sum_{s' \in \Scal_{k,h}} P_{\omdtheta{k}{h}} (s' \mid s^k_h, a^k_h) \| \bar{\varphi}_{k,h,s'} (\omdtheta{k}{h}) \|_{\Bb^{-1}_{k,h}}
        \\
        &\le \gamma_k (\delta) \bigg( \sum_{s' \in \Scal_{k,h}} P_{\omdtheta{k+1}{h}} (s' \mid s^k_h, a^k_h) \left\| \bar{\varphib}_{k,h,s'} (\omdtheta{k+1}{h}) \right\|_{\Bb^{-1}_{k,h}}
        \\
        &\phantom{-----} + \frac{16 \eta \Fnorm}{\sqrt{\lambda}} \max_{s' \in \Scal_{k,h}} \left\| \bar{\varphib}_{k,h,s'} (\omdtheta{k+1}{h}) \right\|^2_{\Bb^{-1}_{k,h}} \bigg) \, .
    \end{align*}
\end{proof}

\subsubsection{Proof of\texorpdfstring{~\Cref{lemma:zhang_lemma20}}{Lemma 25}}

\begin{proof}[Proof of Lemma~\ref{lemma:zhang_lemma20}]
    We provide a proof for~\Cref{lemma:zhang_lemma20} since it is slight modification of Lemma 20 of~\cite{zhang2023online}.
    From the definition, we know that
    \begin{align*}
        \left( \omdtheta{k+1}{h} \right)^\top \nabla \ell_{k,h}(\omdtheta{k}{h}) + \frac{1}{2 \eta} \left\| \omdtheta{k+1}{h} - \omdtheta{k}{h} \right\|^2_{\tilde{\Bb}_{k,h}}  
        \le \left( \omdtheta{k}{h} \right)^\top \nabla \ell_{k,h}(\omdtheta{k}{h}) \, .
    \end{align*}
    By rearranging the terms, the following holds:
    \begin{align*}
        \frac{1}{2 \eta} \left\| \omdtheta{k+1}{h} - \omdtheta{k}{h} \right\|^2_{\tilde{\Bb}_{k,h}}  
        &\le \left( \omdtheta{k}{h} - \omdtheta{k+1}{h} \right)^\top \nabla \ell_{k,h}(\omdtheta{k}{h}) \\
        &\le \left\| \omdtheta{k}{h} - \omdtheta{k+1}{h} \right\|_{\tilde{\Bb}_{k,h}} \left\| \nabla \ell_{k,h}(\omdtheta{k}{h}) \right\|_{\tilde{\Bb}^{-1}_{k,h}}
    \end{align*}
    Thus, we get
    \begin{align*}
        \left\| \omdtheta{k+1}{h} - \omdtheta{k}{h} \right\|_{\tilde{\Bb}_{k,h}}  
        \le 2\eta \left\| \nabla \ell_{k,h}(\omdtheta{k}{h}) \right\|_{\tilde{\Bb}^{-1}_{k,h}} \, .
    \end{align*}
    Since $\Bb_{k,h} \preceq \tilde{\Bb}_{k,h}$ and $\tilde{\Bb}^{-1}_{k,h} \preceq \lambda^{-1} \Ib_d$, we obtain
    \begin{align}
        \left\| \omdtheta{k+1}{h} - \omdtheta{k}{h} \right\|_{\Bb_{k,h}} 
        \le \left\| \omdtheta{k+1}{h} - \omdtheta{k}{h} \right\|_{\tilde{\Bb}_{k,h}}  
        \le 2\eta \left\| \nabla \ell_{k,h}(\omdtheta{k}{h}) \right\|_{\tilde{\Bb}^{-1}_{k,h}} 
        \le \frac{2 \eta}{\sqrt{\lambda}} \left\| \nabla \ell_{k,h}(\omdtheta{k}{h}) \right\|_2 
        \le \frac{4 \eta \Fnorm}{\sqrt{\lambda}} \, .
        \label{eq:modified_zhang_lemma_20}
    \end{align}
    For the last inequality of~\eqref{eq:modified_zhang_lemma_20}, we provide the upper bound of $l_2$-norm of $\nabla \ell_{k,h} (\thetab)$.
    Since
    \begin{equation*}
        \ell_{k,h} (\thetab) = - \sum_{s' \in \Scal_{k,h}} y^k_h(s') \log P_{\thetab}(s' \mid s^k_h, a^k_h) \, ,
    \end{equation*}
    the gradient of the loss function is given by
    \begin{align*}
        \nabla \ell_{k,h} (\thetab) 
        &= - \sum_{s' \in \Scal_{k,h}} y^k_h(s') \left( \varphib_{s,a,s'} - \sum_{s'' \in \Scal_{k,h}} P_{\thetab}(s'' \mid s^k_h, a^k_h) \varphib_{s,a,s''} \right) \\
        &=  \sum_{s' \in \Scal_{k,h}} y^k_h(s') \sum_{s'' \in \Scal_{k,h}} P_{\thetab}(s'' \mid s^k_h, a^k_h) \varphib_{s,a,s''} - \sum_{s' \in \Scal_{k,h}}  y^k_h(s') \varphib_{s,a,s'} \\
        &= \sum_{s'' \in \Scal_{k,h}} P_{\thetab}(s'' \mid s^k_h, a^k_h) \varphib_{s,a,s''} - \sum_{s' \in \Scal_{k,h}}  y^k_h(s') \varphib_{s,a,s'} \\
        &= \sum_{s' \in \Scal_{k,h}} \left( P_{\thetab}(s' \mid s^k_h, a^k_h) - y^k_h(s') \right) \varphib_{s,a,s'} \, .
    \end{align*}
    Therefore, we have
    \begin{align*}
        \left\| \nabla \ell_{k,h} (\thetab) \right\|_2 
        &= \left\| \sum_{s' \in \Scal_{k,h}} \left( P_{\thetab}(s' \mid s^k_h, a^k_h) - y^k_h(s') \right) \varphib_{s,a,s'} \right\|_2 \\
        &\le \sum_{s' \in \Scal_{k,h}} \left| P_{\thetab}(s' \mid s^k_h, a^k_h) - y^k_h(s') \right| \| \varphib_{s,a,s'} \|_2
        \\
        &\le 2 \Fnorm 
    \end{align*}
    and this concludes the proof.
\end{proof}

\subsection{Bound on Pessimism Part}
In this section, we provide the upper bound on the pessimism part of the regret: $\sum_{k=1}^K (V^*_1 - \tilde{V}^k_{1})(s^k_1)$.
\begin{lemma}[Bound on pessimism] \label{lemma:pessimism bound for alg2}
    For any $\delta$ with $0 < \delta < \Phi(-1)/2$, let $\sigma_k = H \beta_k$.
    If $\lambda = \Ocal(\Fnorm^2 d \log \Ucal)$ and we take multiple sample size $M = \lceil 1 - \frac{\log(H \Ucal)}{\log \Phi(1)} \rceil$, then with probability at least $1 - \delta/2$, we have
    \begin{equation*}
        \sum_{k=1}^K (V^*_1 - V^k_{1})(s^k_1)
        = \BigOTilde \left( d^{3/2} H^{3/2} \sqrt{T} + \kappa^{-1} d^2 H^2 \right) \, .
    \end{equation*}
\end{lemma}

\begin{proof}[Proof of Lemma~\ref{lemma:pessimism bound for alg2}]
    As seen in Lemma~\ref{lemma:stochastic optimism of alg2}, by using multiple sampling technique we show that 
    the optimistic randomized value function $\tilde{V}$ of $\texttt{ORRL-MNL}$ is optimistic than the true optimal value with constant probability
    Hence, with the same argument used in Lemma~\ref{lemma:bound of pessimism part}, we can show that the pessimism term of $\texttt{ORRL-MNL}$ is upper bounded by a bound of the estimation term times the inverse probability of being optimistic, i.e.,
    \begin{equation*}
        \sum_{k=1}^K \left(V^*_1 - V^k_{1} \right)(s^k_1) \le \tilde{\Ocal} \left( \frac{1}{\Phi(-1)} \sum_{k=1}^K \left(V^k_{1} - V^{\pi^k}_1 \right)(s^k_1) \right) \, .
    \end{equation*}
\end{proof}

\subsection{Regret Bound of \texorpdfstring{$\texttt{ORRL-MNL}$}{ORRL-MNL}}
\begin{proof}[Proof of Theorem~\ref{thm:alg 2}]
    Since both Lemma~\ref{lemma:estimation bound for alg2} and Lemma~\ref{lemma:pessimism bound for alg2} holds with probability at least $1 - \delta / 2$ respectively, by taking the union bound we conclude the proof.
\end{proof}

\section{Optimistic Exploration Extension} \label{appx:ucrl-mnl plus}

In this section, we introduce $\AlgUCB$ (Algorithm~\ref{alg:ucrl-mnl plus}), which is both \textit{computationally} and \textit{statistically} efficient for MNL-MDPs with UCB-based exploration. 
The main difference compared to $\texttt{ORRL-MNL}$ is that $\AlgUCB$ constructs an \textit{optimistic value function} that is greater than the optimal value function with high probability.
At each episode $k \in [K]$, with the estimated transition core parameter $\omdtheta{k}{h}$~\eqref{eq:online mirror descent theta}, for $(s,a) \in \Scal \times \Acal$, set $\hat{Q}^k_{H+1}(s,a) = 0$.
For each $h \in [H]$, 
\begin{equation} \label{eq:q-function for ucb alg}
    \hat{Q}^k_h (s,a) := r(s,a) + \sum_{s' \in \Scal_{s,a}} P_{\omdtheta{k}{h}} (s' \mid s, a) \hat{V}^k_{h+1}(s') + \obonus_{k,h}(s,a) \, ,
\end{equation}
where $\hat{V}^k_{h}(s) := \min \{ \max_{a \in \Acal} \hat{Q}^k_h(s,a), H \}$ and $\obonus_{k,h}(s,a)$ is the \textit{optimistic bonus term} defined by
\begin{equation*}
    \obonus_{k,h}(s,a) := 
        H \beta_k \sum_{s' \in \Scal_{s,a}} P_{\omdtheta{k}{h}}(s' \mid s,a) \| \bar{\varphib}(s,a,s'; \omdtheta{k}{h}) \|_{\Bb_{k,h}^{-1}}
        + 3 H \beta_k^2 \max_{s' \in \Scal_{s,a}} \| \varphib(s,a,s') \|^2_{\Bb_{k,h}^{-1}} \, .
\end{equation*}
Based on these \textit{optimistic value function} $\hat{Q}^k_h$, at each episode the agent plays a greedy action with respect to $\hat{Q}^k_h$ as summarized in Algorithm~\ref{alg:ucrl-mnl plus}.

\begin{algorithm}[h!] 
    \caption{$\AlgUCB$ (Upper Confidence RL for MNL-MDPs)} \label{alg:ucrl-mnl plus}
    \begin{algorithmic}[1]
        \STATE \textbf{Inputs:} Episodic MDP $\Mcal$, Feature map $\varphib:\Scal \times \Acal \times \Scal \rightarrow \RR^{d}$, Number of episodes~$K$, Regularization parameter $\lambda$, Confidence radius $\{ \beta_k \}_{k=1}^K$, Step size $\eta$
        \STATE \textbf{Initialize: } $\omdtheta{1}{h} = \zero_d $, $\Bb_{1,h} = \lambda \Ib_d$ for all $h \in [H]$
        \FOR{episode $k=1,2, \cdots, K$}
            \STATE Observe $s^k_1$ and set $\left\{ \hat{Q}^k_h (\cdot, \cdot) \right\}_{h \in [H]}$ as described in~\eqref{eq:q-function for ucb alg}
            \FOR{horizon $h=1, 2, \cdots, H$} 
                \STATE Select $a^k_h = \argmax_{a \in \Acal} \hat{Q}^k_h(s^k_h, a)$ and observe $s^k_{h+1}$
                \STATE Update $\tilde{\Bb}_{k,h} = \Bb_{k,h} + \eta \nabla^2 \ell_{k,h} (\omdtheta{k}{h})$ and $\omdtheta{k+1}{h}$ as in~\eqref{eq:online mirror descent theta}
                \STATE Update $\Bb_{k+1,h} = \Bb_{k,h} + \nabla^2 \ell_{k,h} (\omdtheta{k+1}{h})$
            \ENDFOR
        \ENDFOR
    \end{algorithmic}
\end{algorithm}

The main difference in regret analysis lies in ensuring the optimism of the estimated value function $\hat{Q}^k_h$ (Lemma~\ref{lemma:optimsm}). 
In the following statement (formal statement of Corollary~\ref{coro:ucrl-mnl plus}), we provide a regret guarantee for $\AlgUCB$, which enjoys the tightest regret bound for MNL-MDPs. 

\begin{theorem}[Regret Bound of $\AlgUCB$] \label{thm:ucrl-mnl plus}
    Suppose that Assumption~\ref{assm:mnl-mdp}-~\ref{assm:positive kappa} hold.
    For any $\delta \in (0,1)$, if we set the input parameters in Algorithm~\ref{alg:ucrl-mnl plus} as $\lambda = \Ocal(\Fnorm^2 d \log \Ucal), \beta_k = \Ocal(\sqrt{d} \log \Ucal \log(kH))$ $\eta = \Ocal(\log \Ucal)$, then with probability at least $1 - \delta$, the cumulative regret of the $\AlgUCB$ policy $\pi$ is upper-bounded by
    \begin{align*}
        \Regret_\pi (K)
        = \BigOTilde \left( d H^{3/2} \sqrt{T} + \kappa^{-1} d^2 H^2 \right) \, ,
    \end{align*}        
    where $T=KH$ is the total number of time steps.
\end{theorem}

\begin{proof}[Proof of Theorem~\ref{thm:ucrl-mnl plus}]
    By Lemma~\ref{lemma:good event prob_2}, suppose that the good event $\Gfrak(K, \delta')$ holds with probability at least $1 - \delta$.
    Then, we show that the optimistic value function $\hat{Q}^k_h$ is deterministically greater than the true optimal value function as follows:
    \begin{lemma}[Optimism] \label{lemma:optimsm}
        Suppose that the event $\Gfrak^{\Delta}_{k,h}(\delta)$ holds for all $k \in [K]$ and $h \in [H]$. 
        Then for any $(s,a) \in \Scal \times \Acal$, we have
        \begin{equation*}
            Q^*_h(s,a) \le \hat{Q}^k_h(s,a) \, .
        \end{equation*}
    \end{lemma}

    Conditioned on $\Gfrak(K, \delta')$, by Lemma~\ref{lemma:optimsm} we have
    \begin{align}
        (V^*_1 - V^{\pi^k}_1) (s^k_1)
        & = Q^*_1(s^k_1, \pi^*(s^k_1)) - Q^{\pi^k}_1 (s^k_1, a^k_1)
        \nonumber
        \\
        & \le \hat{Q}^k_1(s^k_1, \pi^*(s^k_1)) - Q^{\pi^k}_1 (s^k_1, a^k_1)
        \nonumber
        \\
        & \le \hat{Q}^k_1(s^k_1, a^k_1) - Q^{\pi^k}_1 (s^k_1, a^k_1)
            = \obonus_{k,1}(s^k_1, a^k_1) + P_1(\hat{V}^k_2 - V^{\pi^k}_2)(s^k_1, a^k_1) \, .
        \nonumber
    \end{align}
    Note that
    \begin{align*}
        P_1(\hat{V}^k_2 - V^{\pi^k}_2)(s^k_1, a^k_1)
        = \EE_{\tilde{s} \mid s^k_1, a^k_1} \left[ (\hat{V}^k_2 - V^{\pi^k}_2)(\tilde{s}) \right]
        = (\hat{V}^k_2 - V^{\pi^k}_2)(s^k_2) + \dot{\zeta}^k_1 \, ,
    \end{align*}
    where we denote $\zeta^k_h := (\hat{V}^k_{h+1} - V^{\pi^k}_{h+1})(s^k_{h+1}) - \EE_{\tilde{s} \mid s^k_h, a^k_h} \left[ (\hat{V}^k_{h+1} - V^{\pi^k}_{h+1})(\tilde{s}) \right]$.
    Then, with the same argument, we have
    \begin{align*}
        (V^*_1 - V^{\pi^k}_1) (s^k_1)
        \le \sum_{h=1}^{H} \obonus_{k,h}(s^k_h, a^k_h) + \sum_{h=1}^{H} \dot{\zeta}^k_h \, .
    \end{align*}
    By summing over all episodes, we have
    \begin{align}
        \Regret_\pi (K)
        \le \sum_{k=1}^K \sum_{h=1}^{H} \obonus_{k,h}(s^k_h, a^k_h) + \sum_{k=1}^K \sum_{h=1}^{H} \dot{\zeta}^k_h \, .
    \end{align}

    On the other hand, note that
    \begin{align*}
        & \sum_{k=1}^K \sum_{h=1}^{H} \obonus_{k,h}(s^k_h, a^k_h)
        \\
        & = \sum_{k=1}^K \sum_{h=1}^H H \beta_k \sum_{s' \in \Scal_{k,h}} P_{\omdtheta{k}{h}}(s' \mid s^k_h,a^k_h) \| \bar{\varphib}_{k,h,s'}(\omdtheta{k}{h}) \|_{\Bb_{k,h}^{-1}}
        + \sum_{k=1}^K \sum_{h=1}^H 3 H \beta_k^2 \max_{s' \in \Scal_{k,h}} \| \varphib_{k,h,s'} \|^2_{\Bb_{k,h}^{-1}}
        \\
        & \le  H \beta_K \sum_{k=1}^K \sum_{h=1}^H \sum_{s' \in \Scal_{k,h}} P_{\omdtheta{k}{h}}(s' \mid s^k_h,a^k_h) \| \bar{\varphib}_{k,h,s'}(\omdtheta{k}{h}) \|_{\Bb_{k,h}^{-1}}
        \\
        &\phantom{-} + 3 H \beta_K^2 \sum_{k=1}^K \sum_{h=1}^H \max_{s' \in \Scal_{k,h}} \| \varphib_{k,h,s'} \|^2_{\Bb_{k,h}^{-1}} 
        \\
        & \le \underbrace{H \beta_K \sum_{k=1}^K \sum_{h=1}^H \sum_{s' \in \Scal_{k,h}} P_{\omdtheta{k+1}{h}} (s' \mid s^k_h, a^k_h) \left\| \bar{\varphib}_{k,h,s'} (\omdtheta{k+1}{h}) \right\|_{\Bb^{-1}_{k,h}}}_{(\mathrm{i})}
            \\
            & \phantom{{}={}}
            + \underbrace{\frac{16 \eta \Fnorm}{\sqrt{\lambda}} H \beta_K \sum_{k=1}^K \sum_{h=1}^H \max_{s' \in \Scal_{k,h}} \left\| \bar{\varphib}_{k,h,s'} (\omdtheta{k+1}{h}) \right\|^2_{\Bb^{-1}_{k,h}}}_{(\mathrm{ii})}            
            + \underbrace{3 H \beta_K^2 \sum_{k=1}^K \sum_{h=1}^H \max_{s' \in \Scal_{k,h}} \| \varphib_{k,h,s'} \|^2_{\Bb_{k,h}^{-1}}}_{(\mathrm{iii})} \, ,
    \end{align*}
    where the last inequality follows by Lemma~\ref{lemma:auxliary bound for prediction error}.

    Term $(\mathrm{i})$ can be bounded as in Eq.~\eqref{eq:bound on estimation for alg 2 final 2}:
    \begin{align*}
        H \beta_K \sum_{k=1}^K \sum_{h=1}^H \sum_{s' \in \Scal_{k,h}} P_{\omdtheta{k+1}{h}} (s' \mid s^k_h, a^k_h) \left\| \bar{\varphib}_{k,h,s'} (\omdtheta{k+1}{h}) \right\|_{\Bb^{-1}_{k,h}} 
        = \tilde{\Ocal}(d H^{3/2} \sqrt{T}) \, .
        \numberthis
        \label{eq:ucrl-mnl final 1}
    \end{align*}

    For term $(\mathrm{ii})$, recall that as in Eq.~\eqref{eq:eq:bound on estimation for alg 2 final 4} we have
    \begin{align*}
        \sum_{k=1}^K \sum_{h=1}^H \max_{s' \in \Scal_{k,h}} \left\| \bar{\varphib}_{k,h,s'} (\omdtheta{k+1}{h}) \right\|^2_{\Bb^{-1}_{k,h}} 
        \le 16 \kappa^{-1} d H \log \left( 1 + \frac{K \Ucal \Fnorm^2}{d \lambda} \right) \, .
    \end{align*}
    Then, we have
    \begin{align*}
        \frac{16 \eta \Fnorm}{\sqrt{\lambda}} H \beta_K \sum_{k=1}^K \sum_{h=1}^H \max_{s' \in \Scal_{k,h}} \left\| \bar{\varphib}_{k,h,s'} (\omdtheta{k+1}{h}) \right\|^2_{\Bb^{-1}_{k,h}}
        = \tilde{\Ocal}(\kappa^{-1} d H^2) \, .
        \numberthis
        \label{eq:ucrl-mnl final 2}
    \end{align*}

    For term $(\mathrm{iii})$, since we have
    \begin{align*}
        3 H \beta_K^2 \sum_{k=1}^K \sum_{h=1}^H \max_{s' \in \Scal_{k,h}} \| \varphib_{k,h,s'} \|^2_{\Bb_{k,h}^{-1}}
        & \le 3 H \beta_K^2 \sum_{k=1}^K \sum_{h=1}^H \max_{s' \in \Scal_{k,h}} \| \varphib_{k,h,s'} \|^2_{\Ab_{k,h}^{-1}}
        \\
        & \le 12 \kappa^{-1} d H^2 \beta_K^2 \log \left( 1 + K \Ucal \Fnorm^2/(d \lambda) \right)
        \\
        & = \tilde{\Ocal}(\kappa^{-1} d^2 H^2) \, .
        \numberthis
        \label{eq:ucrl-mnl final 3}
    \end{align*}

    Combining the results of Eq.~\eqref{eq:ucrl-mnl final 1},~\eqref{eq:ucrl-mnl final 2}, and~\eqref{eq:ucrl-mnl final 3}, we have 
    \begin{equation*}
        \sum_{k=1}^K \sum_{h=1}^{H} \obonus_{k,h}(s^k_h, a^k_h)
        = \tilde{\Ocal}(d H^{3/2} \sqrt{T} + \kappa^{-1} d^2 H^2) \, .
    \end{equation*}

    Finally, by Azuma-Hoeffiding inequality as in Eq.~\eqref{eq:eq:bound on estimation for alg 2 final 7} we have
    \begin{equation*}
        \sum_{k=1}^K \sum_{h=1}^H \dot{\zeta}^k_h = \tilde{\Ocal}(H \sqrt{T}) \, .
    \end{equation*}
    This concludes the proof.    
\end{proof}

In the following, we provide the proof of Lemma~\ref{lemma:optimsm}.
\subsection{Optimism}

\begin{proof}[Proof of Lemma~\ref{lemma:optimsm}]
    We prove this by backwards induction on $h$.
    For the base case $h = H$, since $V^*_{H+1}(s) = \hat{V}^k_{H+1}(s) = 0$ for all $s \in \Scal$, we have
    \begin{equation*}
        \hat{Q}^k_H(s,a) = r(s,a) = Q^*_H (s,a) \, .
    \end{equation*}
    Suppose that the statement holds for $h+1$ where $h \in [H-1]$. 
    Then, for $h$ and for any $(s,a) \in \Scal \times \Acal$, 
    \begin{align*}
        & \hat{Q}^k_h(s,a)
        \\
        & =  r(s,a) + \sum_{s' \in \Scal_{s,a}} P_{\omdtheta{k}{h}} (s' \mid s, a) \hat{V}^k_{h+1}(s') + \obonus_{k,h}(s,a)
        \\
        & \ge r(s,a) + \sum_{s' \in \Scal_{s,a}} P_{\omdtheta{k}{h}} (s' \mid s, a) V^*_{h+1}(s') + \obonus_{k,h}(s,a)
        \\
        & = r(s,a) + \sum_{s' \in \Scal_{s,a}} P_{\thetab^*_h} (s' \mid s, a) V^*_{h+1}(s')
        \\
        &\phantom{-} + \sum_{s' \in \Scal_{s,a}} \left( P_{\omdtheta{k}{h}} (s' \mid s, a) - P_{\thetab^*_h} (s' \mid s, a) \right) V^*_{h+1}(s')
        + \obonus_{k,h}(s,a)
        \\
        & \ge r(s,a) + \sum_{s' \in \Scal_{s,a}} P_{\thetab^*_h} (s' \mid s, a) V^*_{h+1}(s')
        \\
        & = Q^*_h(s,a) \, ,
    \end{align*}
    where the first inequality follows from the induction hypothesis and the second inequality holds by Lemma~\ref{lemma:prediction error bound (alg2)}.
\end{proof}

\section{Experiment Details} \label{appx:experiment details}
\begin{figure}[H]
    \tikzstyle{every node}=[font=\small]
    \centering
    \resizebox{\columnwidth}{!}{
    \begin{tikzpicture}[->,>=stealth',shorten >=2pt, 
        line width=0.7 pt, node distance=1.6cm,
        scale=1, 
        transform shape, align=center, 
        state/.style={circle, minimum size=0.5cm, text width=5mm}]
        \node[state, draw] (one) {$s_1$};
        \node[state, draw] (two) [right of=one] {$s_2$};
        \node[state, draw=none] (dots) [right of=two] {$...$};
        
        \node[state, draw] (n-1) [right of=dots] {$s_{n-1}$};
        \node[state, draw] (n) [right of=n-1] {$s_{n}$};

        \path (one) edge [ loop above ] node {$0.4$} (one) ;
        \path (one) edge [ bend left ] node [above]{$0.6$} (two) ;
        \draw[->] (two.145) [bend left] to node[below]{$0.05$} (one.35);
        \path (two) edge [ loop above ] node {$0.6$} (two) ;
        \path (two) edge [ bend left ] node [above]{$0.35$} (dots) ;
        \draw[->] (dots.145) [bend left] to node[below]{$0.05$} (two.35); 
        \path (n-1) edge [ loop above ] node {$0.6$} (n-1) ;
        \path (dots) edge [ bend left ] node [above]{$0.35$} (n-1) ;
        \draw[->] (n-1.145) [bend left] to node[below]{$0.05$} (dots.35); 
        \path (n-1) edge [ bend left ] node [above]{$0.35$} (n) ;
        \draw[->] (n.145) [bend left] to node[below]{$0.4$} (n-1.35); 
        
        \path[densely dashed] (one) edge [ loop left ] node {$(1, \red{r=\frac{5}{1000}})$} (one) ;
        \draw[densely dashed, ->] (two.-100) [bend left] to node[below]{$1$} (one.-80);
        \draw[densely dashed, ->] (dots.-100) [bend left] to node[below]{$1$} (two.-80);
        \draw[densely dashed, ->] (n-1.-100) [bend left] to node[below]{$1$} (dots.-80);
        \draw[densely dashed, ->] (n.-100) [bend left] to node[below]{$1$} (n-1.-80);
        
        \path (n) edge [ loop right ] node {$(0.6, \blue{r=1})$} (n);    
    \end{tikzpicture}
    }
     \caption{The ``RiverSwim'' environment with $n$ states~\citep{osband2013more}}
    \label{fig:riverswim env}
\end{figure}

The RiverSwim environment (Figure~\ref{fig:riverswim env}) consists of $n$ states that are arranged in a chain.    
The agent starts in the leftmost state with a relatively small reward of $0.005$ and aims to reach the rightmost state, which has a relatively large reward of $1$.
Choosing to swim to the left moves the agent deterministically to the left, while swimming to the right has a probability of transitioning the agent toward the right state, but also a high chance of remaining in the current state or even moving left due to the strong current of river.
Therefore, efficient exploration is crucial in order to learn the optimal policy for this environment. 

We fine-tuned the hyperparameters for each algorithm within specific ranges. 
Figures~\ref{fig:exp_instance_1} and~\ref{fig:exp_instance_2} show the episodic returns in the RiverSwim environment over 10 independent runs with $|\Scal| = 4, H = 12$, and $K = 10,000$ and $|\Scal| = 8, H = 24$, and $K = 10,000$, respectively.
The shaded areas represent the standard deviations (1-sigma error). 
Figure~\ref{fig:exp_barchart} compares the running time of the algorithms over the first 1,000 episodes. All experiments were conducted on a Xeon(R) Gold 6226R CPU @ 2.90GHz (16 cores).


\section{Auxiliary Lemmas}
\begin{lemma}[Determinant-trace inequality~\cite{abbasi2011improved}] \label{lemma:det-trace ineq}
    Suppose $\xb_1, \ldots, \xb_t \in \RR^d$ and for any $1 \le \tau \le t$, $\| \xb_\tau \|_2 \le L$. 
    Let $\Vb_t = \lambda \Ib_d + \sum_{\tau = 1}^t \xb_\tau \xb_\tau^\top$ for some $\lambda > 0$.
    Then, 
    \begin{equation*}
        \det (\Vb_t) \le (\lambda + t L^2/d)^d \, . 
    \end{equation*}    
\end{lemma}

\begin{lemma}[Freedman's inequality~\cite{freedman1975tail}] \label{lemma:freedman ineq}
    Consider a real-valued martingale $\{Y_k : k = 0,1,2, \ldots \}$ with difference sequence $\{ X_k : k=0,1,2,3,\ldots\}$. 
    Assume that the difference sequence is uniformly bounded, $X_k \le R$ almost surely for $k=1,2,3,\ldots$.
    Define the predictable quadratic variation process of the martingale:
    \begin{equation*}
        W_k := \sum_{j=1}^k \EE_{j-1} [ X_j^2] \quad \text{ for } k=1,2,3, \ldots .
    \end{equation*}
    Then, for all $t \ge 0$ and $\sigma^2 > 0$,
    \begin{equation*}
        \PP \left( \exists k \ge 0 : Y_k \ge t \, \text{ and } \, W_k \le \sigma^2 \right) \le \exp \left( - \frac{-t^2 /2}{\sigma^2 + Rt/3} \right) \, .
    \end{equation*}
\end{lemma}

\begin{lemma}[Gaussian noise concentration (Lemma D.2 in \cite{ishfaq2021randomized})] \label{auxiliary lemma: gaussian noise concentration}
    Let $\xib^{(1)}, \xib^{(2)}, \ldots, \xib^{(M)}$ be $M$ independent $d$-dimensional multivariate normal distributed vector with mean $\zero_d$ and covariance $\sigma^2 \Ab^{-1}$ for some $\sigma > 0 $ and a positive definite matrix $\Ab^{-1}$, i.e., $ \xib^{(m)} \sim \Ncal (\zero_d, \sigma^2 \Ab^{-1})$ for $m \in [M]$. 
    Then for any $\delta \in (0,1) $, with probability at least $1 - \delta$, we have
        \begin{equation*}
            \max_{m \in [M]} \| \xib^{(m)} \|_\Ab \le C_{\xib} \sigma \sqrt{d \log (M d/ \delta) } := \gamma(\delta) \, ,
        \end{equation*}
    where $C_{\xib}$ is an absolute constant.
\end{lemma}

\begin{lemma}[Proposition 4.1 of~\citealp{campolongo2020temporal}]
\label{lemma:loss_firstorder_decomposition}
    Let the $w_{t+1}$ be the solution of the update rule
    \[
    w_{t+1} = \arg\min_{w \in \mathcal{V}} \eta\ell_t(w) + D_{\psi}(w, w_t),
    \]
    where $\mathcal{V} \subseteq \mathcal{W} \subseteq \mathbb{R}^d$ is a non-empty convex set and $D_{\psi}(w_1, w_2) = \psi(w_1) - \psi(w_2) - \langle \nabla\psi(w_2), w_1 - w_2 \rangle$ is the Bregman Divergence w.r.t. a strictly convex and continuously differentiable function $\psi : \mathcal{W} \rightarrow \mathbb{R}$. Further supposing $\psi(w)$ is $1$-strongly convex w.r.t. a certain norm $\|\cdot\|$ in $\mathcal{W}$, then there exists a $g_t \in \partial\ell_t(w_{t+1})$ such that
    \begin{align*}
        \langle \eta_tg_t', w_{t+1} - u \rangle 
        \leq \langle \nabla \psi(w_t) - \nabla \psi(w_{t+1}), w_{t+1} - u \rangle     
    \end{align*}
    for any $u \in \mathcal{W}$.
\end{lemma}
\begin{lemma} \label{lemma:zhang_lemma15}
Let $\{\mathcal{F}_t\}_{t=1}^{\infty}$ be a filtration. Let $\{\zb_t\}_{t=1}^{\infty}$ be a stochastic process in $\mathcal{B}_2(\Ucal) = \{\zb \in \mathbb{R}^\Ucal \mid \|\zb\|_{\infty} \leq 1\}$ such that $\zb_t$ is $\mathcal{F}_t$ measurable. 
Let $\{\boldsymbol{\varepsilon}_t\}_{t=1}^{\infty}$ be a martingale difference sequence such that $\boldsymbol{\varepsilon}_t \in \mathbb{R}^\Ucal$ is $\mathcal{F}_{t+1}$ measurable. 
Furthermore, assume that conditional on $\mathcal{F}_t$, we have $\|\boldsymbol{\varepsilon}_{t}\|_1 \leq 2$ almost surely, and denote by $\Sigma_t = \mathbb{E}[\boldsymbol{\varepsilon}_t \boldsymbol{\varepsilon}_t^\top | \mathcal{F}_t]$. 
Let $\lambda > 0$ and for any $t \geq 1$ define
    \begin{align*}
    U_t = \sum_{i=1}^{t-1} \langle \boldsymbol{\varepsilon}_i, \zb_i \rangle \quad \text{and} 
    \quad \Bb_t = \lambda + \sum_{i=1}^{t-1} \|\zb_i\|_{\Sigma_i}^2,        
    \end{align*}
    Then, for any $\delta \in (0,1]$, we have
    \begin{align*}
    \Pr\left[ \exists t \geq 1, U_t \geq \sqrt{\Bb_t} \left( \frac{\sqrt{\lambda}}{4} + \frac{4}{\sqrt{\lambda}} \log \left( \sqrt{\frac{\Bb_t}{\lambda}} \right) + \frac{4}{\sqrt{\lambda}} \log\left(\frac{2}{\delta}\right) \right) \right] \leq \delta.        
    \end{align*}
\end{lemma}
\begin{lemma}[Lemma 1 of \citealp{zhang2023online}] \label{lemma:zhang_lemma1}
    Let $\ell(\zb, y) = \sum_{k=0}^{K} \mathbf{1}\{y = k\} \cdot \log\left(\frac{1}{[\sigma(\zb)]_k}\right)$, $\mathbf{a} \in [-C,C]^K$, $y \in \{ 0\}\cup [K]$ and $\mathbf{b} \in \RR^K$ where $C>0$.
    Then, we have
    \begin{align*}
        \ell(\mathbf{a}, y) \geq 
        \ell(\mathbf{b}, y) + \nabla \ell(\mathbf{b}, y)^\top (\mathbf{a} - \mathbf{b})
        + \frac{1}{\log(K+1) + 2(C+1)} (\mathbf{a} - \mathbf{b})^\top \nabla^2 \ell(\mathbf{b},y)  (\mathbf{a} - \mathbf{b}).
    \end{align*}
\end{lemma}
\begin{lemma}[Lemma 17 of \citealp{zhang2023online}] \label{lemma:zhang_lemma17}
    Let $\ell(\zb, y) = \sum_{k=0}^{K} \mathbf{1}\{y = k\} \cdot \log\left(\frac{1}{[\sigma(\zb)]_k}\right)$ and $\zb \in \mathbb{R}^K$ be a $K$-dimensional vector. 
    Define $\zb^{\mu} \triangleq \sigma^+\left(\operatorname{smooth}_\mu(\sigma(\zb))\right)$, where $\operatorname{smooth}_\mu(\mathbf{p}) = (1 - \mu)\mathbf{p} + \mu \mathbf{1}/(K+1)$. Then, for $\mu \in [0, 1/2]$, we have
    \begin{align*}
        \ell(\zb^{\mu}, y) - \ell(\zb, y) \leq 2\mu
    \end{align*}
    for any $y \in \{0\} \cup [K]$. We also have $\|\zb^{\mu}\|_{\infty} \leq \log(K/\mu)$.
\end{lemma}
\begin{lemma}[Lemma 18 of \citealp{zhang2023online}] \label{lemma:zhang_lemma18}
    Let $L_{i,h}(\thetab) := \ell_{i,h}(\thetab) + \frac{1}{2c} \| \thetab - \thetab^{i}_h \|_{\Bb_{i,h}}^2$. 
    Assume that $\ell_{i,h}$ is a $\sqrt{N}$-self-concordant-like function.
    Then, for any $\thetab,  \thetab^{i}_h \in \Bcal(\zero_d, 1)$, the quadratic approximation $\tilde{L}_{i,h}(\thetab) = L_{i,h}(\omdtheta{i+1}{h}) 
            + \langle \nabla L_{i,h}(\omdtheta{i+1}{h}) , \thetab - \omdtheta{i+1}{h} \rangle
            + \frac{1}{2c}\left\| \thetab - \omdtheta{i+1}{h} \right\|_{\Bb_{i,h}}^2$ satisfies
    \begin{align*}
         L_{i,h}(\thetab) \leq \tilde{L}_{i,h}(\thetab) + \exp \left( N\left\| \thetab - \omdtheta{i+1}{h} \right\|_2^2 \right) \left\| \thetab - \omdtheta{i+1}{h} \right\|_{\nabla \ell_{i,h}(\omdtheta{i+1}{h})}^2.
    \end{align*}
\end{lemma}

\section{Limitations}~\label{appx:limitations}
We make an assumption about the transition model of MDPs by using the MNL model, which is a specific parametric model. This assumption implies that we assume the realizability of the MNL model. It's worth noting that the realizability assumption has also been commonly made in previous literature on provable reinforcement learning with function approximation, including works such as \citep{yang2020reinforcement, jin2020provably, zanette2020frequentist, modi2020sample, du2020is, cai2020provably, ayoub2020model, wang2020reinforcement_eluder, weisz2021exponential, he2021logarithmic, zhou2021nearly, zhou2021provably, ishfaq2021randomized, hwang2023model}. However, we hope that this condition can be relaxed in the future work.

\end{document}